\documentclass[sigconf, screen]{acmart}

\acmYear{2022}
\setcopyright{acmcopyright}
\acmConference[ICMR '22] {Proceedings of the 2022 International Conference on Multimedia Retrieval}{Accepted version}{Newark, NJ, USA.}
\acmBooktitle{Proceedings of the 2022 International Conference on Multimedia Retrieval (ICMR '22), June 27--30, 2022, Newark, NJ, USA}
\acmISBN{978-1-4503-9238-9/22/06}
\acmDOI{10.1145/3512527.3531369}
 
\usepackage{titlesec}
\titlespacing*{\subsection}{0pt}{1.0ex plus 0.2ex minus .2ex}{1.0ex plus .2ex}

\usepackage[aboveskip=2pt, belowskip=-2pt]{subcaption} 
\usepackage{tabu} 
\usepackage{etex}
\usepackage{times}
\usepackage{epsfig} 
\usepackage{amsmath}
\usepackage{mathrsfs}
\usepackage{datatool}
\usepackage{booktabs}
\usepackage{graphicx}
\usepackage{caption,tabularx,booktabs,calc}
\usepackage{array, booktabs, makecell, multirow}
\usepackage{fnpos}
\usepackage{footmisc}
\usepackage{lipsum}
\usepackage{datatool,filecontents}
\usepackage{xstring}
\usepackage{siunitx} 

\usepackage{bigstrut}

\usepackage{xcolor}
\usepackage{colortbl} 
\usepackage{caption}  
\usepackage{subcaption}  

\usepackage{xspace}

\usepackage{amsbsy}

\usepackage[export]{adjustbox}

\usepackage{multirow}
\usepackage{array}
\setcellgapes{5pt}
\usepackage{arydshln} 
\usepackage{csvsimple} 
\usepackage{enumitem}   
\usepackage{threeparttable,booktabs}
\usepackage{etoolbox}
\appto\TPTnoteSettings{\footnotesize}
\usepackage{csvsimple} 
\usepackage{gensymb}

\usepackage{subcaption}
\usepackage{float}
\usepackage{booktabs,arydshln} 
\usepackage{datatool,filecontents}
\usepackage{xstring}
\usepackage{multirow}
\usepackage{mfirstuc}
\usepackage{hyperref}
\usepackage{xr}

\usepackage{romannum}
\usepackage{tikz}

\makeatletter
\def\adl@drawiv#1#2#3{%
	\hskip.5\tabcolsep
	\xleaders#3{#2.5\@tempdimb #1{1}#2.5\@tempdimb}%
	#2\z@ plus1fil minus1fil\relax
	\hskip.5\tabcolsep}
\newcommand{\cdashlinelr}[1]{%
	\noalign{\vskip\aboverulesep
		\global\let\@dashdrawstore\adl@draw
		\global\let\adl@draw\adl@drawiv}
	\cdashline{#1}
	\noalign{\global\let\adl@draw\@dashdrawstore
		\vskip\belowrulesep}}
\makeatother

\newcommand{\LinearAttenLayer}{linear attention layer}
\newcommand{\LinearAttenLayerCap}{Linear Attention layer}
\newcommand{\PairwiseAttenLayer}{pairwise neighborhood layer}
\newcommand{\LocalNeighSelect}{\textit{Local Neigh. Selection}}

\newcommand{\PairwiseAtten}{pairwise neighborhood attention}
\newcommand{\PairwiseAttenCapLayer}{Pairwise Neighborhood Attention layer}
\newcommand{\PairwiseAttenCap}{Pairwise neighborhood attention}
\newcommand{\RefC}[2]{#1~\cite{#2}}
\newcommand{\etal}{\textit{et al.}}

\usepackage{array}  
\newcolumntype{R}[1]{>{\raggedleft\let\newline\\\arraybackslash\hspace{0pt}}m{#1}} 
\newcolumntype{P}[1]{>{\centering\arraybackslash}p{#1}} 
\newcommand{\OursInputs}[1]{%
	\IfEqCase{#1}{%
		{0}{$\hatvect{f}^s$, $\hatvect{f}^t$}%
		{1}{$\hatvect{f}^s$, $\hatvect{f}^t$ }
		{2}{$\vect{x}^s,\vect{x}^t$}%
		{#1}{#1}
	}[\PackageError{Ours}{Undefined option to Ours: #1}{}]%
}%

\newcommand{\OursSeps}[1]{%
	\IfEqCase{#1}{%
		{0}{\xmark}%
		{1}{\cmark}
		{2}{\xmark}%
		{#1}{#1}
	}[\PackageError{Ours}{Undefined option to Ours: #1}{}]%
}%

\newcommand{\OursLin}[1]{%
	\IfEqCase{#1}{%
		{3}{$L_1$=8}%
		{4}{$L_1$=8}
		{5}{$L_1$=10}%
		{6}{$L_1$=10}%
		{10}{$L_1$=8}%
		{9}{-}%
		{#1}{#1}
	}[\PackageError{Ours}{Undefined option to Ours: #1}{}]%
}%

\newcommand{\OursPair}[1]{%
	\IfEqCase{#1}{%
		{3}{$L_2$=2}%
		{4}{$L_2$=2}
		{5}{\xmark}%
		{6}{\xmark}%
		{10}{$L_2$=2}%
		{9}{-}%
		{#1}{#1}
	}[\PackageError{Ours}{Undefined option to Ours: #1}{}]%
}%

\newcommand{\IsMatcher}[1]{%
	\IfEqCase{#1}{%
		{3}{\cmark}%
		{4}{\cmark}
		{5}{\cmark}%
		{6}{\cmark}%
		{10}{\cmark}%
		{9}{\cmark}%
		{#1}{#1}
	}[\PackageError{Ours}{Undefined option to Ours: #1}{}]%
}%

\newcommand{\IsFilter}[1]{%
	\IfEqCase{#1}{%
		{3}{\cmark}%
		{4}{\xmark}
		{5}{\cmark}%
		{6}{\xmark}%
		{10}{\cmark}%
		{9}{\cmark}%
		{#1}{#1}
	}[\PackageError{Ours}{Undefined option to Ours: #1}{}]%
}%

\newcommand{\IsFeat}[1]{%
	\IfEqCase{#1}{%
		{3}{64} 
		{4}{64} 
		{5}{64} 
		{6}{64} 
		{10}{256} 
		{9}{256}%
		{#1}{#1}
	}[\PackageError{Ours}{Undefined option to Ours: #1}{}]%
}%

\newcommand{\IsSize}[1]{%
	\IfEqCase{#1}{%
		{3}{S}%
		{4}{S}
		{5}{S}%
		{6}{S}%
		{10}{L}%
		{9}{-}%
		{#1}{#1}
	}[\PackageError{Ours}{Undefined option to Ours: #1}{}]%
}%

\newcommand{\Ours}[1]{%
	\IfEqCase{#1}{%
		{0}{Our \textit{Pair.-w/oSep.}}%
		{1}{Our \textit{Pair.Neigh.}}
		{2}{Our \textit{Pair.-w/oSep-Inp.}}%
		{3}{Our \textit{Pair.Neigh.}}
		{4}{Our \textit{Pair.Neigh.} (\textbf{No Filt.}) }
		{5}{Our \textit{Linear.}}
		{6}{Our \textit{Linear.} (\textbf{No Filt.}) }
		{7}{ Our \textit{Lin.-Pair.}\hspace{22pt}\textbf{+ Dist.Match.+Filt.} }
		{8}{ Our \textit{Lin.-Pair.}(\textit{Prior})~\textbf{+ Dist.Match.+Filt.} }
		{9}{\textbf{Dist.Match.+Filt}~\cite{AdaLAM2020}}
		{10}{Our \textit{Pair.Neigh.}-L}
		{11}{Our \textit{Linear.}}
		{12}{Our \textit{Linear.} + Sinkhorn}
		{13}{Our \textit{Linear.} + Dist.Match.\& Filt.}
		{#1}{#1}
	}[\PackageError{Ours}{Undefined option to Ours: #1}{}]%
}%

\newcommand{\ours}[1]{%
	\IfEqCase{#1}{%
		{0}{\textit{Pair.-w/oSep.}}%
		{1}{\textit{Pair.Neigh.}}
		{2}{\textit{Pair.-w/oSep-Inp.}}%
		{3}{\textit{Pair.Neigh.}}
		{4}{\textit{Pair.Neigh.} (\textbf{No Filt.})}
		{5}{\textit{Linear.}}
		{6}{\textit{Linear.} (\textbf{No Filt.}) }
		{7}{\textit{Lin.-Pair.}\hspace{22pt}\textbf{+ Dist.Match.+Filt.} }
		{8}{\textit{Lin.-Pair.}(\textit{Prior})~\textbf{+ Dist.Match.+Filt.} }
		{9}{\textbf{Dist.Match.+Filt}~\cite{AdaLAM2020}}
		{10}{\textit{Pair.Neigh.}-L}
		{11}{\textit{Linear.}}
		{#1}{#1}
	}[\PackageError{Ours}{Undefined option to Ours: #1}{}]%
}%

\newcommand{\Option}[1]{%
	\IfEqCase{#1}{%
		{0}{ \textbf{\textit{MMA}} }%
		{1}{ \textbf{\textit{Time(ms)}} }
	}[\PackageError{Ours}{Undefined option to Ours: #1}{}]%
}%

\newcommand{\addspace}[1]{%
	\IfEqCase{#1}{%
		{0}{\hspace{15pt}}%
		{1}{\hspace{19pt}}
		{2}{\hspace{9pt}}%
	}[\PackageError{Ours}{Undefined option to Ours: #1}{}]%
}%

\newcommand{\suppl}[1]{\textit{suppl.}~\textit{#1}}

\newcommand{\MMA}{\textit{MMA}}
\newcommand{\NumMatches}{\textit{\#Matches}}
\newcommand{\InlierRat}{\textit{Inl.Ratio}}
\newcommand{\NumParam}{\textit{\#Param.}}
\newcommand{\TotalTime}{\textit{Total Time}}

\newcommand{\Step}[1]{\textit{Step.#1}}
\newcommand{\No}[1]{\textit{No.#1}}

\newcommand{\Hquad}{\hspace{0.5em}} 
\newcommand{\cosim}[1][]{\text{cosim~}} 
 
\newcommand{\elu}[1]{\text{elu}(#1)}

\newcommand{\StructAtt}{\textbf{PairAtt}}
\newcommand{\StructAttCdot}{\textbf{PairAtt}$(\cdot)$~}
\newcommand{\vect}[1]{\boldsymbol{\textit{$#1$}}}
\newcommand{\hatvect}[1]{\boldsymbol{\textit{$\hat{#1}$}}}

\newcommand\numberthis{\addtocounter{equation}{1}\tag{\theequation}}

\newcommand{\eg}{\textit{e.g.}}
\newcommand{\ie}{\textit{i.e.}}
\newcommand{\eqreff}[1]{Eq.~\eqref{#1}}
\newcommand{\Table}[1]{Table~\ref{#1}} 
\newcommand{\Fig}[1]{Fig.~\ref{#1}}

\newcommand{\roundK}[1]{\sisetup{round-precision=3, round-mode = figures}\hphantom{-\,}\num[scientific-notation=true]{#1}}

\newcommand{\roundMK}[1]{\sisetup{round-precision=3, fixed-exponent=6, round-mode = figures}\hphantom{-\,}\num[scientific-notation = fixed]{#1}}

\newcommand{\roundKK}[1]{\sisetup{round-mode=places, round-precision=2, fixed-exponent=2}\hphantom{-\,}\num[scientific-notation = fixed]{#1}}

\newcommand{\intpres}[1]{\sisetup{round-precision=0}\num{#1}} 
\newcommand{\onepres}[1]{\sisetup{round-precision=1}\num{#1}}  
\newcommand{\numDB}[1]{\sisetup{round-precision=2}\num{#1}}
\newcommand{\numTP}[1]{\sisetup{round-precision=3}\num{#1}}

\newcommand{\x}[1][]{\times}  
\newcommand{\mean}[1][]{\text{mean~}}

\newcommand{\degc}[1]{#1\degree}

\usepackage{pifont}
\newcommand{\cmark}{\ding{51}}%
\newcommand{\xmark}{\ding{55}}%

\newcommand{\BoldUndLineLARGDB}[3]{%
	\DTLifnumlt{#1}{#2}{\numDB{#1}}{\DTLifnumlt{#1}{#3}{\bfseries\numDB{#1}}{\bfseries\underline{\numDB{#1}}}} } 

\newcommand{\BoldUndLineLARGDBU}[4]{%
	\DTLifnumlt{#1}{#2}{\numDB{#1}{#4}}{\DTLifnumlt{#1}{#3}{\bfseries{\numDB{#1}{#4}}}{\bfseries\underline{\numDB{#1}{#4}}}} }

\newcommand{\BoldUndLineLARGSingU}[4]{%
	\DTLifnumlt{#1}{#2}{\onepres{#1}{#4}}{\DTLifnumlt{#1}{#3}{\bfseries{\onepres{#1}{#4}}}{\bfseries\underline{\onepres{#1}{#4}}}} }

\newcommand{\BoldUndLineLARGSing}[3]{%
	\DTLifnumlt{#1}{#2}{\onepres{#1}}{\DTLifnumlt{#1}{#3}{\bfseries\onepres{#1}}{\bfseries\underline{\onepres{#1}}}} }

\newcommand{\BoldUndLineLARGK}[3]{%
	\DTLifnumlt{#1}{#2}{\roundK{#1}}{\DTLifnumlt{#1}{#3}{\bfseries\roundK{#1}}{\bfseries\underline{\roundK{#1}}}} }

\newcommand{\BoldUndLineLARGINT}[3]{%
	\DTLifnumlt{#1}{#2}{\intpres{#1}}{\DTLifnumlt{#1}{#3}{\bfseries\intpres{#1}}{\bfseries\underline{\intpres{#1}}}} }

\newcommand{\BoldUndLineLARG}[3]{%
	\DTLifnumlt{#1}{#2}{\num{#1}}{\DTLifnumlt{#1}{#3}{\bfseries\num{#1}}{\bfseries\underline{\num{#1}}}} }

\newcommand{\BoldTextSmall}[4]{%
	\DTLifnumlt{#1}{#2}{\DTLifnumlt{#1}{#3}{\textbf{\underline{#4}}}{\textbf{#4}}}{#4} }

\newcommand{\BoldUndLineSMALLESTINT}[3]{%
	\DTLifnumlt{#1}{#2}{\DTLifnumlt{#1}{#3}{\bfseries{\underline{\intpres{#1}}}}{\bfseries{\intpres{#1}}}}{\intpres{#1}} } 

\newcommand{\BoldUndLineSMALLEST}[3]{%
	\DTLifnumlt{#1}{#2}{\DTLifnumlt{#1}{#3}{\bfseries{\underline{\num{#1}}}}{\bfseries{\num{#1}}}}{\num{#1}} } 

\newcommand{\BoldUndLineSMALLESTThree}[3]{%
	\DTLifnumlt{#1}{#2}{\DTLifnumlt{#1}{#3}{\bfseries{\underline{\numTP{#1}}}}{\bfseries{\numTP{#1}}}}{\numTP{#1}} } 

\newcommand{\BoldUndLineSMALLESTMK}[3]{%
	\DTLifnumlt{#1}{#2}{\DTLifnumlt{#1}{#3}{\bfseries{\underline{\roundMK{#1}}}}{\bfseries{\roundMK{#1}}}}{\roundMK{#1}} } 

\newcommand{\BoldUndLineSMALLESTKK}[3]{%
	\DTLifnumlt{#1}{#2}{\DTLifnumlt{#1}{#3}{\bfseries{\underline{\roundKK{#1}}}}{\bfseries{\roundKK{#1}}}}{\roundKK{#1}} }

\newcommand{\MNN}{\textit{MNN}}

\newcommand{\RegImg}{\textit{\#Reg. Img.}}
\newcommand{\Reproj}{\textit{Reproj. Error}} 
\newcommand{\Track}{\textit{Track. Len.}}

\newcommand{\DensePoint}{\textit{Dense Points}}
\newcommand{\MatchTime}{\textit{Match. Time}}
\newcommand{\LshR}{\rotatebox[origin=c]{180}{$\Lsh$}}

\DTLloaddb{Ablation1}{pics/scalability/abalation1.csv} 
\DTLloaddb{Ablation2}{pics/scalability/abalation2.csv} 
\DTLloaddb{DiffTime}{pics/supp/scalability-2/diff_time.csv} 
\DTLloaddb{HPatch}{pics/HPatch-image-matching/hseq.csv}  

\DTLloaddb{SfM:Fountain}{pics/SfM/csv_3DRecon_Fountain-10K_smallSfM.csv}  
\DTLloaddb{SfM:Herzjesu}{pics/SfM/csv_3DRecon_Herzjesu-10K_smallSfM.csv} 
\DTLloaddb{SfM:SouthBuilding}{pics/SfM/csv_3DRecon_South-Building-10K_smallSfM.csv}

\DTLloaddb{SfM:Madrid}{pics/SfM/csv_3DRecon_Madrid_Metropolis-10K_smallSfM.csv}  
\DTLloaddb{SfM:Gendark}{pics/SfM/csv_3DRecon_Gendarmenmarkt-10K_smallSfM.csv}  
\DTLloaddb{SfM:Tower}{pics/SfM/csv_3DRecon_Tower_of_London-10K_smallSfM.csv}

\DTLloaddb[keys={sg,sgmnet,ourmethod}]{App:VisualResultI}{pics/supp/visual/methods_I.csv} 
\DTLloaddb[keys={sg,sgmnet,ourmethod}]{App:VisualResultV}{pics/supp/visual/methods_v.csv} 

\DTLloaddb{Ablation1Supp}{pics/supp/scalability-3/abalation11.csv} 
\DTLloaddb{Ablation2Supp}{pics/supp/scalability-3/abalation12.csv} 
\DTLloaddb{Ablation3Supp}{pics/supp/scalability-3/abalation13.csv} 

\DTLloaddb{Ablation21Supp}{pics/supp/scalability-3/abalation21.csv} 
\DTLloaddb{Ablation22Supp}{pics/supp/scalability-3/abalation22.csv} 
\DTLloaddb{Ablation23Supp}{pics/supp/scalability-3/abalation23.csv} 

\DTLloaddb{HPatchSupp}{pics/supp/hpatches/hseq.csv}  
\DTLloaddb{HPatchSuppSink}{pics/HPatch-image-matching/hseq-sink.csv}  

\DTLloaddb{SfMSupp:Fountain}{pics/supp/SfM/csv_3DRecon_Fountain-10K_smallSfM.csv}  
\DTLloaddb{SfMSupp:Herzjesu}{pics/supp/SfM/csv_3DRecon_Herzjesu-10K_smallSfM.csv} 
\DTLloaddb{SfMSupp:SouthBuilding}{pics/supp/SfM/csv_3DRecon_South-Building-10K_smallSfM.csv} 

\DTLloaddb{SfMSupp:Madrid}{pics/supp/SfM/csv_3DRecon_Madrid_Metropolis-10K_smallSfM.csv}  
\DTLloaddb{SfMSupp:Gendark}{pics/supp/SfM/csv_3DRecon_Gendarmenmarkt-10K_smallSfM.csv}  
\DTLloaddb{SfMSupp:Tower}{pics/supp/SfM/csv_3DRecon_Tower_of_London-10K_smallSfM.csv}  

\DTLloaddb[keys={SuperGlue,SGMNet,Ours,Image}]{App:VisualResultIup}{pics/supp/visual/methods_i_small_up.csv}  
\DTLloaddb[keys={SuperGlue,SGMNet,Ours,Image}]{App:VisualResultVup}{pics/supp/visual/methods_v_small_up.csv}  

\acmSubmissionID{98}

\begin{document}
\begin{filecontents*}{pics/scalability/abalation1.csv}
,Method,OneK,TwoK,ThreeK,FourK
0,9,41.8,71.4,79.6,76.5
1,6,61.2,72.4,73.5,76.5
2,4,66.3,74.5,73.5,76.5
3,5,48.0,73.5,76.5,77.6
4,3,58.2,72.4,78.6,80.6
5,10,63.3,73.5,74.5,78.6
\end{filecontents*}

\begin{filecontents*}{pics/scalability/abalation2.csv}
,Method,OneK,TwoK,ThreeK,FourK
0,2,59.2,75.5,74.5,77.6
1,0,57.1,71.4,74.5,78.6
2,1,58.2,72.4,78.6,80.6
\end{filecontents*}

\begin{filecontents*}{pics/supp/scalability-2/diff_time.csv}
,Method,NetOne,NetTwo,NetFour,NetEight,NetSix,OneK,TwoK,FourK,EightK,SixTK
0,SuperGlue \cite{SuperGlue_CVPR2020},37.09171630859375,53.304736328125,163.337080078125,751.504921875,3392.434375,6.712849121093747,21.91525390625,78.13959960937498,292.03179687500005,1162.2246875
1,SGMNet \cite{SGMNet_ICCV2021},69.2074853515625,70.380263671875,76.8100732421875,122.199091796875,340.675,10.978525390625009,38.778974609375,140.3785986328125,561.590595703125,2369.50625
2,1,32.92932373046875,33.6403857421875,35.434990234375,44.3169580078125,100.351455078125,6.41973876953125,6.906472167968751,9.861240234374996,22.334345703124995,67.196279296875
3,10,48.3752978515625,50.654990234375,45.4961962890625,96.389755859375,234.65326171875,5.894121093750002,5.240454101562499,9.651445312499995,26.317763671875,83.82900390625002
\end{filecontents*}

\begin{filecontents*}{pics/HPatch-image-matching/hseq.csv}
Meth,Matc,Inli,Rtim,Para,NoteF,NoteM,SpaceVal,Pub
D2-Net \cite{CVPR2019:D2Net},2495.92037037037,0.42033407263207667,487.0580653861717,7635264,,,0,${CVPR}$'19
R2D2 \cite{NIPS2019:R2D2},2051.7166666666667,0.7441303028492827,1755.160609549063,1038899,,,1,${NIPS}$'19
SuperPoint \cite{CVPR2018:SuperPoint},1082.2740740740742,0.6514757406872843,34.3011573932789,1300865,,,2,${CVPR}$'18
NCNet \cite{NCNet_CVPR2017},1480.3537037037038,0.4612552053927761,297.1770542921843,21287281,,,,${CVPR}$'17
Patch2Pix \cite{Patch2pix_CVPR2021},1264.0777777777778,0.7643729003575932,453.06807948924876,31577467,,,,${CVPR}$'21
LoFTR \cite{LOFTR_CVPR2021},4705.698148148148,0.8690239158449509,212.9790438192862,11561456,,,,${CVPR}$'21
SuperPoint + SuperGlue \cite{SuperGlue_CVPR2020},831.6981481481481,0.842752784975722,115.13686828613281,12023297,,,,${CVPR}$'20
SuperPoint + SGMNet \cite{SGMNet_ICCV2021},866.2518518518518,0.8020107801433908,116.1131974114312,31064820,,,,${ICCV}$'21
SuperPoint + Our Pair.-w/o-Sep-Inp,714.6648148148148,0.7965348497734142,68.48863112131754,840960,SuperPoint,2,,-
SuperPoint + Our Pair.-w/o-Sep,697.4185185185186,0.8087594949139317,68.57522231207953,840960,SuperPoint,0,,-
SuperPoint + Our Pair.Neigh.,710.6925925925926,0.8030023774327798,73.52289443545871,840960,SuperPoint,1,,-
\end{filecontents*}

\begin{filecontents*}{pics/SfM/csv_3DRecon_Fountain-10K_smallSfM.csv}
num_reg_images,mean_reproj_error,mean_track_length,num_sparse_points,num_observations,num_dense_points,num_sparse_points_round,num_sparse_points_unit,num_observations_round,num_observations_unit,num_dense_points_round,num_dense_points_unit,MatchingTime,MatchingRunTime,TotalTime,TotalRunTime,name,note,ref,fnote
11,0.960808,5.139617,11374,58458,1835492,11.37,k,58.458,k,1.835,M,00:13:10,790.3253137207031,00:13:14,,SuperGlue,SuperGlue,SuperGlue_CVPR2020,\tnote{\textdagger}
11,0.959837,5.136771,11384,58477,1831535,11.38,k,58.477,k,1.832,M,00:07:22,442.27888732910156,00:07:25,,SuperGlue-10,SuperGlue-10,SuperGlue_CVPR2020,
11,0.959142,4.929624,11879,58559,1840800,11.88,k,58.559,k,1.841,M,00:03:58,238.49054901123046,00:04:03,,SGMNet,SGMNet,SGMNet_ICCV2021,\tnote{\S}
11,0.966314,4.916778,11932,58667,1835075,11.93,k,58.667,k,1.835,M,00:01:52,112.47411056518557,00:01:55,,SGMNet-10,SGMNet-10,SGMNet_ICCV2021,
11,0.909373,5.16597,10044,51887,1831238,10.04,k,51.887,k,1.831,M,00:00:40,40.93387596130371,00:00:41,,$Our$-Lin.Att.$+$Pair.Neigh.Att.(Prior),2,,
11,0.905392,5.155611,9980,51453,1830840,9.98,k,51.453,k,1.831,M,00:00:41,41.84380332946776,00:00:46,,$Our$-Lin.$+$Pair.Att.,0,,
11,0.90334,5.141639,9969,51257,1828599,9.97,k,51.257,k,1.829,M,00:00:41,41.06628139495851,00:00:45,,$Our$-$Lin.$+$Pair.Neigh.Att.,1,,
\end{filecontents*}

\begin{filecontents*}{pics/SfM/csv_3DRecon_Herzjesu-10K_smallSfM.csv}
num_reg_images,mean_reproj_error,mean_track_length,num_sparse_points,num_observations,num_dense_points,num_sparse_points_round,num_sparse_points_unit,num_observations_round,num_observations_unit,num_dense_points_round,num_dense_points_unit,MatchingTime,MatchingRunTime,TotalTime,TotalRunTime,name,note,ref,fnote
8,0.921172,4.450526,8459,37647,1138689,8.46,k,37.647,k,1.139,M,00:07:50,470.57103149414064,00:07:53,,SuperGlue,SuperGlue,SuperGlue_CVPR2020,\tnote{\textdagger}
8,0.921044,4.443894,8484,37702,1147223,8.48,k,37.702,k,1.147,M,00:04:28,268.4166857910156,00:04:30,,SuperGlue-10,SuperGlue-10,SuperGlue_CVPR2020,
8,0.959642,4.15802,9638,40075,1137319,9.64,k,40.075,k,1.137,M,00:02:23,143.43813873291015,00:02:26,,SGMNet,SGMNet,SGMNet_ICCV2021,\tnote{\S}
8,0.955408,4.137699,9804,40566,1136598,9.8,k,40.566,k,1.137,M,00:01:06,66.41989532470703,00:01:08,,SGMNet-10,SGMNet-10,SGMNet_ICCV2021,
8,0.881257,4.543225,7299,33161,1143222,7.3,k,33.161,k,1.143,M,00:00:24,24.181332931518554,00:00:27,,$Our$-Lin.Att.$+$Pair.Neigh.Att.(Prior),2,,
8,0.872208,4.537283,7242,32859,1147495,7.24,k,32.859,k,1.147,M,00:00:24,24.475317077636717,00:00:27,,$Our$-Lin.$+$Pair.Att.,0,,
8,0.873364,4.527263,7281,32963,1142254,7.28,k,32.963,k,1.142,M,00:00:22,22.957648315429694,00:00:23,,$Our$-$Lin.$+$Pair.Neigh.Att.,1,,
\end{filecontents*}

\begin{filecontents*}{pics/SfM/csv_3DRecon_South-Building-10K_smallSfM.csv}
num_reg_images,mean_reproj_error,mean_track_length,num_sparse_points,num_observations,num_dense_points,num_sparse_points_round,num_sparse_points_unit,num_observations_round,num_observations_unit,num_dense_points_round,num_dense_points_unit,MatchingTime,MatchingRunTime,TotalTime,TotalRunTime,name,note,ref,fnote
128,0.947322,7.898724,114400,903614,12525820,114.4,k,903.614,k,12.526,M,11:10:23,40223.225684814446,11:11:53,,SuperGlue,SuperGlue,SuperGlue_CVPR2020,\tnote{\textdagger}
128,0.948647,7.882079,114840,905178,12510149,114.84,k,905.178,k,12.51,M,06:20:43,22843.4656829834,06:21:44,,SuperGlue-10,SuperGlue-10,SuperGlue_CVPR2020,
128,0.979234,6.945302,132234,918405,12389681,132.23,k,918.405,k,12.39,M,03:01:07,10867.305214233398,03:02:36,,SGMNet,SGMNet,SGMNet_ICCV2021,\tnote{\S}
128,0.981177,6.969137,131194,914309,12331824,131.19,k,914.309,k,12.332,M,01:21:18,4878.507625427246,01:22:14,,SGMNet-10,SGMNet-10,SGMNet_ICCV2021,
128,0.836813,8.312234,94788,787900,12395986,94.79,k,787.9,k,12.396,M,00:24:42,1482.8196996307372,00:26:08,,$Our$-Lin.Att.$+$Pair.Neigh.Att.(Prior),2,,
128,0.832466,8.312484,94232,783302,12417699,94.23,k,783.302,k,12.418,M,00:23:05,1385.7566271972653,00:24:26,,$Our$-Lin.$+$Pair.Att.,0,,
128,0.836057,8.269952,95091,786398,12453059,95.09,k,786.398,k,12.453,M,00:22:09,1329.4081134033206,00:23:32,,$Our$-$Lin.$+$Pair.Neigh.Att.,1,,
\end{filecontents*}

\begin{filecontents*}{pics/SfM/csv_3DRecon_Madrid_Metropolis-10K_smallSfM.csv}
num_reg_images,mean_reproj_error,mean_track_length,num_sparse_points,num_observations,num_dense_points,num_sparse_points_round,num_sparse_points_unit,num_observations_round,num_observations_unit,num_dense_points_round,num_dense_points_unit,MatchingTime,MatchingRunTime,TotalTime,TotalRunTime,name,note,ref,fnote
513,1.065812,7.279507,65687,478169,2948949,65.687,k,478.169,k,2.949,M,00:04:51,291.30176577806475,00:08:37,,MNN+Lowe's.,MNN+Lowe's.(baseline),LowesThresholding,
519,1.11303,8.291315,60498,501608,3040413,60.498,k,501.608,k,3.04,M,00:10:44,644.9230838572978,00:13:51,,AdaLAM,AdaLAM,AdaLAM2020,
732,1.239746,7.593043,127790,970315,3388990,127.79,k,970.315,k,3.389,M,05:15:30,18930.395976428983,05:21:45,,SuperGlue-10,SuperGlue-10,SuperGlue_CVPR2020,
759,1.21081,6.993536,137225,959688,3436939,137.225,k,959.688,k,3.437,M,05:27:46,19666.779626129417,05:33:37,,SGMNet-10,SGMNet-10,SGMNet_ICCV2021,
563,1.106189,8.190356,64416,527590,2535906,64.416,k,527.59,k,2.536,M,01:16:21,4581.11477973938,01:18:08,,$Our$-Lin.Att.$+$Pair.Neigh.Att.(Prior),2,,
508,1.113715,8.669722,56301,488114,2879983,56.301,k,488.114,k,2.88,M,02:40:49,9649.46700881958,02:43:09,,$Our$-Lin.$+$Pair.Att.,0,,
523,1.117829,8.512511,60147,512002,3021718,60.147,k,512.002,k,3.022,M,01:22:36,4956.734559249878,01:25:44,,$Our$-$Lin.$+$Pair.Neigh.Att.,1,,
\end{filecontents*}

\begin{filecontents*}{pics/SfM/csv_3DRecon_Gendarmenmarkt-10K_smallSfM.csv}
num_reg_images,mean_reproj_error,mean_track_length,num_sparse_points,num_observations,num_dense_points,num_sparse_points_round,num_sparse_points_unit,num_observations_round,num_observations_unit,num_dense_points_round,num_dense_points_unit,MatchingTime,MatchingRunTime,TotalTime,TotalRunTime,name,note,ref,fnote
1034,1.069284,6.573178,167257,1099410,7115466,167.257,k,1.099,M,7.115,M,00:05:23,323.289823101759,00:12:53,,MNN+Lowe's.,MNN+Lowe's.(baseline),LowesThresholding,
1041,1.135364,8.051043,134435,1082342,6770601,134.435,k,1.082,M,6.771,M,00:20:43,1243.3631982040404,00:28:49,,AdaLAM,AdaLAM,AdaLAM2020,
1060,1.2224,8.358206,190999,1596409,7243640,190.999,k,1.596,M,7.244,M,07:08:39,25719.439620018005,07:17:43,,SuperGlue-10,SuperGlue-10,SuperGlue_CVPR2020,
1124,1.20284,7.530515,221202,1665765,6641237,221.202,k,1.666,M,6.641,M,05:12:59,18779.480979118347,05:22:54,,SGMNet-10,SGMNet-10,SGMNet_ICCV2021,
1039,1.11719,7.894919,142423,1124418,6998626,142.423,k,1.124,M,6.999,M,02:29:20,8960.862035713195,02:36:42,,$Our$-Lin.Att.$+$Pair.Neigh.Att.(Prior),2,,
1039,1.103462,7.911397,142275,1125594,7059049,142.275,k,1.126,M,7.059,M,02:56:32,10592.040603141784,03:04:18,,$Our$-Lin.$+$Pair.Att.,0,,
1030,1.115741,7.920752,141681,1122220,6955495,141.681,k,1.122,M,6.955,M,02:02:18,7338.042459030152,02:09:14,,$Our$-$Lin.$+$Pair.Neigh.Att.,1,,
\end{filecontents*}

\begin{filecontents*}{pics/SfM/csv_3DRecon_Tower_of_London-10K_smallSfM.csv}
num_reg_images,mean_reproj_error,mean_track_length,num_sparse_points,num_observations,num_dense_points,num_sparse_points_round,num_sparse_points_unit,num_observations_round,num_observations_unit,num_dense_points_round,num_dense_points_unit,MatchingTime,MatchingRunTime,TotalTime,TotalRunTime,name,note,ref,fnote
733,1.014239,7.440179,98719,734487,5091296,98.719,k,734.487,k,5.091,M,00:05:49,349.3587542060018,00:10:03,,MNN+Lowe's.,MNN+Lowe's.(baseline),LowesThresholding,
777,1.044331,8.460647,89179,754512,5468015,89.179,k,754.512,k,5.468,M,00:13:28,808.7301130771639,00:17:37,,AdaLAM,AdaLAM,AdaLAM2020,
941,1.145184,7.26506,163989,1191390,5970176,163.989,k,1.191,M,5.97,M,06:22:13,22933.47729429245,06:27:59,,SuperGlue-10,SuperGlue-10,SuperGlue_CVPR2020,
879,1.144753,6.776365,169204,1146588,5663380,169.204,k,1.147,M,5.663,M,05:02:00,18120.057329559328,05:08:38,,SGMNet-10,SGMNet-10,SGMNet_ICCV2021,
766,1.040839,8.507543,89218,759026,5526968,89.218,k,759.026,k,5.527,M,02:17:24,8244.717055168152,02:21:01,,$Our$-Lin.Att.$+$Pair.Neigh.Att.(Prior),2,,
776,1.039546,8.628882,87196,752404,5413714,87.196,k,752.404,k,5.414,M,03:19:36,11976.101388397214,03:23:58,,$Our$-Lin.$+$Pair.Att.,0,,
779,1.03845,8.487473,90481,767955,5433822,90.481,k,767.955,k,5.434,M,01:46:00,6360.91247982025,01:49:46,,$Our$-$Lin.$+$Pair.Neigh.Att.,1,,
\end{filecontents*}

\begin{filecontents*}{pics/supp/visual/methods_I.csv}
SuperGlue,SGMNet,Ours,Image,NoteV,NoteC
NONE,NONE,NONE,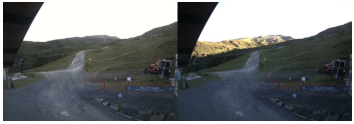,illum.,autannes
SuperGlue,SGMNet,Ours,Image,illum.,autannes
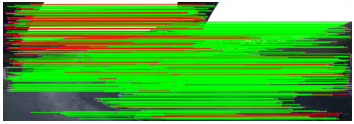,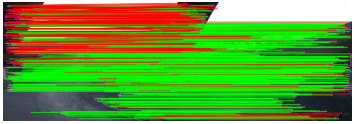,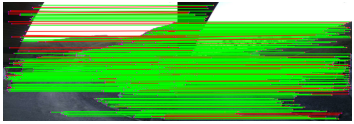,NONE,illum.,autannes
0.8406504065040651,0.7190684133915575,0.8954918032786885,0,illum.,autannes
NONE,NONE,NONE,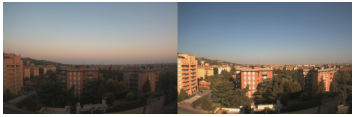,illum.,bologna
SuperGlue,SGMNet,Ours,Image,illum.,bologna
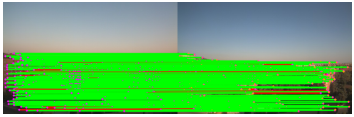,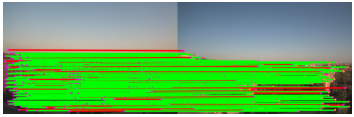,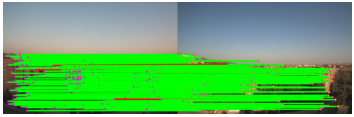,NONE,illum.,bologna
0.972027972027972,0.8945686900958466,0.9795918367346939,0,illum.,bologna
NONE,NONE,NONE,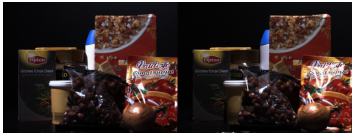,illum.,chestnuts
SuperGlue,SGMNet,Ours,Image,illum.,chestnuts
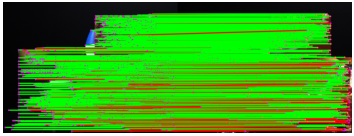,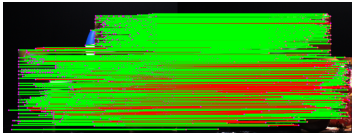,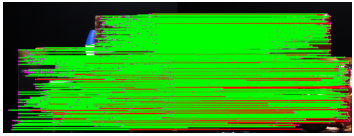,NONE,illum.,chestnuts
0.8701171875,0.845213849287169,0.9161528976572133,0,illum.,chestnuts
NONE,NONE,NONE,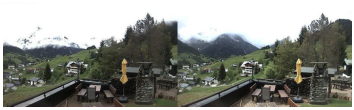,illum.,gonnenberg
SuperGlue,SGMNet,Ours,Image,illum.,gonnenberg
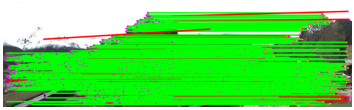,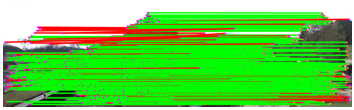,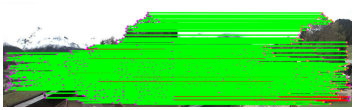,NONE,illum.,gonnenberg
0.9714964370546318,0.920704845814978,0.9875,0,illum.,gonnenberg
NONE,NONE,NONE,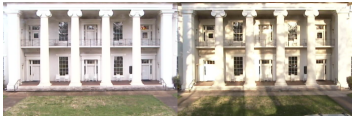,illum.,kions
SuperGlue,SGMNet,Ours,Image,illum.,kions
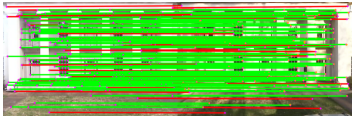,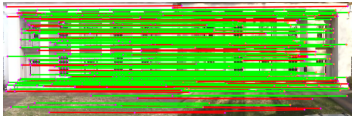,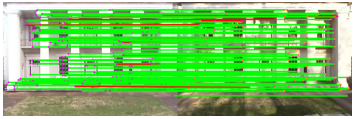,NONE,illum.,kions
0.9019073569482289,0.8434343434343434,0.9136363636363637,0,illum.,kions
NONE,NONE,NONE,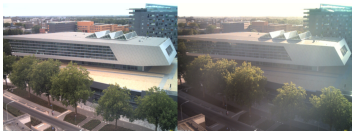,illum.,nijmegen
SuperGlue,SGMNet,Ours,Image,illum.,nijmegen
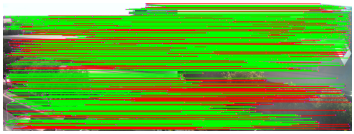,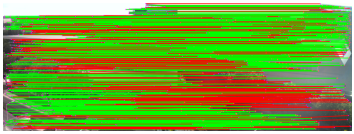,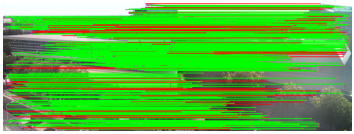,NONE,illum.,nijmegen
0.8268774703557312,0.7833209233060313,0.8669833729216152,0,illum.,nijmegen
NONE,NONE,NONE,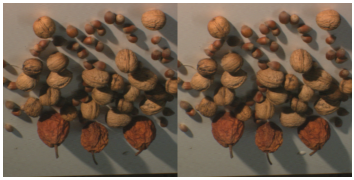,illum.,nuts
SuperGlue,SGMNet,Ours,Image,illum.,nuts
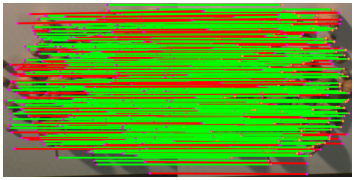,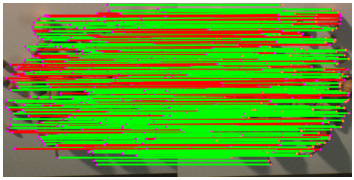,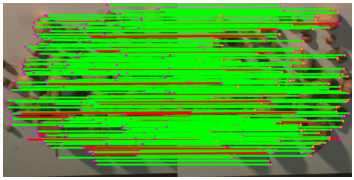,NONE,illum.,nuts
0.7589041095890411,0.7608695652173914,0.8959276018099548,0,illum.,nuts
NONE,NONE,NONE,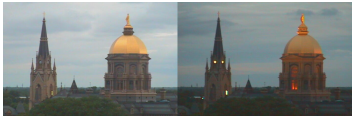,illum.,troulos
SuperGlue,SGMNet,Ours,Image,illum.,troulos
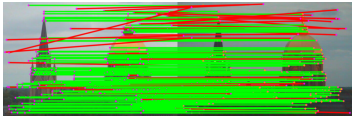,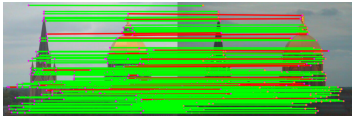,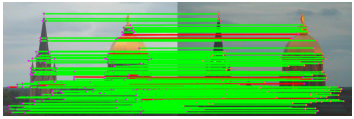,NONE,illum.,troulos
0.8870967741935484,0.9008264462809917,0.9569377990430622,0,illum.,troulos
\end{filecontents*}

\begin{filecontents*}{pics/supp/visual/methods_v.csv}
SuperGlue,SGMNet,Ours,Image,NoteV,NoteC
NONE,NONE,NONE,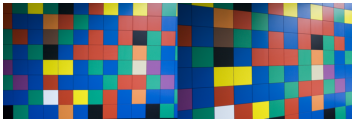,view.,colors
SuperGlue,SGMNet,Ours,Image,view.,colors
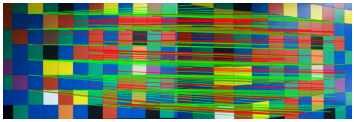,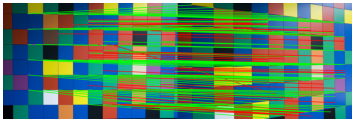,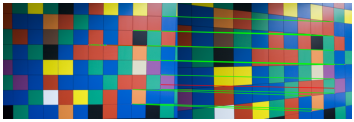,NONE,view.,colors
0.5759493670886076,0.7241379310344828,0.9,0,view.,colors
NONE,NONE,NONE,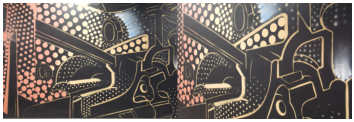,view.,underground
SuperGlue,SGMNet,Ours,Image,view.,underground
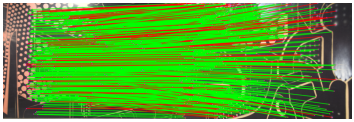,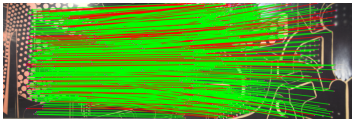,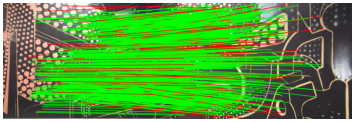,NONE,view.,underground
0.8322981366459627,0.8283450704225352,0.8698412698412699,0,view.,underground
NONE,NONE,NONE,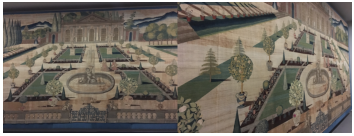,view.,gardens
SuperGlue,SGMNet,Ours,Image,view.,gardens
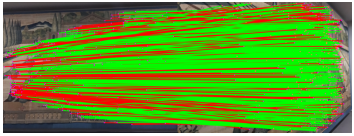,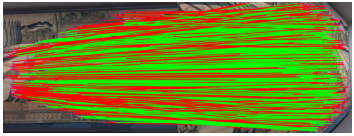,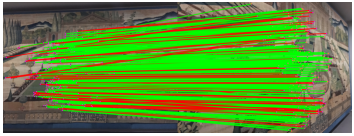,NONE,view.,gardens
0.7559591373439274,0.6885964912280702,0.7857142857142857,0,view.,gardens
NONE,NONE,NONE,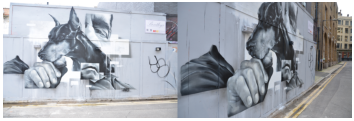,view.,dogman
SuperGlue,SGMNet,Ours,Image,view.,dogman
pics/supp/visual/HPatches_eval_superpoint--SuperGlue-outdoor_viz/v_dogman/drawmatch-Homography-A2B-idx-000004.png,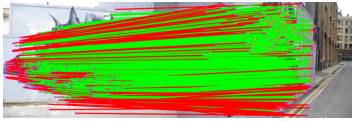,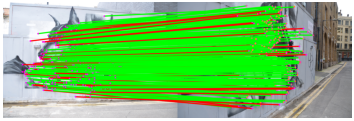,NONE,view.,dogman
0.7680890538033395,0.7249544626593807,0.8314176245210728,0,view.,dogman
NONE,NONE,NONE,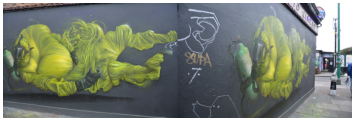,view.,maskedman
SuperGlue,SGMNet,Ours,Image,view.,maskedman
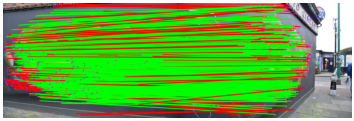,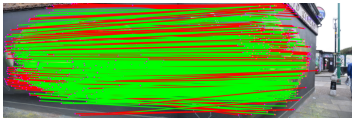,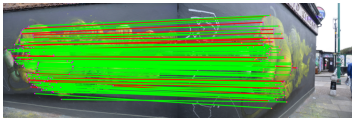,NONE,view.,maskedman
0.7226666666666667,0.6692307692307692,0.8050847457627118,0,view.,maskedman
NONE,NONE,NONE,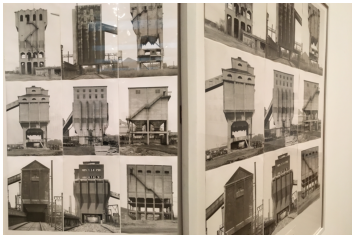,view.,machines
SuperGlue,SGMNet,Ours,Image,view.,machines
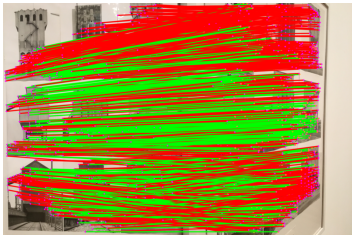,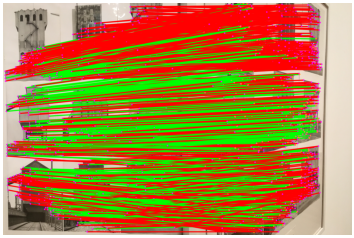,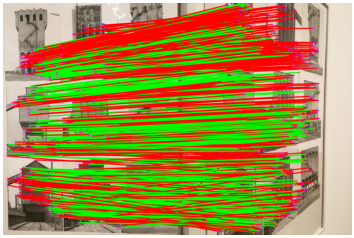,NONE,view.,machines
0.47660311958405543,0.45585738539898135,0.5261904761904762,0,view.,machines
NONE,NONE,NONE,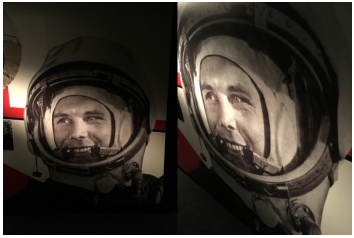,view.,yuri
SuperGlue,SGMNet,Ours,Image,view.,yuri
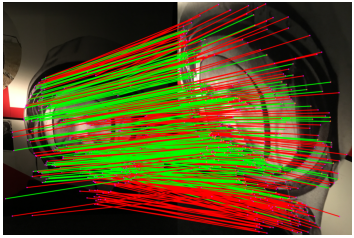,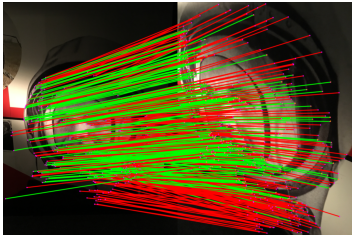,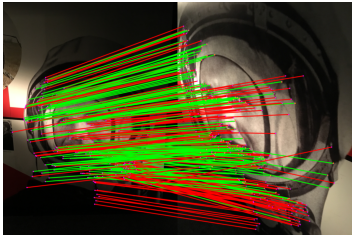,NONE,view.,yuri
0.47884187082405344,0.46320346320346323,0.5121107266435986,0,view.,yuri
\end{filecontents*}

\begin{filecontents*}{pics/supp/scalability-3/abalation11.csv}
,Method,OneK,TwoK,ThreeK,FourK
0,9,41.8,71.4,79.6,76.5
1,6,61.2,72.4,73.5,76.5
2,4,66.3,74.5,73.5,76.5
3,5,48.0,73.5,76.5,77.6
4,3,58.2,72.4,78.6,80.6
5,10,63.3,73.5,74.5,78.6
\end{filecontents*}

\begin{filecontents*}{pics/supp/scalability-3/abalation12.csv}
,Method,OneK,TwoK,ThreeK,FourK
0,9,52.0,80.6,85.7,86.7
1,6,71.4,81.6,81.6,87.8
2,4,74.5,83.7,83.7,83.7
3,5,59.2,82.7,84.7,84.7
4,3,65.3,85.7,86.7,86.7
5,10,70.4,85.7,87.8,87.8
\end{filecontents*}

\begin{filecontents*}{pics/supp/scalability-3/abalation13.csv}
,Method,OneK,TwoK,ThreeK,FourK
0,9,61.2,87.8,92.9,95.9
1,6,75.5,87.8,91.8,95.9
2,4,82.7,91.8,93.9,93.9
3,5,64.3,90.8,92.9,98.9
4,3,73.5,93.9,95.9,95.9
5,10,76.5,94.9,95.9,96.9
\end{filecontents*}

\begin{filecontents*}{pics/supp/scalability-3/abalation21.csv}
,Method,OneK,TwoK,ThreeK,FourK
0,2,59.2,75.5,74.5,77.6
1,0,57.1,71.4,74.5,78.6
2,1,58.2,72.4,78.6,80.6
\end{filecontents*}

\begin{filecontents*}{pics/supp/scalability-3/abalation22.csv}
,Method,OneK,TwoK,ThreeK,FourK
0,2,69.4,84.7,84.7,84.7
1,0,65.3,82.7,86.7,86.7
2,1,65.3,85.7,86.7,86.7
\end{filecontents*}

\begin{filecontents*}{pics/supp/scalability-3/abalation23.csv}
,Method,OneK,TwoK,ThreeK,FourK
0,2,75.5,93.9,94.9,94.9
1,0,71.4,93.9,94.9,95.9
2,1,73.5,93.9,95.9,95.9
\end{filecontents*}

\begin{filecontents*}{pics/supp/hpatches/hseq.csv}
Meth,Matc,Inli,Rtim,Para,NoteF,NoteM,SpaceVal,Pub
D2-Net \cite{CVPR2019:D2Net},2495.92037037037,0.42033407263207667,487.0580653861717,7635264,,,0,${CVPR}$'19
R2D2 \cite{NIPS2019:R2D2},2051.7166666666667,0.7441303028492827,1755.160609549063,1038899,,,1,${NIPS}$'19
SuperPoint \cite{CVPR2018:SuperPoint},1082.2740740740742,0.6514757406872843,34.3011573932789,1300865,,,2,${CVPR}$'18
NCNet \cite{NCNet_CVPR2017},1480.3537037037038,0.4612552053927761,297.1770542921843,21287281,,,,${CVPR}$'17
Patch2Pix \cite{Patch2pix_CVPR2021},1264.0777777777778,0.7643729003575932,453.06807948924876,31577467,,,,${CVPR}$'21
LoFTR \cite{LOFTR_CVPR2021},4705.698148148148,0.8690239158449509,212.9790438192862,11561456,,,,${CVPR}$'21
SuperPoint + SuperGlue \cite{SuperGlue_CVPR2020},831.6981481481481,0.842752784975722,115.13686828613281,12023297,,,,${CVPR}$'20
SuperPoint + SGMNet \cite{SGMNet_ICCV2021},866.2518518518518,0.8020107801433908,116.1131974114312,31064820,,,,${ICCV}$'21
SuperPoint + Our Pair.-w/o-Sep-Inp,714.6648148148148,0.7965348497734142,68.48863112131754,840960,SuperPoint,2,,-
SuperPoint + Our Pair.-w/o-Sep,697.4185185185186,0.8087594949139317,68.57522231207953,840960,SuperPoint,0,,-
SuperPoint + Our Pair.Neigh.,710.6925925925926,0.8030023774327798,73.52289443545871,840960,SuperPoint,1,-,-
SuperPoint + Our Linear.,672.7277777777778,0.8192766605159091,73.59688265058729,924288,SuperPoint,11,,-
SuperPoint + Our Pair.Neigh.-L,718.5722222222222,0.7993989018355728,88.63918230975116,12505344,SuperPoint,10,,-
\end{filecontents*}

\begin{filecontents*}{pics/HPatch-image-matching/hseq-sink.csv}
Meth,Matc,Inli,Rtim,Para,NoteF,NoteM,SpaceVal,Pub,File
SuperPoint + SuperGlue \cite{SuperGlue_CVPR2020},831.6981481481481,0.842752784975722,115.13686828613281,12023297,,,,${CVPR}$`20,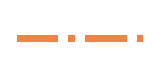
SuperPoint + Our Linear. + Sink.-100,666.7203703703703,0.7642616284913378,110.97875008053249,10530817,SuperPoint,12,,-,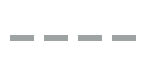
SuperPoint + Our Linear. + Dist.Match. + Filt.,672.7277777777778,0.8192766605159091,73.59688265058729,924288,SuperPoint,13,,-,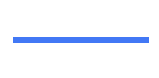
\end{filecontents*}

\begin{filecontents*}{pics/supp/SfM/csv_3DRecon_Fountain-10K_smallSfM.csv}
num_reg_images,mean_reproj_error,mean_track_length,num_sparse_points,num_observations,num_dense_points,num_sparse_points_round,num_sparse_points_unit,num_observations_round,num_observations_unit,num_dense_points_round,num_dense_points_unit,MatchingTime,MatchingRunTime,TotalTime,TotalRunTime,name,note,ref,fnote
11,0.960808,5.139617,11374,58458,1835492,11.37,k,58.458,k,1.835,M,00:13:10,790.3253137207031,00:13:14,,SuperGlue,SuperGlue,SuperGlue_CVPR2020,\tnote{\textdagger}
11,0.959837,5.136771,11384,58477,1831535,11.38,k,58.477,k,1.832,M,00:07:22,442.27888732910156,00:07:25,,SuperGlue-10,SuperGlue-10,SuperGlue_CVPR2020,
11,0.959142,4.929624,11879,58559,1840800,11.88,k,58.559,k,1.841,M,00:03:58,238.49054901123046,00:04:03,,SGMNet,SGMNet,SGMNet_ICCV2021,\tnote{\S}
11,0.966314,4.916778,11932,58667,1835075,11.93,k,58.667,k,1.835,M,00:01:52,112.47411056518557,00:01:55,,SGMNet-10,SGMNet-10,SGMNet_ICCV2021,
11,0.909373,5.16597,10044,51887,1831238,10.04,k,51.887,k,1.831,M,00:00:40,40.93387596130371,00:00:41,,$Our$-Lin.$+$Pair.Neigh.Att.(Prior),2,,
11,0.905392,5.155611,9980,51453,1830840,9.98,k,51.453,k,1.831,M,00:00:41,41.84380332946776,00:00:46,,$Our$-Lin.$+$Pair.Att.,0,,
11,0.90334,5.141639,9969,51257,1828599,9.97,k,51.257,k,1.829,M,00:00:41,41.06628139495851,00:00:45,,$Our$-Lin.$+$Pair.Neigh.Att.,1,,
11,0.889774,5.121098,9802,50197,1825564,9.8,k,50.197,k,1.826,M,00:00:42,42.97651268005371,00:00:47,,$Our$-Linear.,11,,
11,0.900916,5.124415,10039,51444,1833357,10.04,k,51.444,k,1.833,M,00:01:11,71.26612045288086,00:01:15,,$Our$-Lin.$+$Pair.Att.-L,10,,
\end{filecontents*}

\begin{filecontents*}{pics/supp/SfM/csv_3DRecon_Herzjesu-10K_smallSfM.csv}
num_reg_images,mean_reproj_error,mean_track_length,num_sparse_points,num_observations,num_dense_points,num_sparse_points_round,num_sparse_points_unit,num_observations_round,num_observations_unit,num_dense_points_round,num_dense_points_unit,MatchingTime,MatchingRunTime,TotalTime,TotalRunTime,name,note,ref,fnote
8,0.921172,4.450526,8459,37647,1138689,8.46,k,37.647,k,1.139,M,00:07:50,470.57103149414064,00:07:53,,SuperGlue,SuperGlue,SuperGlue_CVPR2020,\tnote{\textdagger}
8,0.921044,4.443894,8484,37702,1147223,8.48,k,37.702,k,1.147,M,00:04:28,268.4166857910156,00:04:30,,SuperGlue-10,SuperGlue-10,SuperGlue_CVPR2020,
8,0.959642,4.15802,9638,40075,1137319,9.64,k,40.075,k,1.137,M,00:02:23,143.43813873291015,00:02:26,,SGMNet,SGMNet,SGMNet_ICCV2021,\tnote{\S}
8,0.955408,4.137699,9804,40566,1136598,9.8,k,40.566,k,1.137,M,00:01:06,66.41989532470703,00:01:08,,SGMNet-10,SGMNet-10,SGMNet_ICCV2021,
8,0.881257,4.543225,7299,33161,1143222,7.3,k,33.161,k,1.143,M,00:00:24,24.181332931518554,00:00:27,,$Our$-Lin.$+$Pair.Neigh.Att.(Prior),2,,
8,0.872208,4.537283,7242,32859,1147495,7.24,k,32.859,k,1.147,M,00:00:24,24.475317077636717,00:00:27,,$Our$-Lin.$+$Pair.Att.,0,,
8,0.873364,4.527263,7281,32963,1142254,7.28,k,32.963,k,1.142,M,00:00:22,22.957648315429694,00:00:23,,$Our$-Lin.$+$Pair.Neigh.Att.,1,,
8,0.85719,4.518404,7118,32162,1147475,7.12,k,32.162,k,1.147,M,00:00:23,23.26359580993653,00:00:26,,$Our$-Linear.,11,,
8,0.87738,4.531464,7310,33125,1144713,7.31,k,33.125,k,1.145,M,00:00:39,39.49683166503906,00:00:42,,$Our$-Lin.$+$Pair.Att.-L,10,,
\end{filecontents*}

\begin{filecontents*}{pics/supp/SfM/csv_3DRecon_South-Building-10K_smallSfM.csv}
num_reg_images,mean_reproj_error,mean_track_length,num_sparse_points,num_observations,num_dense_points,num_sparse_points_round,num_sparse_points_unit,num_observations_round,num_observations_unit,num_dense_points_round,num_dense_points_unit,MatchingTime,MatchingRunTime,TotalTime,TotalRunTime,name,note,ref,fnote
128,0.947322,7.898724,114400,903614,12525820,114.4,k,903.614,k,12.526,M,11:10:23,40223.225684814446,11:11:53,,SuperGlue,SuperGlue,SuperGlue_CVPR2020,\tnote{\textdagger}
128,0.948647,7.882079,114840,905178,12510149,114.84,k,905.178,k,12.51,M,06:20:43,22843.4656829834,06:21:44,,SuperGlue-10,SuperGlue-10,SuperGlue_CVPR2020,
128,0.979234,6.945302,132234,918405,12389681,132.23,k,918.405,k,12.39,M,03:01:07,10867.305214233398,03:02:36,,SGMNet,SGMNet,SGMNet_ICCV2021,\tnote{\S}
128,0.981177,6.969137,131194,914309,12331824,131.19,k,914.309,k,12.332,M,01:21:18,4878.507625427246,01:22:14,,SGMNet-10,SGMNet-10,SGMNet_ICCV2021,
128,0.836813,8.312234,94788,787900,12395986,94.79,k,787.9,k,12.396,M,00:24:42,1482.8196996307372,00:26:08,,$Our$-Lin.$+$Pair.Neigh.Att.(Prior),2,,
128,0.832466,8.312484,94232,783302,12417699,94.23,k,783.302,k,12.418,M,00:23:05,1385.7566271972653,00:24:26,,$Our$-Lin.$+$Pair.Att.,0,,
128,0.836057,8.269952,95091,786398,12453059,95.09,k,786.398,k,12.453,M,00:22:09,1329.4081134033206,00:23:32,,$Our$-Lin.$+$Pair.Neigh.Att.,1,,
128,0.820359,8.322275,92207,767372,12379455,92.21,k,767.372,k,12.379,M,00:24:02,1442.6152793884276,00:25:24,,$Our$-Linear.,11,,
128,0.84058,8.238788,96102,791764,12424552,96.1,k,791.764,k,12.425,M,00:46:11,2771.0120755004878,00:47:31,,$Our$-Lin.$+$Pair.Att.-L,10,,
\end{filecontents*}

\begin{filecontents*}{pics/supp/SfM/csv_3DRecon_Madrid_Metropolis-10K_smallSfM.csv}
num_reg_images,mean_reproj_error,mean_track_length,num_sparse_points,num_observations,num_dense_points,num_sparse_points_round,num_sparse_points_unit,num_observations_round,num_observations_unit,num_dense_points_round,num_dense_points_unit,MatchingTime,MatchingRunTime,TotalTime,TotalRunTime,name,note,ref,fnote
513,1.065812,7.279507,65687,478169,2948949,65.687,k,478.169,k,2.949,M,00:04:51,291.30176577806475,00:08:37,,MNN+Lowe's.,MNN+Lowe's.(baseline),LowesThresholding,
519,1.11303,8.291315,60498,501608,3040413,60.498,k,501.608,k,3.04,M,00:10:44,644.9230838572978,00:13:51,,AdaLAM,AdaLAM,AdaLAM2020,
732,1.239746,7.593043,127790,970315,3388990,127.79,k,970.315,k,3.389,M,05:15:30,18930.395976428983,05:21:45,,SuperGlue-10,SuperGlue-10,SuperGlue_CVPR2020,
759,1.21081,6.993536,137225,959688,3436939,137.225,k,959.688,k,3.437,M,05:27:46,19666.779626129417,05:33:37,,SGMNet-10,SGMNet-10,SGMNet_ICCV2021,
563,1.106189,8.190356,64416,527590,2535906,64.416,k,527.59,k,2.536,M,01:16:21,4581.11477973938,01:18:08,,$Our$-Lin.$+$Pair.Neigh.Att.(Prior),2,,
508,1.113715,8.669722,56301,488114,2879983,56.301,k,488.114,k,2.88,M,02:40:49,9649.46700881958,02:43:09,,$Our$-Lin.$+$Pair.Att.,0,,
523,1.117829,8.512511,60147,512002,3021718,60.147,k,512.002,k,3.022,M,01:22:36,4956.734559249878,01:25:44,,$Our$-Lin.$+$Pair.Neigh.Att.,1,,
552,1.089818,8.257163,58492,482978,2941514,58.492,k,482.978,k,2.942,M,01:08:09,4089.5646510696406,01:11:06,,$Our$-Linear.,11,,
562,1.11095,8.338221,63630,530561,2978097,63.63,k,530.561,k,2.978,M,01:51:17,6677.901119461061,01:54:25,,$Our$-Lin.$+$Pair.Att.-L,10,,
\end{filecontents*}

\begin{filecontents*}{pics/supp/SfM/csv_3DRecon_Gendarmenmarkt-10K_smallSfM.csv}
num_reg_images,mean_reproj_error,mean_track_length,num_sparse_points,num_observations,num_dense_points,num_sparse_points_round,num_sparse_points_unit,num_observations_round,num_observations_unit,num_dense_points_round,num_dense_points_unit,MatchingTime,MatchingRunTime,TotalTime,TotalRunTime,name,note,ref,fnote
1034,1.069284,6.573178,167257,1099410,7115466,167.257,k,1.099,M,7.115,M,00:05:23,323.289823101759,00:12:53,,MNN+Lowe's.,MNN+Lowe's.(baseline),LowesThresholding,
1041,1.135364,8.051043,134435,1082342,6770601,134.435,k,1.082,M,6.771,M,00:20:43,1243.3631982040404,00:28:49,,AdaLAM,AdaLAM,AdaLAM2020,
1060,1.2224,8.358206,190999,1596409,7243640,190.999,k,1.596,M,7.244,M,07:08:39,25719.439620018005,07:17:43,,SuperGlue-10,SuperGlue-10,SuperGlue_CVPR2020,
1124,1.20284,7.530515,221202,1665765,6641237,221.202,k,1.666,M,6.641,M,05:12:59,18779.480979118347,05:22:54,,SGMNet-10,SGMNet-10,SGMNet_ICCV2021,
1039,1.11719,7.894919,142423,1124418,6998626,142.423,k,1.124,M,6.999,M,02:29:20,8960.862035713195,02:36:42,,$Our$-Lin.$+$Pair.Neigh.Att.(Prior),2,,
1039,1.103462,7.911397,142275,1125594,7059049,142.275,k,1.126,M,7.059,M,02:56:32,10592.040603141784,03:04:18,,$Our$-Lin.$+$Pair.Att.,0,,
1030,1.115741,7.920752,141681,1122220,6955495,141.681,k,1.122,M,6.955,M,02:02:18,7338.042459030152,02:09:14,,$Our$-Lin.$+$Pair.Neigh.Att.,1,,
1027,1.101499,7.94567,133205,1058403,6997877,133.205,k,1.058,M,6.998,M,01:57:50,7070.0725672531125,02:04:41,,$Our$-Linear.,11,,
1001,1.134002,8.111485,132170,1072095,6672701,132.17,k,1.072,M,6.673,M,02:45:46,9946.947644424437,02:52:40,,$Our$-Lin.$+$Pair.Att.-L,10,,
\end{filecontents*}

\begin{filecontents*}{pics/supp/SfM/csv_3DRecon_Tower_of_London-10K_smallSfM.csv}
num_reg_images,mean_reproj_error,mean_track_length,num_sparse_points,num_observations,num_dense_points,num_sparse_points_round,num_sparse_points_unit,num_observations_round,num_observations_unit,num_dense_points_round,num_dense_points_unit,MatchingTime,MatchingRunTime,TotalTime,TotalRunTime,name,note,ref,fnote
733,1.014239,7.440179,98719,734487,5091296,98.719,k,734.487,k,5.091,M,00:05:49,349.3587542060018,00:10:03,,MNN+Lowe's.,MNN+Lowe's.(baseline),LowesThresholding,
777,1.044331,8.460647,89179,754512,5468015,89.179,k,754.512,k,5.468,M,00:13:28,808.7301130771639,00:17:37,,AdaLAM,AdaLAM,AdaLAM2020,
941,1.145184,7.26506,163989,1191390,5970176,163.989,k,1.191,M,5.97,M,06:22:13,22933.47729429245,06:27:59,,SuperGlue-10,SuperGlue-10,SuperGlue_CVPR2020,
879,1.144753,6.776365,169204,1146588,5663380,169.204,k,1.147,M,5.663,M,05:02:00,18120.057329559328,05:08:38,,SGMNet-10,SGMNet-10,SGMNet_ICCV2021,
766,1.040839,8.507543,89218,759026,5526968,89.218,k,759.026,k,5.527,M,02:17:24,8244.717055168152,02:21:01,,$Our$-Lin.$+$Pair.Neigh.Att.(Prior),2,,
776,1.039546,8.628882,87196,752404,5413714,87.196,k,752.404,k,5.414,M,03:19:36,11976.101388397214,03:23:58,,$Our$-Lin.$+$Pair.Att.,0,,
779,1.03845,8.487473,90481,767955,5433822,90.481,k,767.955,k,5.434,M,01:46:00,6360.91247982025,01:49:46,,$Our$-Lin.$+$Pair.Neigh.Att.,1,,
796,1.023506,8.520808,85425,727890,5500840,85.425,k,727.89,k,5.501,M,01:21:16,4876.837587928772,01:24:44,,$Our$-Linear.,11,,
768,1.045772,8.4947,91407,776475,5449753,91.407,k,776.475,k,5.45,M,02:22:14,8534.310445289613,02:25:59,,$Our$-Lin.$+$Pair.Att.-L,10,,
\end{filecontents*}

\begin{filecontents*}{pics/supp/visual/methods_i_small_up.csv}
SuperGlue,SGMNet,Ours,Image,NoteV,NoteC
SuperGlue,SGMNet,Ours,Autannes (Original),illum.,autannes
pics/supp/visual/HPatches_eval_superpoint--SuperGlue-outdoor_viz/i_autannes/drawmatch-Homography-A2B-idx-000002.png,pics/supp/visual/HPatches_eval_superpoint--SGM-ICCV2021_viz/i_autannes/drawmatch-Homography-A2B-idx-000002.png,pics/supp/visual/HPatches_eval_superpoint--FullLoFTR-Block-8-SE-2-FixedSeeds_viz/i_autannes/drawmatch-Homography-A2B-idx-000002.png,pics/supp/visual/HPatches_eval_superpoint--NONE_viz/i_autannes/drawmatch-Homography-A2B-idx-000002.png,illum.,autannes
0.8406504065040651,0.7190684133915575,0.8954918032786885,0,illum.,autannes
SuperGlue,SGMNet,Ours,Bologna (Original),illum.,bologna
pics/supp/visual/HPatches_eval_superpoint--SuperGlue-outdoor_viz/i_bologna/drawmatch-Homography-A2B-idx-000006.png,pics/supp/visual/HPatches_eval_superpoint--SGM-ICCV2021_viz/i_bologna/drawmatch-Homography-A2B-idx-000006.png,pics/supp/visual/HPatches_eval_superpoint--FullLoFTR-Block-8-SE-2-FixedSeeds_viz/i_bologna/drawmatch-Homography-A2B-idx-000006.png,pics/supp/visual/HPatches_eval_superpoint--NONE_viz/i_bologna/drawmatch-Homography-A2B-idx-000006.png,illum.,bologna
0.972027972027972,0.8945686900958466,0.9795918367346939,0,illum.,bologna
SuperGlue,SGMNet,Ours,Chestnuts (Original),illum.,chestnuts
pics/supp/visual/HPatches_eval_superpoint--SuperGlue-outdoor_viz/i_chestnuts/drawmatch-Homography-A2B-idx-000005.png,pics/supp/visual/HPatches_eval_superpoint--SGM-ICCV2021_viz/i_chestnuts/drawmatch-Homography-A2B-idx-000005.png,pics/supp/visual/HPatches_eval_superpoint--FullLoFTR-Block-8-SE-2-FixedSeeds_viz/i_chestnuts/drawmatch-Homography-A2B-idx-000005.png,pics/supp/visual/HPatches_eval_superpoint--NONE_viz/i_chestnuts/drawmatch-Homography-A2B-idx-000005.png,illum.,chestnuts
0.8701171875,0.845213849287169,0.9161528976572133,0,illum.,chestnuts
SuperGlue,SGMNet,Ours,Gonnenberg (Original),illum.,gonnenberg
pics/supp/visual/HPatches_eval_superpoint--SuperGlue-outdoor_viz/i_gonnenberg/drawmatch-Homography-A2B-idx-000004.png,pics/supp/visual/HPatches_eval_superpoint--SGM-ICCV2021_viz/i_gonnenberg/drawmatch-Homography-A2B-idx-000004.png,pics/supp/visual/HPatches_eval_superpoint--FullLoFTR-Block-8-SE-2-FixedSeeds_viz/i_gonnenberg/drawmatch-Homography-A2B-idx-000004.png,pics/supp/visual/HPatches_eval_superpoint--NONE_viz/i_gonnenberg/drawmatch-Homography-A2B-idx-000004.png,illum.,gonnenberg
0.9714964370546318,0.920704845814978,0.9875,0,illum.,gonnenberg
SuperGlue,SGMNet,Ours,Kions (Original),illum.,kions
pics/supp/visual/HPatches_eval_superpoint--SuperGlue-outdoor_viz/i_kions/drawmatch-Homography-A2B-idx-000003.png,pics/supp/visual/HPatches_eval_superpoint--SGM-ICCV2021_viz/i_kions/drawmatch-Homography-A2B-idx-000003.png,pics/supp/visual/HPatches_eval_superpoint--FullLoFTR-Block-8-SE-2-FixedSeeds_viz/i_kions/drawmatch-Homography-A2B-idx-000003.png,pics/supp/visual/HPatches_eval_superpoint--NONE_viz/i_kions/drawmatch-Homography-A2B-idx-000003.png,illum.,kions
0.9019073569482289,0.8434343434343434,0.9136363636363637,0,illum.,kions
SuperGlue,SGMNet,Ours,Nijmegen (Original),illum.,nijmegen
pics/supp/visual/HPatches_eval_superpoint--SuperGlue-outdoor_viz/i_nijmegen/drawmatch-Homography-A2B-idx-000004.png,pics/supp/visual/HPatches_eval_superpoint--SGM-ICCV2021_viz/i_nijmegen/drawmatch-Homography-A2B-idx-000004.png,pics/supp/visual/HPatches_eval_superpoint--FullLoFTR-Block-8-SE-2-FixedSeeds_viz/i_nijmegen/drawmatch-Homography-A2B-idx-000004.png,pics/supp/visual/HPatches_eval_superpoint--NONE_viz/i_nijmegen/drawmatch-Homography-A2B-idx-000004.png,illum.,nijmegen
0.8268774703557312,0.7833209233060313,0.8669833729216152,0,illum.,nijmegen
SuperGlue,SGMNet,Ours,Nuts (Original),illum.,nuts
pics/supp/visual/HPatches_eval_superpoint--SuperGlue-outdoor_viz/i_nuts/drawmatch-Homography-A2B-idx-000002.png,pics/supp/visual/HPatches_eval_superpoint--SGM-ICCV2021_viz/i_nuts/drawmatch-Homography-A2B-idx-000002.png,pics/supp/visual/HPatches_eval_superpoint--FullLoFTR-Block-8-SE-2-FixedSeeds_viz/i_nuts/drawmatch-Homography-A2B-idx-000002.png,pics/supp/visual/HPatches_eval_superpoint--NONE_viz/i_nuts/drawmatch-Homography-A2B-idx-000002.png,illum.,nuts
0.7589041095890411,0.7608695652173914,0.8959276018099548,0,illum.,nuts
SuperGlue,SGMNet,Ours,Troulos (Original),illum.,troulos
pics/supp/visual/HPatches_eval_superpoint--SuperGlue-outdoor_viz/i_troulos/drawmatch-Homography-A2B-idx-000002.png,pics/supp/visual/HPatches_eval_superpoint--SGM-ICCV2021_viz/i_troulos/drawmatch-Homography-A2B-idx-000002.png,pics/supp/visual/HPatches_eval_superpoint--FullLoFTR-Block-8-SE-2-FixedSeeds_viz/i_troulos/drawmatch-Homography-A2B-idx-000002.png,pics/supp/visual/HPatches_eval_superpoint--NONE_viz/i_troulos/drawmatch-Homography-A2B-idx-000002.png,illum.,troulos
0.8870967741935484,0.9008264462809917,0.9569377990430622,0,illum.,troulos
\end{filecontents*}

\begin{filecontents*}{pics/supp/visual/methods_v_small_up.csv}
SuperGlue,SGMNet,Ours,Image,NoteV,NoteC
SuperGlue,SGMNet,Ours,Colors (Original),view.,colors
pics/supp/visual/HPatches_eval_superpoint--SuperGlue-outdoor_viz/v_colors/drawmatch-Homography-A2B-idx-000003.png,pics/supp/visual/HPatches_eval_superpoint--SGM-ICCV2021_viz/v_colors/drawmatch-Homography-A2B-idx-000003.png,pics/supp/visual/HPatches_eval_superpoint--FullLoFTR-Block-8-SE-2-FixedSeeds_viz/v_colors/drawmatch-Homography-A2B-idx-000003.png,pics/supp/visual/HPatches_eval_superpoint--NONE_viz/v_colors/drawmatch-Homography-A2B-idx-000003.png,view.,colors
0.5759493670886076,0.7241379310344828,0.9,0,view.,colors
SuperGlue,SGMNet,Ours,Underground (Original),view.,underground
pics/supp/visual/HPatches_eval_superpoint--SuperGlue-outdoor_viz/v_underground/drawmatch-Homography-A2B-idx-000004.png,pics/supp/visual/HPatches_eval_superpoint--SGM-ICCV2021_viz/v_underground/drawmatch-Homography-A2B-idx-000004.png,pics/supp/visual/HPatches_eval_superpoint--FullLoFTR-Block-8-SE-2-FixedSeeds_viz/v_underground/drawmatch-Homography-A2B-idx-000004.png,pics/supp/visual/HPatches_eval_superpoint--NONE_viz/v_underground/drawmatch-Homography-A2B-idx-000004.png,view.,underground
0.8322981366459627,0.8283450704225352,0.8698412698412699,0,view.,underground
SuperGlue,SGMNet,Ours,Gardens (Original),view.,gardens
pics/supp/visual/HPatches_eval_superpoint--SuperGlue-outdoor_viz/v_gardens/drawmatch-Homography-A2B-idx-000003.png,pics/supp/visual/HPatches_eval_superpoint--SGM-ICCV2021_viz/v_gardens/drawmatch-Homography-A2B-idx-000003.png,pics/supp/visual/HPatches_eval_superpoint--FullLoFTR-Block-8-SE-2-FixedSeeds_viz/v_gardens/drawmatch-Homography-A2B-idx-000003.png,pics/supp/visual/HPatches_eval_superpoint--NONE_viz/v_gardens/drawmatch-Homography-A2B-idx-000003.png,view.,gardens
0.7559591373439274,0.6885964912280702,0.7857142857142857,0,view.,gardens
SuperGlue,SGMNet,Ours,Dogman (Original),view.,dogman
pics/supp/visual/HPatches_eval_superpoint--SuperGlue-outdoor_viz/v_dogman/drawmatch-Homography-A2B-idx-000004.png,pics/supp/visual/HPatches_eval_superpoint--SGM-ICCV2021_viz/v_dogman/drawmatch-Homography-A2B-idx-000004.png,pics/supp/visual/HPatches_eval_superpoint--FullLoFTR-Block-8-SE-2-FixedSeeds_viz/v_dogman/drawmatch-Homography-A2B-idx-000004.png,pics/supp/visual/HPatches_eval_superpoint--NONE_viz/v_dogman/drawmatch-Homography-A2B-idx-000004.png,view.,dogman
0.7680890538033395,0.7249544626593807,0.8314176245210728,0,view.,dogman
SuperGlue,SGMNet,Ours,Maskedman (Original),view.,maskedman
pics/supp/visual/HPatches_eval_superpoint--SuperGlue-outdoor_viz/v_maskedman/drawmatch-Homography-A2B-idx-000006.png,pics/supp/visual/HPatches_eval_superpoint--SGM-ICCV2021_viz/v_maskedman/drawmatch-Homography-A2B-idx-000006.png,pics/supp/visual/HPatches_eval_superpoint--FullLoFTR-Block-8-SE-2-FixedSeeds_viz/v_maskedman/drawmatch-Homography-A2B-idx-000006.png,pics/supp/visual/HPatches_eval_superpoint--NONE_viz/v_maskedman/drawmatch-Homography-A2B-idx-000006.png,view.,maskedman
0.7226666666666667,0.6692307692307692,0.8050847457627118,0,view.,maskedman
SuperGlue,SGMNet,Ours,Machines (Original),view.,machines
pics/supp/visual/HPatches_eval_superpoint--SuperGlue-outdoor_viz/v_machines/drawmatch-Homography-A2B-idx-000003.png,pics/supp/visual/HPatches_eval_superpoint--SGM-ICCV2021_viz/v_machines/drawmatch-Homography-A2B-idx-000003.png,pics/supp/visual/HPatches_eval_superpoint--FullLoFTR-Block-8-SE-2-FixedSeeds_viz/v_machines/drawmatch-Homography-A2B-idx-000003.png,pics/supp/visual/HPatches_eval_superpoint--NONE_viz/v_machines/drawmatch-Homography-A2B-idx-000003.png,view.,machines
0.47660311958405543,0.45585738539898135,0.5261904761904762,0,view.,machines
SuperGlue,SGMNet,Ours,Yuri (Original),view.,yuri
pics/supp/visual/HPatches_eval_superpoint--SuperGlue-outdoor_viz/v_yuri/drawmatch-Homography-A2B-idx-000004.png,pics/supp/visual/HPatches_eval_superpoint--SGM-ICCV2021_viz/v_yuri/drawmatch-Homography-A2B-idx-000004.png,pics/supp/visual/HPatches_eval_superpoint--FullLoFTR-Block-8-SE-2-FixedSeeds_viz/v_yuri/drawmatch-Homography-A2B-idx-000004.png,pics/supp/visual/HPatches_eval_superpoint--NONE_viz/v_yuri/drawmatch-Homography-A2B-idx-000004.png,view.,yuri
0.47884187082405344,0.46320346320346323,0.5121107266435986,0,view.,yuri
\end{filecontents*}

\title{Efficient Linear Attention for Fast and Accurate Keypoint Matching}


\author{Suwichaya Suwanwimolkul} 
\affiliation{%
	\institution{KDDI Research, Inc.} 
	\country{Japan} 
}
\email{xsu-suwanui@kddi.com}

\author{Satoshi Komorita}  
\affiliation{%
	\institution{KDDI Research, Inc.} 
	\country{Japan} 
} 
\email{sa-komorita@kddi.com}

\renewcommand{\shortauthors}{Suwanwimolkul, et al.}

\begin{abstract} 
 Recently Transformers have provided state-of-the-art performance in \textit{sparse} matching, crucial to realize high-performance 3D vision applications. Yet, these Transformers lack efficiency due to the quadratic  computational complexity of their attention mechanism. To solve this problem, we employ an efficient linear attention for the linear computational complexity. Then, we propose a new attentional aggregation that achieves high accuracy by aggregating both the global and local information from sparse keypoints. To further improve the efficiency, we propose the joint learning of feature matching and description. Our learning enables simpler and faster matching than Sinkhorn, often used in matching the learned descriptors from Transformers. Our method achieves competitive performance with only $0.84M$ learnable parameters against the bigger SOTAs,  SuperGlue ($12M$ parameters) and SGMNet ($30M$  parameters), on three benchmarks, HPatch, ETH, and Aachen Day-Night. 
\end{abstract}



%
\begin{teaserfigure}   
	{  	\fontsize{8.20}{8}\selectfont 
		\setlength\tabcolsep{2pt}  	\renewcommand{\arraystretch}{1.2}  
			\begin{tabular}{  P{2cm}  c@{\hskip1pt}c@{\hskip1pt}  c@{\hskip1pt}   p{1.8cm} }    
			  \smash{\raisebox{5pt}{\multirow{4}{1.6cm}{\centering\includegraphics [width=0.09\textwidth, trim={0.2cm 0.0cm 0.0cm 0.0cm},clip]{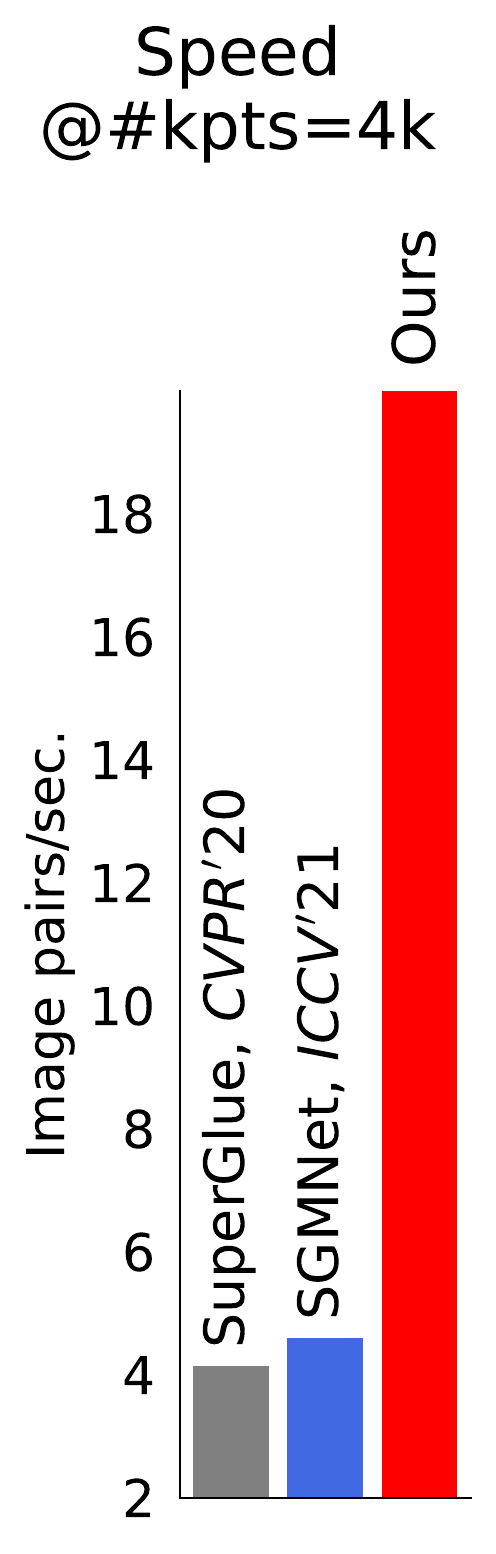}} }}   & SuperGlue, \textit{CVPR'20}  & SGMNet, \textit{ICCV'21} & Ours  &  \\
				  & 
				 	\begin{minipage}[b]{0.26\textwidth} 
							\includegraphics [width=1\textwidth, trim={0.0 0 0.0cm 0.0cm},clip]{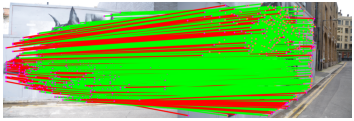}
						\end{minipage} 
					&
					\begin{minipage}[b]{0.26\textwidth} 
							\centering
							\includegraphics [width=1\textwidth, trim={0.0 0 0.0cm 0.0cm},clip]{{pics/supp/visual/HPatches_eval_superpoint--SGM-ICCV2021_viz/v_dogman/drawmatch-Homography-A2B-idx-000004}.png} 
						\end{minipage}
					&
					\begin{minipage}[b]{0.26\textwidth} 
							\includegraphics [width=1.0\textwidth, trim={0.0cm 0.0cm 0.0cm 0.0cm},clip]{{pics/supp/visual/HPatches_eval_superpoint--FullLoFTR-Block-8-SE-2-FixedSeeds_viz/v_dogman/drawmatch-Homography-A2B-idx-000004}.png}
						\end{minipage}  &   \smash{\raisebox{30pt}{ \begin{minipage}{1.25cm}  \centering \textbf{\textit{Image  Matching}}  \end{minipage} }}   \\
				  &  
					\begin{minipage}[b]{0.20\textwidth} 
				\includegraphics [width=1\textwidth, trim={4.5cm 2cm 3.75cm 3.5cm},clip]{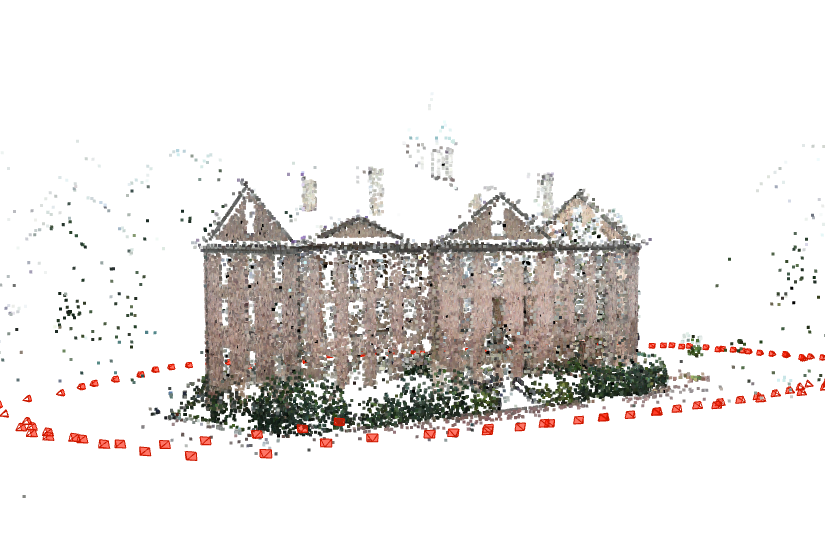} 
					\end{minipage} 
				&
				\begin{minipage}[b]{0.20\textwidth}  
				\includegraphics [width=1\textwidth, trim={4.5cm 2cm 3.75cm 3.5cm},clip]{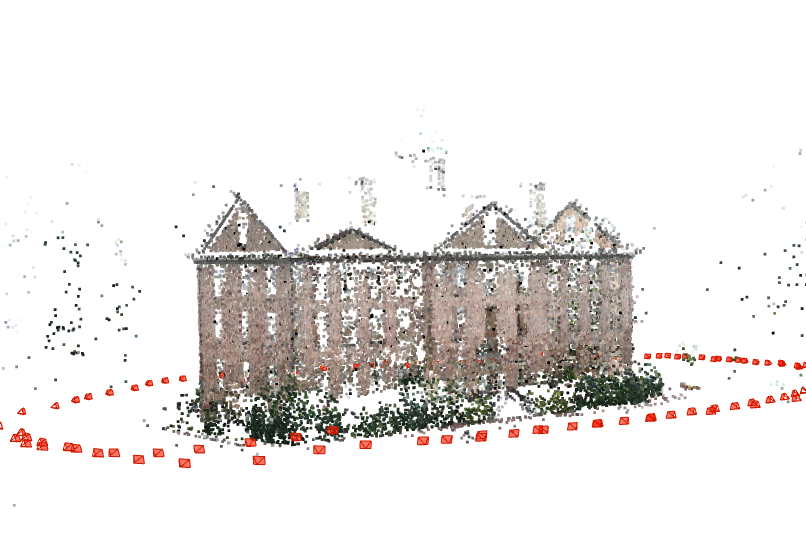} 
				\end{minipage}
				&
				\begin{minipage}[b]{0.20\textwidth} 
					\includegraphics [width=1.0\textwidth, trim={4.5cm 2cm 3.75cm 3.5cm},clip]{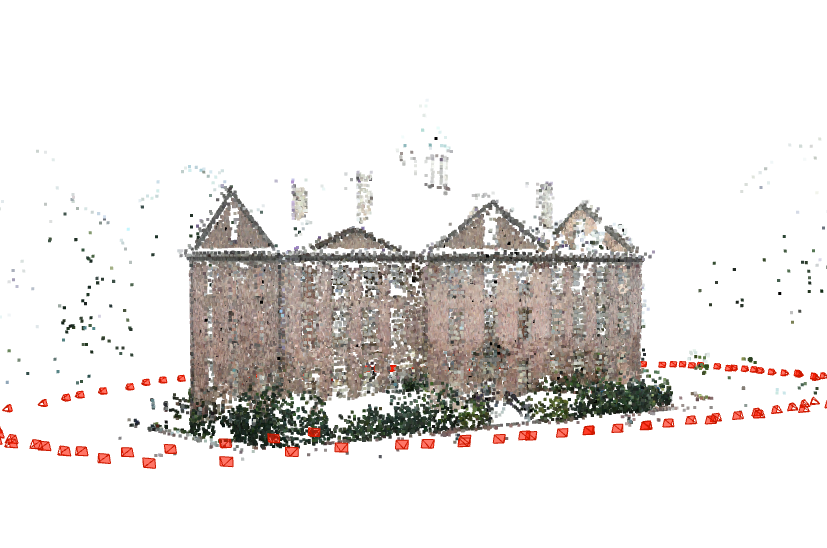} 
				\end{minipage}  & \smash{\raisebox{30pt}{  \textbf{\textit{3D Recon.}}}}  \\	
		&	 \Reproj= 0.95, \Track= 7.90 &  \Reproj= 0.98, \Track= 6.95 & \bfseries \Reproj= 0.84, \Track= 8.27  &
				\end{tabular}  
		}
\vspace*{-0.25cm}
\caption[1]{Our method versus the bigger SOTAs---SuperGlue and SGMNet---on speed (left), image matching\protect\footnotemark (top), and 3D reconstruction (bottom).} 
\label{Fig:Teaser}
\end{teaserfigure}

\maketitle   
\section{Introduction}  
\footnotetext[1]{In image matching,  green~\& red lines indicate the correct and incorrect  matches.}
Local feature matching is a fundamental step to achieve high performance in vision applications, such as visual localization~\cite{Intro:CVPR2017:Sattler},  Structure from Motion (SfM)~\cite{BM:CVPR2016:Colmap}, and 3D reconstruction~\cite{Intro:CVPR2015:RecontheWorldin6days}. Classical local feature matching starts from extracting feature descriptors and keypoints that are robust against various transformations. The local feature matching relies both on the descriptive power of the descriptors and the geometrical consistency of keypoints. The similarity of descriptors is crucial in finding the nearest neighbors in feature space. Recent studies~\cite{SuperGlue_CVPR2020, LOFTR_CVPR2021, Patch2pix_CVPR2021,  DualRC_NIPS2020, SparseNCNet_ECCV2020} focused on using deep learning techniques to boost the descriptive power of the descriptors. \textit{Transformers} have become the core technology to realize state-of-the-art  performance in \textit{sparse} matching~\cite{SuperGlue_CVPR2020,  SGMNet_ICCV2021}. Specifically, the Transformers originated from~\cite{AttentionNIPS2017} were extended to learn the descriptiveness of sparse keypoints through \textit{self-attention} and \textit{cross-attention}~\cite{SuperGlue_CVPR2020, SGMNet_ICCV2021}. Self-attention encodes the descriptiveness by aggregating information within an image; cross-attention aggregates the information between the pair.

Nevertheless, the efficiency of these Transformers~\cite{AttentionNIPS2017, SuperGlue_CVPR2020, SGMNet_ICCV2021} remains a critical issue when the number of keypoints is large. The major cause of the lack of efficiency is the quadratic computational complexity of softmax attention in these Transformers. Although Chen, \etal~\cite{SGMNet_ICCV2021} attempted to improve the complexity of~\cite{SuperGlue_CVPR2020} by using seeds to represent groups of keypoints in matching, the complexity remains quadratic in the number of seeds:  $\mathcal{O}(N^2C)$ for $N$ denoting the number of seeds (or keypoints) and  $C$ denoting feature dimensions. Nevertheless, another reason for the lack of efficiency is the descriptors matching after encoding by the Transformers. In order to match the encoded descriptors, the existing works~\cite{SuperGlue_CVPR2020,  SGMNet_ICCV2021} formulate the learning as an optimal transport problem where Sinkhorn algorithm~\cite{Villani2008OptimalTO, NIPS2013:Sinkhorn} is used to match the descriptors. The computational cost of Sinkhorn, however, is very high. In matching 10$k$ keypoints, Sinkhorn increases the runtime by an order of magnitude of the inference runtime of the Transformer~\cite{SGMNet_ICCV2021}.   

To address this problem, we resort to using the linear attention~\cite{ICML2020_LinearAttention,  LOFTR_CVPR2021} that offers linear computational complexity, \ie,  $\mathcal{O}(NC^2)$. However, it offered a lower or comparable accuracy than the regular softmax attention~\cite{ICLR2021_Rethinking}.  Thus, we further improve the accuracy of the linear attention for sparse keypoint matching by proposing a new attentional aggregation, namely  \textit{pairwise neighborhood attention}, to aggregate the local information from the neighborhoods of candidate matches in addition to the global information from the self-and cross-attention. Despite the accuracy improvement, the resulting complexity is kept low. \Table{Table:Complexity} provides the time complexity of our proposed attention versus the SOTAs. To further improve the efficiency, we propose the joint learning of the description and sparse keypoint matching based on minimizing the feature distance. With the proposed learning, we can employ the feature distance-based matching such  as~\cite{LowesThresholding}, which is simpler and faster than Sinkhorn. Then, the performance can be improved further with efficient filtering based on the feature distance~\cite{AdaLAM2020}. This results in competitive performance with a low computational cost against the existing SOTAs, as shown in \Fig{Fig:Teaser}. Our contributions are:        
 
 \smallskip 
\noindent 
{  	\fontsize{9}{11}\selectfont   
	\setlength\tabcolsep{2pt}  	\renewcommand{\arraystretch}{1.3} 
	\begin{tabular} { p{0.3cm}   p{7.70cm}}    
		$\bullet$ & \PairwiseAttenCap~to boost the performance of existing linear attention.   \\   
		$\bullet$ & Joint learning of the sparse keypoint matching and description via minimizing feature distance, which  improves the feature description and enables the efficient matching and filtering.  \\  
		$\bullet$ & Competitive performance while having only $0.84M$ learnable parameters, against the bigger SOTAs:  SuperGlue~\cite{SuperGlue_CVPR2020} ($12M$ parameters)  and SGMNet~\cite{SGMNet_ICCV2021} ($30M$ parameters) on the benchmarks: HPatch, ETH, Aachen Day-Night.  
	\end{tabular}
}   


\section{Related works}
\label{Section:Related_Works}  
\subsection{Learnable local feature matching}
\textbf{Sparse matching} has recently gained a large improvement over the local feature detection  by learning to match the detected keypoints. Notably, SuperGlue~\cite{SuperGlue_CVPR2020} employed a  Transformer similar to~\cite{AttentionNIPS2017} to exchange both visual and geometric information between the pair of images. Nevertheless, the Transformer has quadratic computational complexity in the number of keypoints.  Recently SGMNet~\cite{SGMNet_ICCV2021} achieves the lower complexity by projecting $N$ keypoints into $K$ seeds. However, SGMNet still employs the softmax attention to aggregate the messages from seeds, which still results in, yet, a quadratic complexity $\mathcal{O}(NKC + K^2C)$.  


\smallskip
\noindent
\textbf{Dense matching}~\cite{NCNet_CVPR2017, DualRC_NIPS2020, SparseNCNet_ECCV2020, LOFTR_CVPR2021,  Patch2pix_CVPR2021} aims to match descriptors in a pixel-wise manner. To enumerate all the possible matches, the works~\cite{NCNet_CVPR2017, DualRC_NIPS2020, SparseNCNet_ECCV2020} employed 4D cost volumes.  Patch2Pix~\cite{Patch2pix_CVPR2021} took a step further from SparseNCNet~\cite{SparseNCNet_ECCV2020} with an end-to-end learnable matching and refinement by regressing on pixel-level matches of local patches. Meanwhile, LoFTR~\cite{LOFTR_CVPR2021} employed a ResNet with linear  Transformer~\cite{ICML2020_LinearAttention} for detector-less matching.  Nevertheless, LoFTR matches every pixel between two images, leading to the large input's sequence length, \ie, $H_1\times W_1$ (or $H_2 \times W_2$), for $H_1 (H_2)$ and $W_1 (W_2)$ denoting the height and width of the image, which requires a much higher time and memory cost than the sparse matching.

\subsection{Graph matching} 
Graph matching aims to establish node-to-node correspondences between two or multiple graphs, which are found in various  applications~\cite{ECCV4FeatCorrGraph,CVPR2017DeepSem, PAMI2013FeatCorr,PR2015GraphFlow,CVPR2001ObjTracking}. Graph matching can be formulated as a Quadratic Assignment Problem (QAP)  known to be NP-hard~\cite{EJOR2007_NPHARD, IJCV2020SurveyImageMatching}. Early works~\cite{ICCV05Spectral, ICCV05Spectral, CVPR2017Dual, PR2019SeededGraph,2007SeedGraph2} improved the feasibility of QAP solvers. Recent works~\cite{Rolinek2020DeepGM, ICLR2020DeepMatch, CVPR2018Deep} leverage the graph matching with deep learning, yet they become less feasible in handling more than hundreds of keypoints~\cite{SuperGlue_CVPR2020}. Alternatively, the matching problem can be formulated as the optimal transport problem~\cite{Villani2008OptimalTO} where the Sinkhorn algorithm can be used to efficiently find the solution~\cite{Pacific1967:Sinkhorn, Book2019:Sinkhorn, NIPS2013:Sinkhorn}. A recent study~\cite{NIPS2013:Sinkhorn} improved the algorithm to achieve the nearly linear runtime of $\mathcal{O}(N^2/ \epsilon^3 )$, where $\epsilon$ is an error tolerance bound. However, Sinkhorn still requires an extreme runtime cost in matching  thousands of keypoints or more, as evidenced by~\cite{SuperGlue_CVPR2020,SGMNet_ICCV2021}.



 \begin{table}[t]
 	\caption{Time complexity of our proposed attention vs. SOTAs'   } 
 	\vspace{-0.2cm}
 	\begin{threeparttable} 
	\makegapedcells 
	\renewrobustcmd{\bfseries}{\fontseries{b}\selectfont} 
	{ 
		 \fontsize{6.75}{8}\selectfont  
		\setlength\tabcolsep{2pt} 
		\setlength{\extrarowheight}{3pt}
		\begin{tabular} { l@{\hskip2pt}   p{2.2cm}   p{4.2cm} }   
			\toprule   
			\bfseries Methods  & \bfseries	Comp. Complex. & \bfseries	Attention Type   \\  
			\midrule  
			SuperGlue~\cite{SuperGlue_CVPR2020}  
			&  $\mathcal{O}(N^2C)$  &  Softmax attention~\cite{AttentionNIPS2017, ICLR2018UniversalTF} \\
			SGMNet~\cite{SGMNet_ICCV2021}       
			&  $\mathcal{O}(NKC + K^2C)$  & Seeding +  Softmax attention~\cite{AttentionNIPS2017, SGMNet_ICCV2021}   \\
			Ours       
			&   $\mathcal{O}(NC'^2 + N_{max} C'^2) $ 	&  Linear Attention~\eqreff{Equation:LinAtten}~\cite{ICML2020_LinearAttention} +  Pairwise  \\
			& $\approx \mathcal{O}(NC'^2)$ &   Neighborhood~Attention~\eqreff{Equation:StructLinAtten}  \\
			\bottomrule  
		\end{tabular}
	}    
	\begin{tablenotes} 
		\item $N$   denotes the number of keypoints;  $C$ or $C'$ denotes the associated feature dimensions after linear projection;  	$K$  denotes the number of seeds in~\cite{SGMNet_ICCV2021}; $N_{max}$ denotes the size of the largest neighborhood, $N_{max} \ll N$. 
	\end{tablenotes} 
	\end{threeparttable} 
\label{Table:Complexity}
\end{table}

 
\begin{figure*}[t]    
	\vspace*{-0.2cm}
	{\fontsize{7}{7.0}\selectfont    
		\setlength\tabcolsep{5pt}  
		\renewcommand{\arraystretch}{1.5}.
		\begin{tabular}{c@{\hskip1pt}     c@{\hskip1pt}   c@{\hskip1pt}     c@{\hskip1pt}    }     
			\multicolumn{2}{c}{ 	 
			\begin{minipage}{0.55\columnwidth}   
					\includegraphics [width=\textwidth, trim={0.0 0 0.0cm 0.0cm},clip] {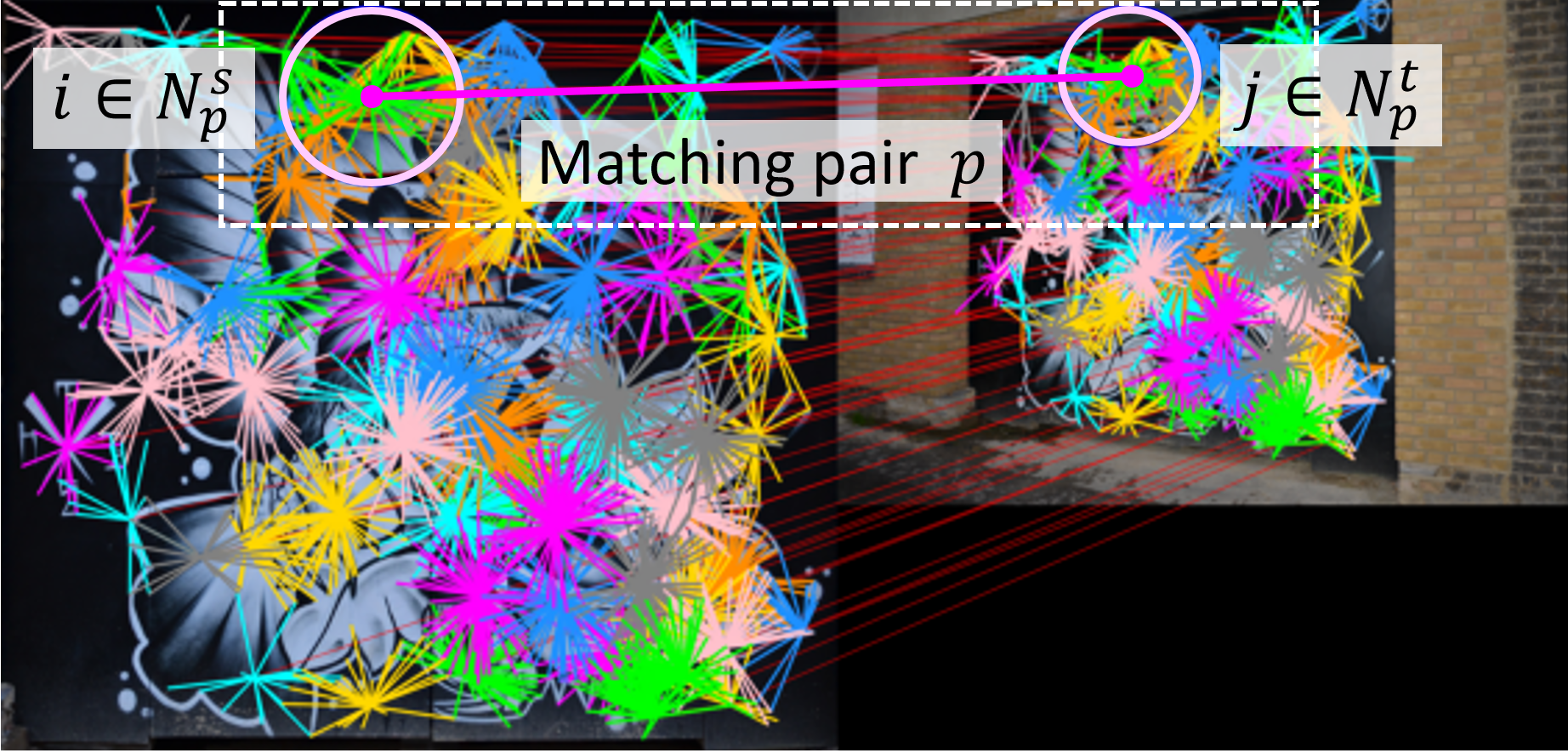}
					\subcaption{Neighborhoods \& matching pair 	\label{Fig:ExampleNeigh} }
			\end{minipage} }
			&
			\multicolumn{2}{c}{
				\begin{minipage}{0.55\columnwidth} 
					\includegraphics [width=\textwidth, trim={0.0 0 0.0cm 0.0cm},clip] {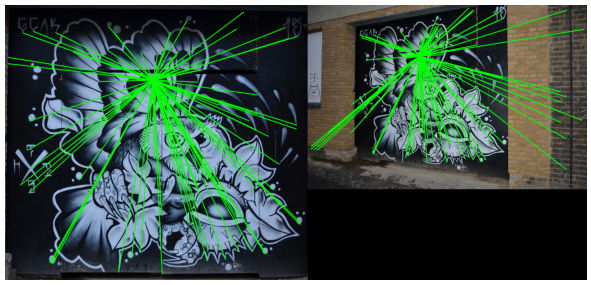} 
					\setcounter{subfigure}{2}%
					\subcaption{ Self-attention (global) }
				\end{minipage} 
			} 
			\\     
			 Source to Target ( $\vect{s} \rightarrow \vect{t}$)  &  Target to Source ($\vect{t} \rightarrow \vect{s}$) & \hspace{2pt} Source to Target ( $\vect{s} \rightarrow \vect{t}$)  & Target to Source ($\vect{t} \rightarrow \vect{s}$)  \\
			\multicolumn{2}{l}{
				\begin{minipage}{\columnwidth}  
					\includegraphics [width=0.45\textwidth, trim={0.0 0 0.0cm 0.0cm},clip]{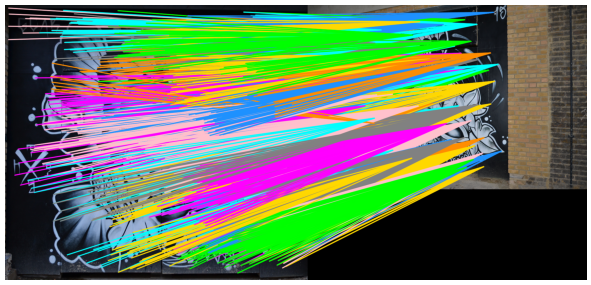}
					\includegraphics [width=0.45\textwidth, trim={0.0 0 0.0cm 0.0cm},clip]   {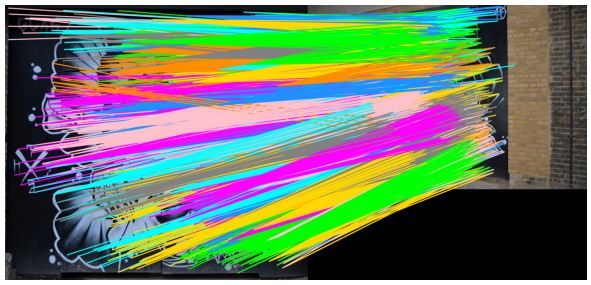}  
					\setcounter{subfigure}{1}%
					\subcaption{Our pairwise neighborhood attention (pairwise, local) \label{Fig:Pairwise}}
			\end{minipage}  }  
			&
			\multicolumn{2}{l}{ 
				\begin{minipage}{\columnwidth} 
					\includegraphics [width=0.475\textwidth, trim={0.0 0 0.0cm 0.0cm},clip] {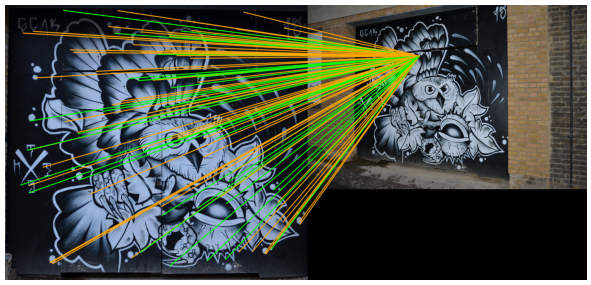}
					\includegraphics [width=0.475\textwidth, trim={0.0 0 0.0cm 0.0cm},clip] {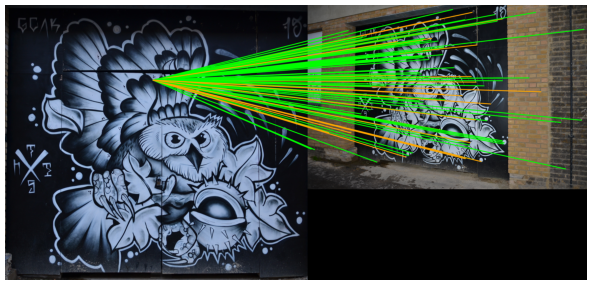} 
						\setcounter{subfigure}{3}%
					\subcaption{Cross-attention (global)  }
			\end{minipage}    } 
	\end{tabular}  	} 
	\vspace{-0.1cm}
	\caption{Visualization of our pairwise neighborhood attention vs. self- and cross-attention. Given (a) sets of neighborhoods \& matching pairs, one can aggregate information with (b) our pairwise neighborhood attention to collect the pairwise, local neighborhood information. Meanwhile, (c) self-attention and (d) cross-attention focus more on the global information within and between images.  }\label{Fig:Attention}
	\vspace{-0.3cm}
\end{figure*}

\subsection{Efficient Attention with Linear Complexity} 

Regular Transformers~\cite{AttentionNIPS2017, ICLR2018UniversalTF} contain the powerful softmax attention. However, the softmax attention has the time complexity and memory scale quadratically with the input sequence length $N$, \ie, $\mathcal{O}(N^2 \max{(D,C)})$ for $D$ and $C$ being the feature dimension of query and key. To solve this, Linear Transformers~\cite{ICML2020_LinearAttention, WACV2021_EffAttention, 2020_Linformer} reduce the computational cost to the linear complexity $\mathcal{O}(NDC)$ by computing the attention from the feature maps of dimensionality $C$, instead of the softmax attention. The feature maps offer lower or comparable accuracy than the softmax  attention in applications such as speech recognition and image generation~\cite{ICML2020_LinearAttention, ICLR2021_Rethinking}; however, it can approximate well~\textit{without imposing any constraints}, which is opposed to the previously developed techniques, \eg, restricting attention~\cite{ICML2018_ImageTransformer}, employing sparsity prior~\cite{OpenAI2019_Sparsetransformer}, pooling-based compression~\cite{Rae2020Compressive}. Others reduced the space complexity by sharing attention weights~\cite{Kitaev2020Reformer} or allowing one-time activation storage in training~\cite{OpenAI2019_Sparsetransformer}. However, these approximations are not sufficient for long-sequence problems.    
 
\smallskip
Our work is inspired by the Linear Transformers such as~\cite{WACV2021_EffAttention, 2020_Linformer,ICML2020_LinearAttention} that offer high efficiency. Meanwhile, the existing sparse matching, \ie, SuperGlue~\cite{SuperGlue_CVPR2020} and SGMNet~\cite{SGMNet_ICCV2021} employ the regular Transformer~\cite{AttentionNIPS2017, ICLR2018UniversalTF} with quadratic computational complexity.  LoFTR~\cite{LOFTR_CVPR2021} also uses Linear Transformer, but for dense matching to match every pixel, which offers the denser and more accurate matches. However, these matches are not suitable for large-scale 3D reconstruction due to the high computational cost caused by the redundant matches~\cite{ICCV2021_PixelPerfect}.   



\section{Proposed method}
\label{Section:Method}
Our main proposal is the efficient Linear Transformer for \textit{sparse matching}, where we employ two different types of attentional aggregation to collect the global and local information. Self- and cross-attention are used to aggregate the global information. Then, our proposed pairwise neighborhood  attention is used to aggregate the local information. The visualization of the two attention is in~\Fig{Fig:Attention}. The formulation of our problem is first discussed. Then, we present the proposed Transformer, where we used a local neighborhood selection to extract the local information. Then, we match the extracted features with distance-based matching and filtering in matching. Finally, we confirm our design choice with the time complexity.

\subsection{Formulation}
We consider the problem  of finding the matched pairs between $N$ and $M$ keypoints in source and target images, $\vect{I}^s$ and $\vect{I}^t$. Let $\vect{k}^s,\vect{k}^t \in \mathcal{{R}_{+}}^2$ denotes the sets of keypoint locations in the 2D images. Our goal is to encode the associated descriptors $\vect{x}^s \in \mathcal{R}^{N \times D},  \vect{x}^t \in \mathcal{R}^{M \times D}$ via a parametric function $\mathcal{F}_{\Phi} (\cdot)$ into new feature space such that it establishes the correct matching. This is formulated as  finding the best set of parameters $\Phi$ for the function $\mathcal{F}_{\Phi} (\cdot)$ via minimizing:
\begin{align*}  
	\label{Equation:Triplet}
	\mathcal{L}_{triplet} &=  \frac{1}{|\mathcal{M}_+|}\sum_{ c \in \mathcal{M}_+}{s_c \cdot \mathcal{R}( {\vect{\hat{x}}^s_c},  {\vect{\hat{x}}^t_c}   ) }  	\numberthis      
	\vspace{-1cm}
\end{align*} 
\noindent  where $\hatvect{x}^s,\hatvect{x}^t=\mathcal{F}_{\Phi} (\vect{x}^s,\vect{x}^t|\vect{k}^s,\vect{k}^t)$ and $\mathcal{M}_+$ is the set of ground truth correspondence. The subscription in $\hatvect{x}^s_{c}$ denotes the coefficient selection where $c$ denotes the selected indices.  The triplet loss $\mathcal{L}_{triplet}$ encourages  the descriptiveness of the encoded descriptors $\hatvect{x}^s,\hatvect{x}^t$ through the ranking loss $\mathcal{R}( {\hatvect{x}^s_c},  {\hatvect{x}^t_c}  )$ by minimizing the distances of matched descriptors while maximizing the unmatched ones~\cite{CVPR2020:ASLFeat}:    
\begin{align*}  
	\label{Equation:Ranking}
	\mathcal{R}( {\vect{\hat{x}}^s_c},  {\vect{\hat{x}}^t_c}  ) =&~\left[ \mathcal{D}({\hatvect{x}^s_c},  {\hatvect{x}^t_c}) - m_p\right]_{+} +  \\ &~[ m_n -  \min(\min_{k \neq c}{\mathcal{D}({\hatvect{x}^s_c},  {\hatvect{x}^t_k} )}, \min_{k \neq c }{ \mathcal{D}({\hatvect{x}^s_k},  {\hatvect{x}^t_c}) }  ) ]_{+} \numberthis   
\end{align*}
where $m_p$ and $m_n$ are small constants to prevent the negative loss value. As $\mathcal{L}_{triplet}$ decreases, $\mathcal{D}({\hatvect{x}^s_c},  {\hatvect{x}^t_c} ) =~|| {\hatvect{x}^s_c} - {\hatvect{x}^t_c} ||^{2}_2$ for $c\in \mathcal{M}_+$ will be minimized. Meanwhile, the distance of the wrong matching, \ie, ${\hatvect{x}^s_c}$ vs. ${\hatvect{x}^t_k}$ (or ${\hatvect{x}^s_k}$ vs. ${\hatvect{x}^t_c}$) for $k \notin \mathcal{M}_+$, will be further enlarged.   

Then, we weigh the distance minimization with confidence $s_c$ for $c\in \mathcal{M}_+$. The confidence $s_c$ is  a scalar product between $\hatvect{f}^s_c$ and $\hatvect{f}^t_c$, where $\hatvect{f}^s, \hatvect{f}^t$  are intermediate outputs from $\mathcal{F}_{\Phi}$, and  $\hatvect{f}^s_{c}, \hatvect{f}^t_{c}$  are column feature vectors:
\begin{align*} 
	\label{Equation:Similarity}
	s_c = {\hatvect{f}^s_c}^T {\hatvect{f}^t_c} .
	\numberthis    
\end{align*} 
\noindent The higher confidence $s_c$ will penalize the feature distance more, resulting in higher descriptiveness, and the lower feature distance can lead to the higher similarity between $\hatvect{f}^s_c$ and $\hatvect{f}^t_c$, which encourages the matching between keypoints. The proposed loss aims at minimizing the feature distance, which is different from the loss used in the existing works (SuperGlue, SGMNet, and LoFTR)  focusing on establishing as many matches as possible with their optimal transport layer, Sinkhorn. Thus, we replace Sinkhorn with feature-distance based matching and filtering (Section~\ref{Section:Matching}) for the better efficiency.

\smallskip
 We implement $\mathcal{F}_{\Phi}$ as a Linear Transformer shown in ~\Fig{Fig:Network} (Section~\ref{Section:NetworkArchitecture}) where self- and cross-attention layers collect global information with linear  attention~\cite{ICML2020_LinearAttention}.  Then, our pairwise neighborhood layers collect the local information from candidate matches. The number of candidate matches is controlled by the global information from the final cross-attention layer in~\Fig{Fig:Network}. Thus, $\hatvect{f}^s$ and $\hatvect{f}^t$  in \eqreff{Equation:Similarity}~are the output from this layer. Meanwhile, ${\hatvect{x}^s}$ and ${\hatvect{x}^t}$ are the combinations of global and local information from the final layer.


\begin{figure*}[t]   
	\vspace{-0.2cm}
	{\fontsize{8}{9.5}\selectfont    
		\setlength\tabcolsep{5.0pt}  
		\captionsetup[subfigure]{justification=centering }  
		\begin{tabular}{ c@{\hskip1pt} c@{\hskip1pt}  c@{\hskip1pt}}    
			\begin{minipage}[b]{0.24\textwidth} 
				\includegraphics [width=1\textwidth, trim={0.0 0 0.0cm 0.0cm},clip]{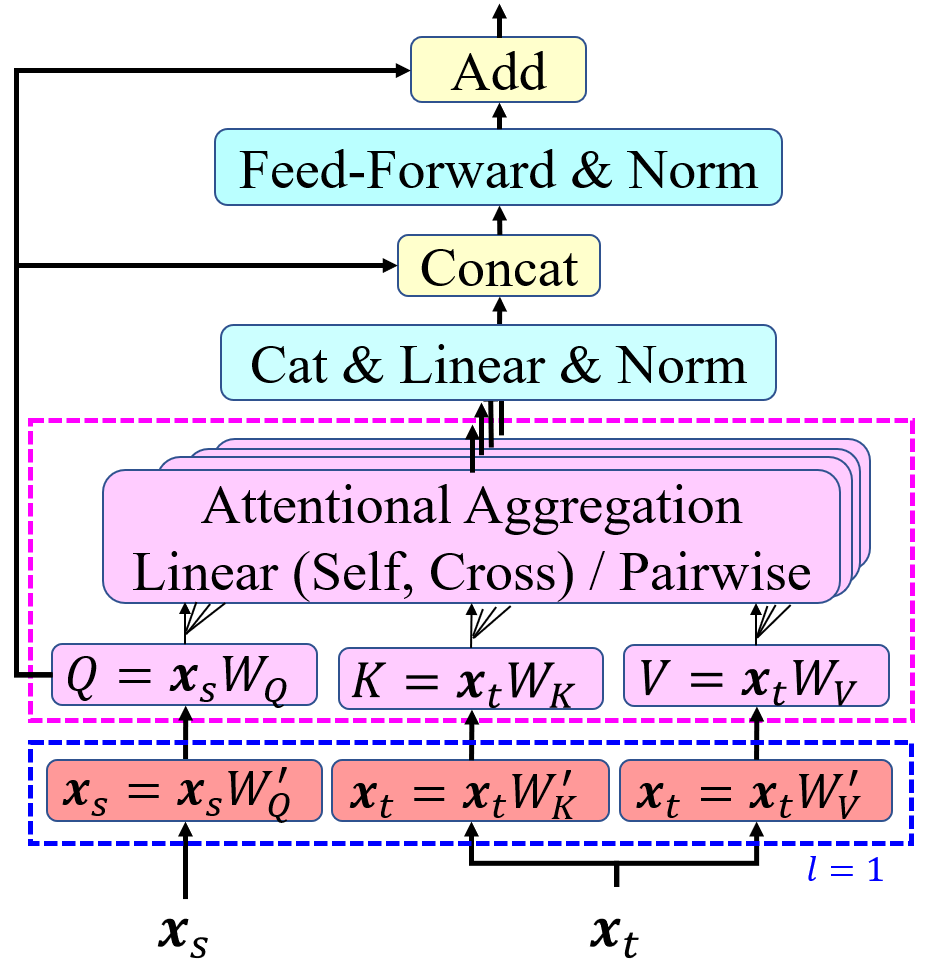}
				\subcaption{Encoder~\label{Fig:Encoder}  }
			\end{minipage} 
			&
			\begin{minipage}[b]{0.20\textwidth} 
				\centering
				\includegraphics [width=0.85\textwidth, trim={-.5cm -1.5cm 0.0cm 0.0cm},clip]{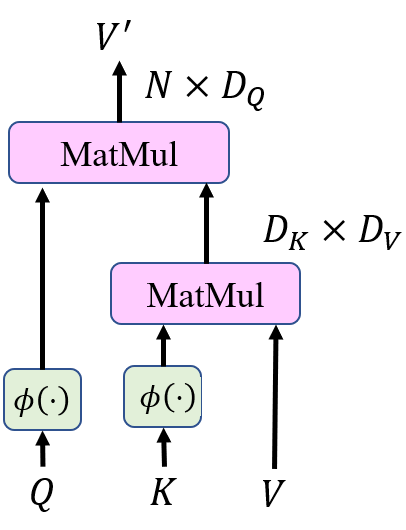} 
				\subcaption{Linear attention ~\eqreff{Equation:LinAtten}~\label{Fig:LinearAttention}  }
			\end{minipage}
			&
			\begin{minipage}[b]{0.44\textwidth} 
				\includegraphics [width=1.0\textwidth, trim={0.0cm 0.0cm 0.0cm 0.0cm},clip]{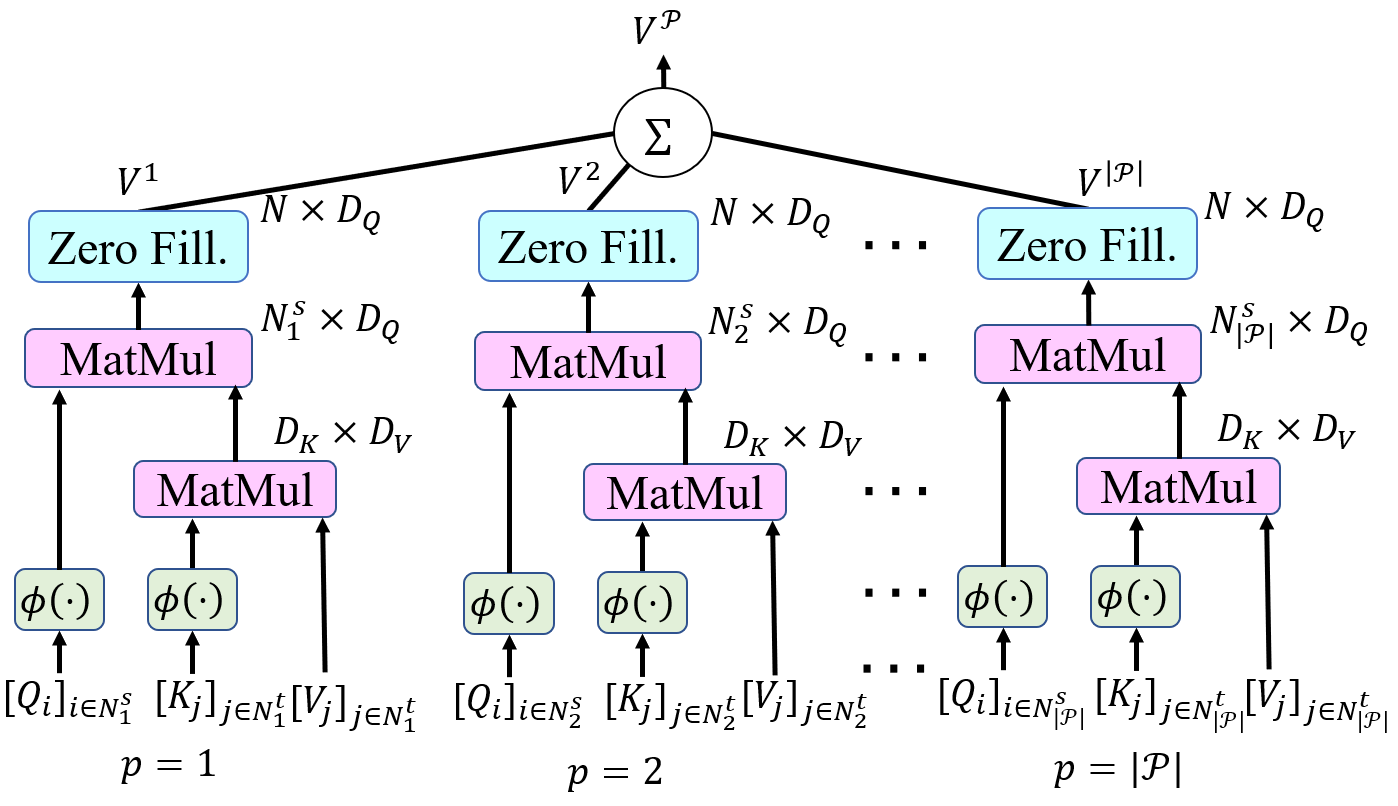}
				\subcaption{Our pairwise neighborhood attention ~\eqreff{Equation:StructLinAtten}   ~\label{Fig:StructuredAttention}  }
			\end{minipage} 
		\end{tabular}  
	}  
	\vspace{-0.2cm}
	\caption{Encoder and attentional aggregation: (a) encoder; (b) linear attention; (c) our pairwise neighborhood  attention.}\label{Fig:Encoder_and_attention}
	\vspace*{-0.2cm}
\end{figure*}

\subsection{Efficient Linear Attention}
\label{Section:LinearAttention}
 Our Transformer $\mathcal{F}_{\Phi}$ contains \textit{multiple encoders}. The function of each encoder is defined by their attention as shown in~\Fig{Fig:Encoder}. We adopt the architecture of the encoder from~\cite{LOFTR_CVPR2021}. Our Transformer consists of two types of attentional aggregation: (1) \textit{linear attention}~\cite{ICML2020_LinearAttention} and (2) \textit{our pairwise neighborhood attention}.

\smallskip
\noindent \textbf{Linear Attention.} At first, we employ the linear attention similar to~\cite{ICML2020_LinearAttention}. The architecture is provided in \Fig{Fig:LinearAttention}. The inputs of attention are vectors resulting from the linear projection of the source and target descriptors with three matrices $W_Q \in \mathcal{R}^{D\times D_Q}$, $W_K \in \mathcal{R}^{D\times D_K}$, and $W_V \in \mathcal{R}^{D\times D_V}$. Let $Q = \vect{x}_s W_Q$, $K = \vect{x}_t W_K$, $V = \vect{x}_t W_V$. Then, the output from the attention $V' = \textbf{LinAtt}(\vect{x}_s, \vect{x}_t)$, is: 
 \begin{align*} 
\label{Equation:LinAtten} 	
      V'  = [{V'}_i]_{i \in [N]}  =   \left[ \frac{ \phi(Q_i)^T \sum_{j \in [M]} \phi(K_j) V_j^T  }{\phi(Q_i)^T \sum_{j \in [M]} \phi(K_j) }  \right]_{i \in [N]}  \numberthis 
\end{align*}
where $\phi(\cdot)  = \elu{\cdot} + 1$. The subscription $i$ on a matrix returns a column vector of the $i$-th row, \eg,  $K_j$ is a vector of size $D_K \times 1$.  \\
\smallskip 
\noindent \textbf{Pairwise Neighborhood Attention.} To improve \eqreff{Equation:LinAtten}, we propose to employ the local information of the neighborhoods area about candidate matches. The architecture is provided in \Fig{Fig:StructuredAttention}. Let $\mathcal{N}^s_p$ and $\mathcal{N}^t_p$ denote a pair of keypoint neighborhood, where $\mathcal{N}^s_p$ is from the  source, and $\mathcal{N}^t_p$ from the target.  Both center around seed points $p_1$, $p_2$ of the matching pair $p =  (p_1,p_2)$. Thus, our attention incorporates the positional information of the matching neighborhood $\mathcal{N}^s_p$ and $\mathcal{N}^t_p$. The output $V^p = \StructAtt(\vect{x}_s, \vect{x}_t | \mathcal{N}^s_p, \mathcal{N}^t_p)$, is  
\begin{align*} 
	\label{Equation:StructLinAtten} 	
{V^p}=[{V^p}_i]_{i \in \mathcal{N}^s_p}  =   \left[ \frac{ \phi(Q_i)^T \sum_{j \in \mathcal{N}^t_p} \phi(K_j) V_j^T  }{\phi(Q_i)^T \sum_{j \in \mathcal{N}^t_p} \phi(K_j) }  \right]_{i \in \mathcal{N}^s_p}  \numberthis 
\end{align*}
Any element outside $\mathcal{N}^s_p$ is filled with zero value, \ie, ${V(k)^{p}} = 0$ for $k \notin \mathcal{N}^s_p$. If there is more than one pair, the output is a superposition of $V^{p}$, \ie,  $V^\mathcal{P} = \sum_{p \in \mathcal{P}} V^{p}$ where $\mathcal{P}$ is the set of matching pairs.  The set of neighboring keypoints  $\mathcal{N}^s_p$ (or $\mathcal{N}^t_p$) can be established using a local neighborhood selection (in  Section~\ref{Section:Neighborhood}). An example of the keypoint neighborhood $\mathcal{N}^s_p$ and $\mathcal{N}^t_p$ of a matching pair $p$ is provided in \Fig{Fig:ExampleNeigh}. The visualization of the attentional aggregation is provided in \Fig{Fig:Pairwise}, which results in the collection of local information in the pairwise neighborhood. Furthermore, the dominating cost of \StructAttCdot  is  $\mathcal{O}(N_{max} C^2)$ which linearly increases with the largest neighborhood size $N_{max}$.  The derivation is in Section~\ref{Section:TimeComplexity}.

\subsection{Network Architecture}
\label{Section:NetworkArchitecture} 
\begin{figure}[t]   
	\includegraphics[trim=0.20cm 0.0cm 0.01cm 0.0cm,clip,width=.975\columnwidth]{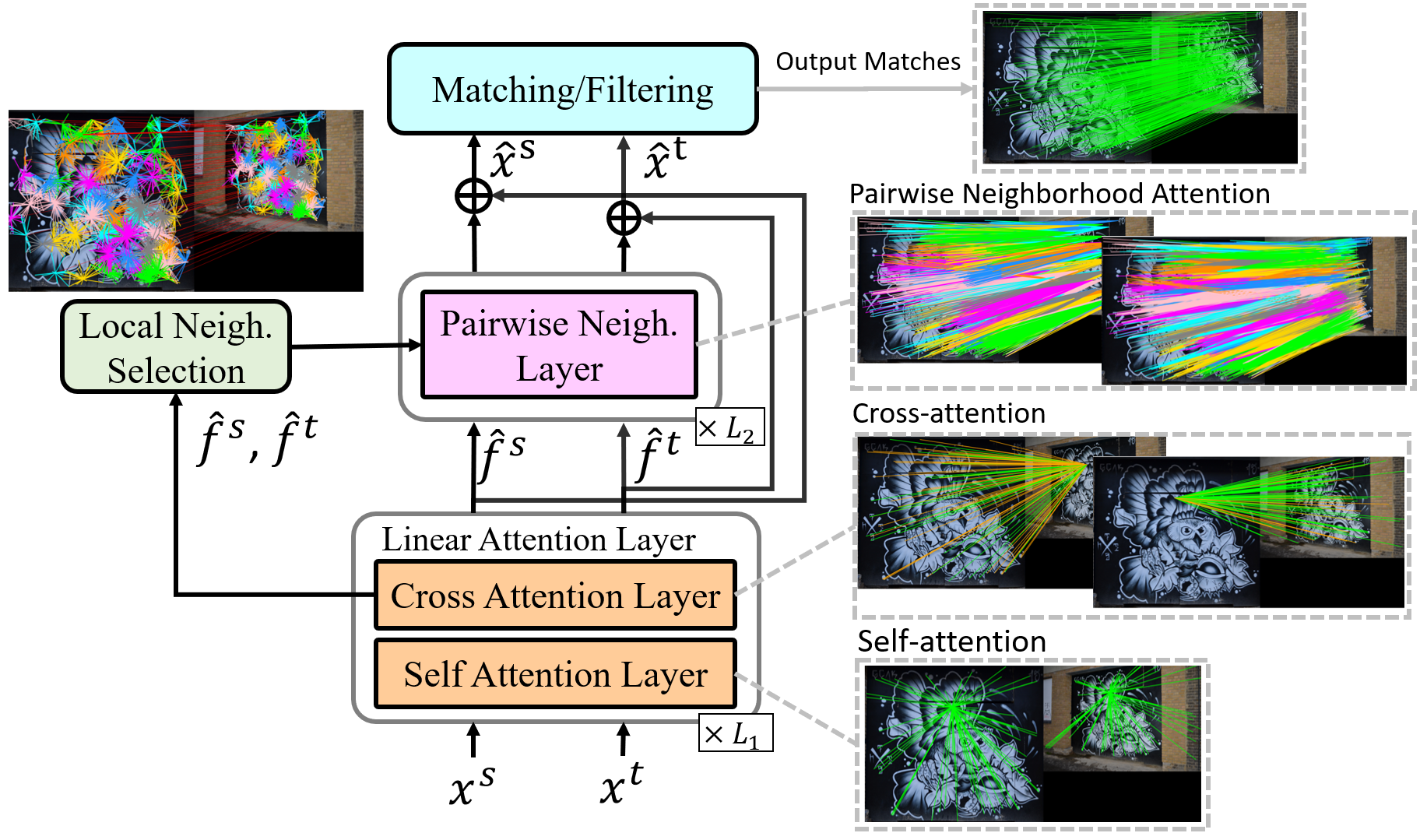}  
	\vspace{-0.25cm}
	\caption{Network Architecture} 	 
	\label{Fig:Network}
\end{figure}

Our network architecture is provided in~\Fig{Fig:Network}. Each layer consists of an encoder layer (\Fig{Fig:Encoder}) with linear or pairwise neighborhood attention, which results in \textit{\LinearAttenLayer}~and \textit{\PairwiseAttenLayer}. We use the linear attention \eqreff{Equation:LinAtten} to perform the self- and cross-attention to collect the global information through intra-and inter-relationship between descriptors. The self-attention layer updates its message by  
\begin{align*} 
	\label{Equation:Self-Attention} 	
	\hatvect{x}^s  =  \textbf{LinAtt}(\vect{x}^s, \vect{x}^s), \quad \quad  \hatvect{x}^t  =  \textbf{LinAtt}(\vect{x}^t, \vect{x}^t) \numberthis 
\end{align*}
The cross-attention layer updates messages with information collected from the inter-relationship between two descriptors~\cite{SuperGlue_CVPR2020}: 
\begin{align*} 
	\label{Equation:Cross-Attention} 	
	\hatvect{x}^s  =  \textbf{LinAtt}(\vect{x}^s, \vect{x}^t), \quad \quad  \hatvect{x}^t  =  \textbf{LinAtt}(\vect{x}^t, \vect{x}^s) \numberthis 
\end{align*}
 
Then, we employ our \PairwiseAtten~\eqreff{Equation:StructLinAtten} to form the \PairwiseAttenLayer~that aggregates the local information around candidate matches. We construct a \PairwiseAttenLayer~using~\eqreff{Equation:StructLinAtten}. Given  $(\mathcal{N}^s_p, \mathcal{N}^t_p)$ extracted by the neighborhood selection (Section~\ref{Section:Neighborhood}), the  message update is 
\begin{align*} 
	\label{Equation:Pair-Attention} 	
	{\hatvect{x}^s}^p   &=  \StructAtt(\vect{x}^s, \vect{x}^t| \mathcal{N}^s_p, \mathcal{N}^t_p ), \\ 
	 {\hatvect{x}^t}^p  &=  \StructAtt(\vect{x}^t, \vect{x}^s| \mathcal{N}^t_p, \mathcal{N}^s_p ) \numberthis 
\end{align*}
where any element outside $\mathcal{N}^s_p$ is filled  with zero value, \ie, ${\hatvect{x}^s}(k)^p = 0$ for $k \notin \mathcal{N}^s_p$  and ${\hatvect{x}^t}(k)^p = 0$ for $k \notin \mathcal{N}^s_t$. Finally, $\hatvect{x}^s = \sum_{p \in \mathcal{P}} {\hatvect{x}^s}^p $ and $\hatvect{x}^t = \sum_{p \in \mathcal{P}} {\hatvect{x}^t}^p$. Then, we perform $L_1$ loop updates between self- and cross-attention layers, and  $L_2$ loop updates over the \PairwiseAttenLayer.  Unlike~\cite{ SuperGlue_CVPR2020, LOFTR_CVPR2021, SGMNet_ICCV2021}, we did not employ any positional encoder. In addition, our first layer ($l=1$) has additional linear weights $W'_Q , W'_K$, and $W'_V$  to reduce the dimension of input descriptors into the lower dimensions $D_Q , D_K$, and $D_V$, leading to the lower computational cost in the multi-head attention of the subsequent layers~\cite{AttentionNIPS2017}. Here, we set $D_Q , D_K, D_V =C'$.

\subsection{Local Neighborhood Selection} 
\label{Section:Neighborhood}
We track the local information from candidate matches for \PairwiseAttenLayer as follows. We employ $\hatvect{f}^s$ and $\hatvect{f}^t$ from the final cross-attention layer to extract the matching pairs. Then, we construct the set of hypothesis matching seeds $\mathcal{P}$, which ensures that the seeds well spread across images. Finally, we extract the set of neighborhoods compatible with the matching seeds to construct the keypoint neighborhood, \ie, $\mathcal{N}^s_p$ and $\mathcal{N}^t_p$, for $p \in \mathcal{P}$, for Eq.~\eqref{Equation:StructLinAtten}.  

\smallskip 
\noindent\textbf{Hypothesis Matching Seeds Selection.}  We start from establishing the set of seed points with high matching confidence and well spread around the image.  Let $\mathcal{M}$ denotes  a set containing the matching pair extracted by the distance ratio algorithm $\text{Dist} (\cdot | \theta)$~\cite{LowesThresholding}  where $\theta$ is an appropriate threshold. Let $\text{distratio}(i,j)$ denotes the distance ratio value corresponding to the match $(i,j)$. Then, the set of matching seeds is defined as follows:   
\begin{align*}  
	\label{Equation:HypothesisSeeds}
	\mathcal{P} :=  \left\lbrace   (i,j) |   \Hquad \text{distratio}(i,j) >    \text{distratio}(i,k), \right. \nonumber \\
	\left. \quad \quad \quad ~\text{for}~ k \in \text{Nei}(j | R), \forall~  (i, j) \in \mathcal{M}  \right\rbrace   	\numberthis  
\end{align*} 
where $\text{Nei}(\cdot| R)$ denotes the index set of neighboring keypoints within radius $R$. We follow~\cite{AdaLAM2020} to employ the seed separation condition where the match index $(i,j)$ is selected to the set of matching seeds  $\mathcal{P}$, if it has the highest distance ratio among its local neighbors. This is to ensure that the matching seeds are well spread.

\smallskip 
\noindent \textbf{Local Neighborhood Set Selection.}  To include candidate matches that are geometrically consistent with the matching seed $p \in \mathcal{P}$, we  collect the  points that locate in a similar neighborhood, following~\cite{AdaLAM2020, SCRAMSAC2009}. Let $(\vect{k}^s_{p}, \vect{k}^t_{p}) $ denote the location of the matched keypoints from source to target corresponding to the matching seed $p \in \mathcal{P}$.  The local neighborhood set $\mathcal{N}_{p}$ is defined as: 
\begin{align*} 
	\label{Equation:Neighborhood}
	\mathcal{N}_p :=    \left\lbrace (p_1, p_2)~~\vert~~||\vect{k}^s_{p_1} - \vect{k}^s_{p} || \leq \lambda R_s~~\& ~~|| \vect{k}^t_{p_2} - \vect{k}^t_{p} || \leq \lambda R_t,  \right. \nonumber \\
	\left.      \forall (p_1, p_2) \in \mathcal{M}  \right\rbrace   	\numberthis  
\end{align*}
where $R_s$ and $R_t$ are the radii to control the coverage of neighboring points around the matching seed $p$ in  $I_s$ and $I_t$, respectively. $\lambda$ is a hyperparameter that regulates the overlapping between  neighborhoods. Then, the pair of keypoint neighborhood $(\mathcal{N}^s_p, \mathcal{N}^t_p)$ is defined as:
\begin{align*} 
	\label{Equation:Neighbor}
	\mathcal{N}^s_p := \left\lbrace i |   i:(i,j) \in \mathcal{N}_p \right\rbrace,  \Hquad \mathcal{N}^t_p := \left\lbrace  j |   j:(i,j) \in \mathcal{N}_p
	\right\rbrace  
	\numberthis  
\end{align*} 
The  pair of keypoint neighborhood $\mathcal{N}^s_p, \mathcal{N}^t_p$ will be used to define the aggregation in ~\eqreff{Equation:StructLinAtten} to emphasize the area of candidate matches.

\subsection{Feature distance-based matching and filtering}
\label{Section:Matching}

 Given the descriptors $\hatvect{x}^s,\hatvect{x}^t$ from our Transformer, one can obtain the set of match pairs $\mathcal{M}_c$ by distance ratio thresholding such as~\cite{LowesThresholding}. However, the fixed thresholding value tends to  restrict the candidate matches overly.  Thus, we employ the similar procedure to Section~\ref{Section:Neighborhood} to include the candidate matches compatible with  $\mathcal{M}_c$:  

\smallskip 
\noindent 
{  
	\fontsize{9}{10}\selectfont   
	\setlength\tabcolsep{2pt}  	\renewcommand{\arraystretch}{1.2} 
	\begin{tabular} { p{0.5cm}   p{7.5cm}}    
	(1) & Extract hypothesis matching seeds $\mathcal{P}_c$ with ~\eqreff{Equation:HypothesisSeeds} where $\hatvect{x}_s,\hatvect{x}_t$ are used to construct the set of matching pairs $\mathcal{M}_c$.    \\   
	(2) & Extract the set of candidate matches, \ie, $ \left\lbrace \mathcal{N}_c | c \in \mathcal{P}_c  \right\rbrace$ where $\mathcal{N}_c$ is extracted with~\eqreff{Equation:Neighborhood}.  \\   
	\end{tabular}
 } 
 
\noindent\textbf{Filtering.}  We employ the filtering process of AdaLAM~\cite{AdaLAM2020} (without refitting) to improve the performance by verifying the local affine consistency in each  $\mathcal{N}_c$  with highly parallel RANSACs~\cite{ParallelRansacs1981, AdaLAM2020}. The filtering scales well with the high number of keypoints ($> 10,000$).

 
The resulting matches $ \left\lbrace \mathcal{N}_c | c \in \mathcal{P}_c  \right\rbrace$ could contain many wrong matches; however, using our network with such procedure (denoted as distance matching or \textbf{DM}) provides comparable performance to AdaLAM~\cite{AdaLAM2020} in most cases (see  Table~\ref{tab:AbationI}). The filtering process in AdaLAM (\textbf{\textit{Filt.}}) improves the performance further, yet the performance gain is more obvious with our \PairwiseAttenLayer.  It can be shown that the runtime cost of the feature distance-based matching and filtering is much lower than Sinkhorn that is used by SuperGlue and SGMNet from \Table{tab:IndividualPart}, and using linear transformer with Sinkhorn does not lead to higher matches (see Section~\ref{Section:App:Sinkhorn}).  

\subsection{Time Complexity} 
\label{Section:TimeComplexity}

 \begin{table}[t]  
	{   \caption{Time Complexity of Linear Attention } 	
		\vspace{-0.2cm}
		\label{tab:Complexity_Lin}  
		\fontsize{6.8}{7.0}\selectfont   
		\setlength\tabcolsep{1pt}  
		\setlength{\extrarowheight}{3pt}
		\begin{tabular} {  c@{\hskip2pt}  | c@{\hskip4pt} |  c@{\hskip4pt} | c@{\hskip2pt} |   l@{\hskip4pt}  }   
			\toprule
			\textit{Step}	& Operation  & Input & Output  & Complexity    \\  \midrule
			\textit{1}. Numerator	& $\sum_{j \in [M]} \phi(K_j) V_j^T $ & two $[C'\times 1]$ & $K_v = [C'\times C']$ &  $\mathcal{O}(M C'^2)$  \\  
			& $\phi(Q_i)^T K_v $   & $[C'\times 1],[C'\times C']$ &  ${Q_k}_i^T = [1\times C']$ & $\mathcal{O}(C'^2)$ \\    \hline 
			\textit{2}. Denominator &	$\sum_{j \in [M]} \phi(K_j)$ & $[C'\times 1]$  & $K_m = [C'\times 1]$ & 	$\mathcal{O}(M)$ \\
			& 	$\phi(Q_i)^T  K_m $ & two $[C'\times 1]$  & $Den_i = [1 \times 1]$ & $\mathcal{O}(C')$ \\     \hline   
			\textit{3}.  Final &		${Q_k}_i^T / Den$  &   $[1\times C'], [1 \times 1]$  & ${V'_i}^T = [1\times C']$ & $\mathcal{O}(C')$  \\    
			& 	$[V'_i]_{i \in N}$ &   $[C' \times  1]$  & $V'= [N\times C']$ & $\mathcal{O}(NC')$  \\ \bottomrule
		\end{tabular}
	}   
	{    
		\caption{Time Complexity of Pairwise Neighborhood Attention }
		\label{tab:Complexity_Pair}    
		\vspace{-0.25cm}
		\fontsize{6.8}{7.0}\selectfont   
		\setlength\tabcolsep{1pt}  
		\setlength{\extrarowheight}{3pt}
		\begin{tabular} {  c@{\hskip2pt}   | c@{\hskip4pt}  |  c@{\hskip4pt}  | c@{\hskip4pt} |  l@{\hskip4pt}  }  
			\toprule
			\textit{Step}	& Operation  & Input & Output  & Complexity    \\  \midrule 
			\textit{1}. Numerator	&	$\sum_{j \in \mathcal{N}^t_p } \phi(K_j) V_j^T $ & two $[C'\times 1]$ & $K_v = [C'\times C']$ &  $\mathcal{O}(|\mathcal{N}^t_p| C'^2)$  \\  
			&	$\phi(Q_i)^T K_v $   & $[C'\times 1],[C'\times C']$ &  ${Q_k}_i^T = [1\times C']$ & $\mathcal{O}(C'^2)$ \\   \hline 
			\textit{2}. Denominator &	$\sum_{j \in  \mathcal{N}^t_p } \phi(K_j)$ & $[C'\times 1]$  & $K_m = [C'\times 1]$ & $\mathcal{O}(|\mathcal{N}^t_p|)$ \\
			& $\phi(Q_i)^T  K_m $ & two $[C'\times 1]$  & $Den_i = [1 \times 1]$ & $\mathcal{O}(C')$ \\    \hline   
			\textit{3}.  Final &	${Q_k}_i^T / Den$  &   $[1\times C'], [1x1]$  & ${V'_i}^T = [1\times C']$ & $\mathcal{O}(C')$  \\    
			& $[V'_i]_{i \in \mathcal{N}^s_p}$ &   $[C'\times 1]$  & $V'= [|\mathcal{N}^s_p| \times C']$ & $\mathcal{O}(|\mathcal{N}^s_p| C')$  \\ \bottomrule 
		\end{tabular}
	} 	
\end{table}  
 
This section provides the time complexity of the two attentional aggregation used in our work: \textit{linear attention \eqreff{Equation:LinAtten}}~and  our \textit{pairwise neighborhood attention \eqreff{Equation:StructLinAtten}}.  Our derivation is based on the size of  $ Q,~\phi(Q) \in \mathcal{R}^{N\times C'}$ and $K,~V,~\phi(K) \in \mathcal{R}^{M\times C'}$. 
 
 \smallskip
 
\noindent \textbf{Linear Attention.} The complexity of \eqreff{Equation:LinAtten} is derived as in~\Table{tab:Complexity_Lin}. The analysis starts from the computations in numerator \Step{1}, denominator  \Step{2}, and final division \Step{3}. The total complexity is $\mathcal{O}(M C'^2 + C'^2 +  M + C' + C' + NC'  ) \approx \mathcal{O}(M C'^2)$. 
  
  \smallskip 
\noindent \textbf{Pairwise neighborhood attention.}  The time complexity of \eqreff{Equation:StructLinAtten} is derived similarly in~\Table{tab:Complexity_Pair}. The only difference is the range of summation operations. The total complexity is $\mathcal{O}(|\mathcal{N}^t_p| C'^2 + C'^2 +  |\mathcal{N}^t_p| + C' + C' + |\mathcal{N}^s_p|C'  )$. Let $N_{max}$ denote the size of the largest neighborhood among $\mathcal{N}^t_p$ for $p \in \mathcal{P}$, \ie, , $ N_{max} = \max_{p \in \mathcal{P}}{ |\mathcal{N}^t_p| }.$
Thus, the dominating complexity is $\approx \mathcal{O}(N_{max} C'^2)$. 
 
\smallskip 
\noindent \textbf{\underline{Total.}} Combining the two, we obtain $\mathcal{O}(M C'^2 + N_{max} C'^2) \approx \mathcal{O}(M C'^2)$ for $N_{max} \ll M$.  ~\Table{Table:Complexity} provides the comparison with SOTAs. Among these methods, our time complexity linear to $M$ (or $N$). In practice, we  set $\mathcal{N}^s_p$ (or $\mathcal{N}^t_p$) to  the same size for parallel computation.

\begin{figure*}[t]   
	{\fontsize{8}{9.5}\selectfont    
		\setlength\tabcolsep{1.0pt}  
		\captionsetup[subfigure]{justification=centering }
		\vspace{-0.05cm}
		\begin{tabular}{c@{\hskip1pt} c@{\hskip1pt}c@{\hskip1pt}  } 
			\multicolumn{3}{l}{ 
				\begin{minipage}{0.825\textwidth} 
					\includegraphics  [width=\textwidth, 
					trim={0.0cm 56.50cm 0.5cm 0.6cm},clip] {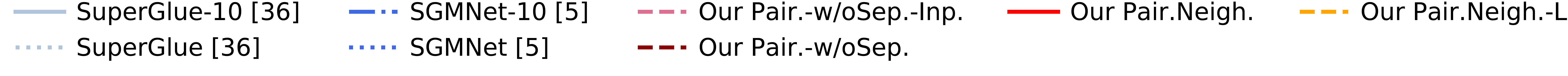}  
			\end{minipage}    }
			\\	 
			\begin{minipage}{0.65\columnwidth} 
				\includegraphics [width=0.9\textwidth, trim={0.23cm 0.0cm 0.0cm 0.0cm},clip] {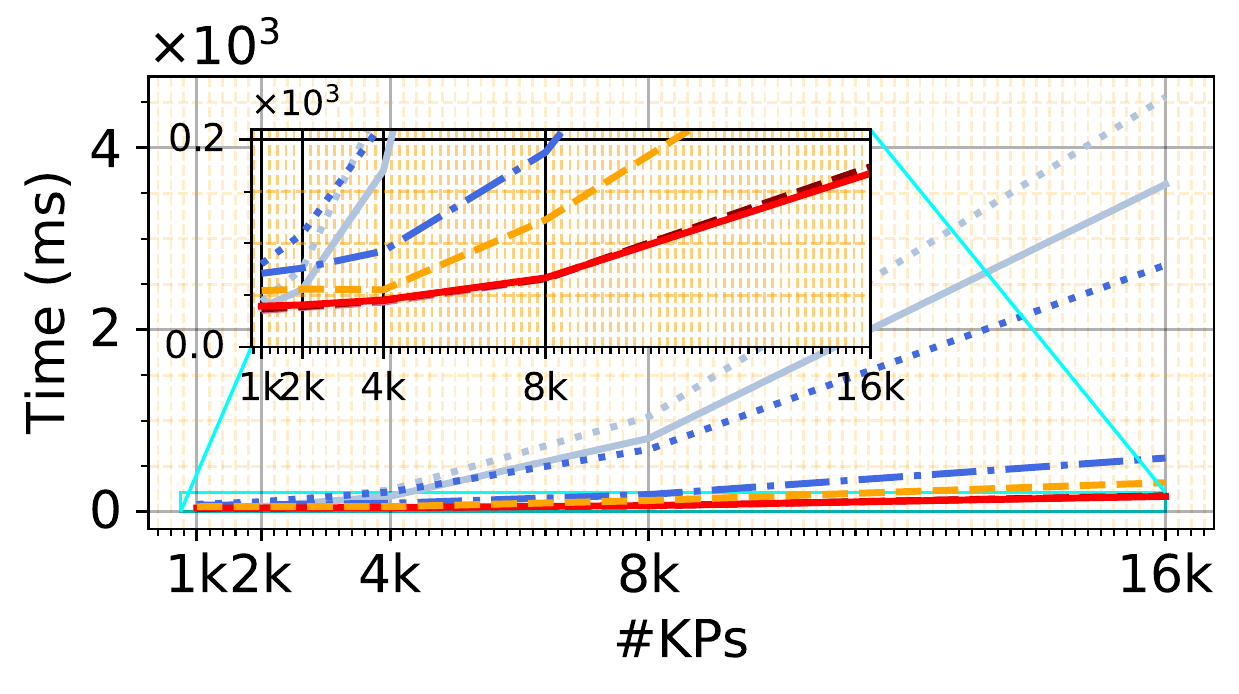} 
				\subcaption{Time Cost \label{Fig:MatchTime}}
			\end{minipage}  
			& 	\begin{minipage}{0.650\columnwidth}
				\includegraphics [width=0.9\textwidth, trim={0.25cm 0.0cm 0.0cm 0.0cm},clip] {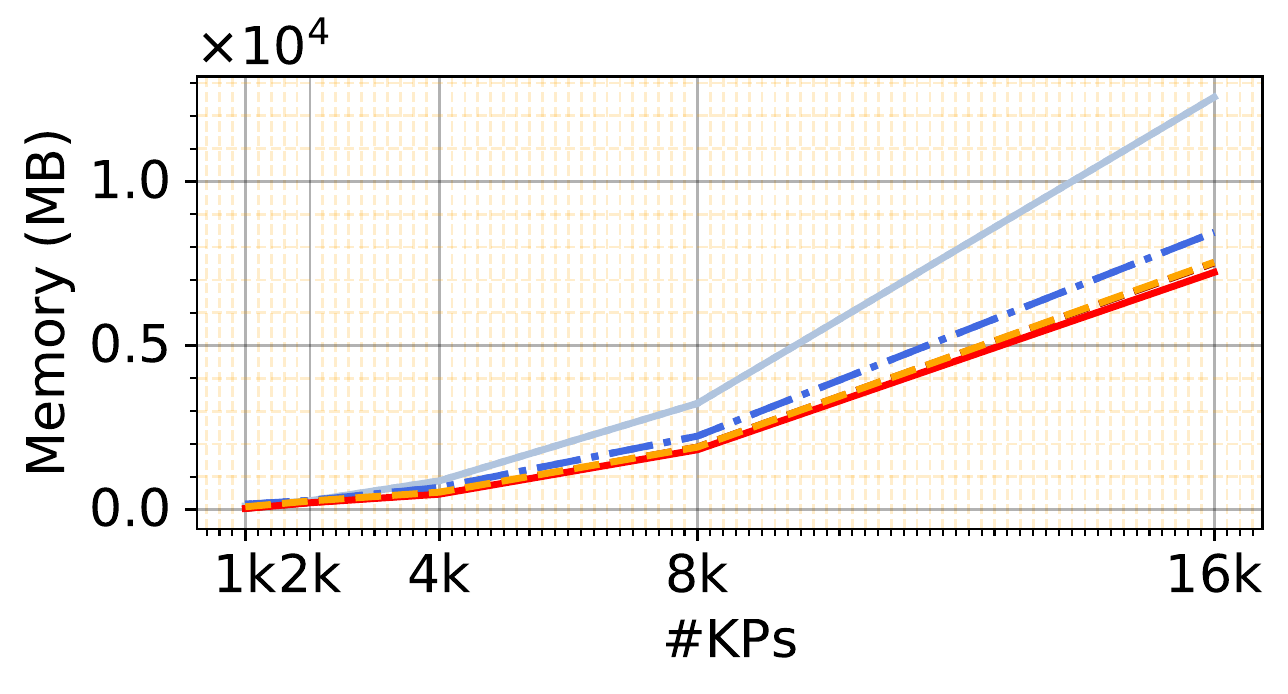} 
				\subcaption{Memory Cost \label{Fig:Memory}}
			\end{minipage}   
			& \begin{minipage}{0.60\columnwidth} 
				\includegraphics [width=0.9\textwidth, trim={0.25cm 0.0cm 0.0cm 0.0cm},clip] {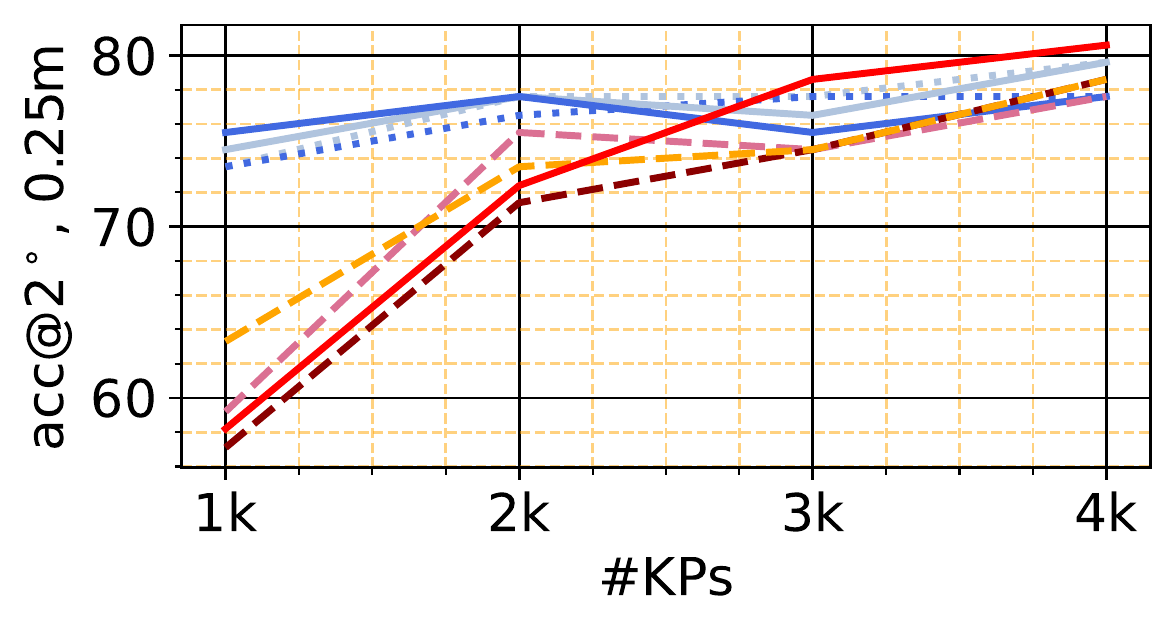}
				\subcaption{Accuracy \label{Fig:Accuracy_kp}}
			\end{minipage}  
		\end{tabular} 
	} 
	\vspace*{-0.25cm}
	\caption{Overall scalability. Our proposed method versus SOTA matching, Superglue and SGMNet  with the official  settings  and Superglue and SGMet with the faster runtime (denoted as -10 for Sinkhorn iter. = 10). }\label{Fig:Scalability}
	\vspace*{-0.1cm}
\end{figure*}

\section{Experimental Results}
\label{Section:Experiment}
We provide the ablation study and the scalability of our work against SOTAs. Then, our method is evaluated on several practical scenarios, \ie, image matching, 3D reconstruction, and visual localization. The model implementation and training are provided in  \suppl{~\ref{Section:App:Implement}}.

\smallskip 
\noindent\textbf{Comparative methods.}  Our work is compared with 
\textbf{1) Sparse matching:} SuperGlue~\cite{SuperGlue_CVPR2020} and SGMNet~\cite{SGMNet_ICCV2021}. 
\textbf{2) Dense matching:}  LoFTR~\cite{LOFTR_CVPR2021}, Patch2Pix~\cite{Patch2pix_CVPR2021}, NCNet~\cite{NCNet_CVPR2017}. 
\textbf{3) Local features:}  SuperPoint~\cite{CVPR2018:SuperPoint}, R2D2~\cite{NIPS2019:R2D2},   D2-Net~\cite{CVPR2019:D2Net}, and  ASLFeat~\cite{CVPR2020:ASLFeat}, where the standard matching, \eg,  \MNN-matching or Lowe's Thresholding, is used for matching local features. 
 \textbf{4) Keypoint filtering:} AdaLAM~\cite{AdaLAM2020} and OANet~\cite{OANet_ICCV2019}.
We report  either results from \textbf{the original papers} or derived from \textbf{the official implementations} with default settings unless otherwise specified. In each table, we highlight the \textit{top two} or \textit{top three} and \textit{underline} the best result. 
  
\smallskip 
In this paper, we apply our method to match the local features of \textbf{SuperPoint}~\cite{CVPR2018:SuperPoint} where keypoints are limited to 2k for image matching, 10k for 3D Reconstruction, and 4k for visual localization.

\subsection{Ablation Study}  
\label{Section:Experiment:Ablation}  
 This study uses localization accuracy on Aachen Day-Night~\cite{Aachen2012BMVC,Aachen2018CVPR}.  
 
\smallskip 
\noindent \textbf{Ablation Study on the Proposed Networks.}  We provide the ablation study on the impact of each component in our network ~\Fig{Fig:Network}, \ie, \LinearAttenLayer, \PairwiseAttenLayer, feature distance-based matching and filtering, and encoded feature dimensions.  

\smallskip
\noindent  From \Table{tab:AbationI}, our \ours{1} with both \LinearAttenLayer~ (\textbf{LA})  and  \PairwiseAttenLayer~(\textbf{PN}) offers the higher accuracy than \ours{5} that uses only \LinearAttenLayer,~in most cases, from 1k to 4k keypoints. Using filtering (\textbf{\textit{Filt.}}) further improves the accuracy, especially for \ours{1}. Next, we compare the model size defined by \textit{\textbf{\#dim}}. The large-size model (L) offers the higher robustness, yet our small model (S) offers the better trade-off with the computational cost. Since the goal is to achieve the high efficiency, our small model is used in subsequent comparisons against SOTAs. 

\smallskip 
\noindent \textbf{Configuration of Local Neighbor Selection.} We consider three configurations:  
$\bullet$ \textit{\ours{2}} omits the seed separation in~\eqreff{Equation:HypothesisSeeds} \& uses  $\vect{x}^s, \vect{x}^t$, instead of $\hatvect{f}^s, \hatvect{f}^t$, in~\Fig{Fig:Network} for \LocalNeighSelect. 
$\bullet$ \textit{\ours{0}} omits the seed separation  in~\eqreff{Equation:HypothesisSeeds} and uses  $\hatvect{f}^s, \hatvect{f}^t$ as input for \LocalNeighSelect.   $\bullet$ \textit{\ours{1}} follows all the steps, similar to \No{5} in \Table{tab:AbationI}.    

\smallskip
 \noindent \Table{tab:AbationII} shows that our \ours{1} and \ours{0} offer the highest accuracy when the number of keypoints  is high (>2k). Meanwhile,~\ours{2} offers higher robustness when the number of keypoints is low. Notice that all of them offer higher accuracy than using only \textit{Linear Attention} (\No{4}).   We report the results of three configurations in the next SOTAs comparison. The detailed results across all the error tolerances, \ie, (0.25m,~\ang{2}),  (0.5m,~\ang{5}), and (5m,~\ang{10}), and visualization are provided  in~\suppl{\ref{Section:App:Abblation}}.

\begin{figure*}[t]  
	\hspace{-0.75cm}  
	\begin{minipage}[b]{0.45\linewidth}
		\includegraphics[trim=10 0 0 0,clip,width=1.0\textwidth]{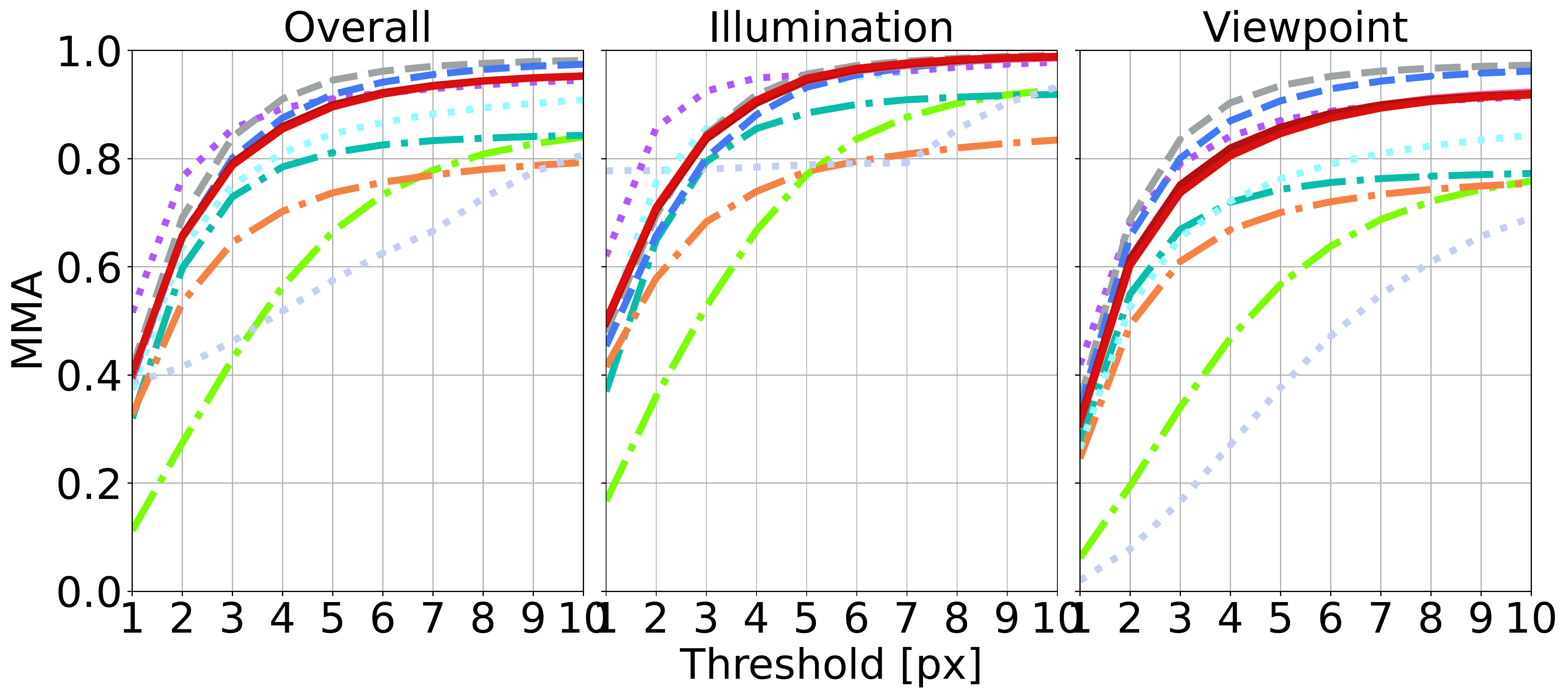}  \\
		\includegraphics[trim=30 1130 30 10,clip,width=1\textwidth]{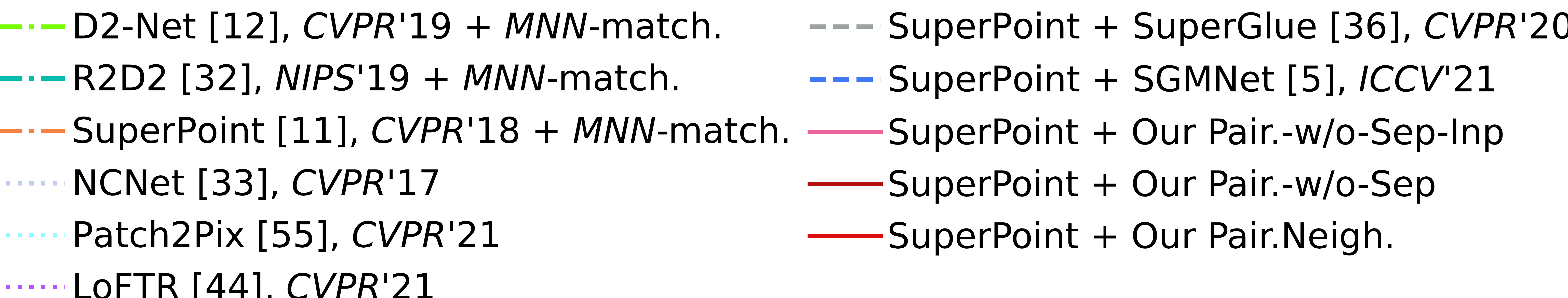}  
	\end{minipage}%
	\hspace{-0.01cm}    
	\begin{minipage}[b]{0.5\linewidth}   
		\renewrobustcmd{\bfseries}{\fontseries{b}\selectfont}   
		\sisetup{detect-weight,mode=text,round-mode=places, round-precision=2}
		{\fontsize{7.05}{8.75}\selectfont   
			{  \setlength\tabcolsep{1pt}  	\renewcommand{\arraystretch}{1.2} 
				\begin{threeparttable} 
					\begin{tabular}{  l@{\hskip2pt} P{1.0cm} P{1.3cm} P{1.00cm} P{1.1cm} P{1.7cm} }  %
							\toprule  
						\bfseries	Methods &  \bfseries	 Pub.   & \bfseries \textit{\#Matches} & \bfseries \textit{Inl.Ratio}    & \bfseries \textit{\#Param.}   & \bfseries \textit{Total Time(ms)\tnote{$\ddagger$}}   
						\DTLforeach{HPatch}{ \Meth=Meth, \Matc=Matc, \Ini=Inli, \Rtim=Rtim, \Para=Para, \SpaceVal=SpaceVal, \NoteF=NoteF, \NoteM=NoteM, \Pubs=Pub}{
							\ifthenelse{\value{DTLrowi}=1}{\tabularnewline \midrule   }{ 
								\ifthenelse{\value{DTLrowi}=4}{ \tabularnewline \cdashlinelr{1-6}}{ 
									\ifthenelse{\value{DTLrowi}=7}{ \tabularnewline \cdashlinelr{1-6}}{	
										\ifthenelse{\value{DTLrowi}=9}{ \tabularnewline \midrule}{ \tabularnewline } } } 
							} %
							\ifthenelse{\value{DTLrowi}<4}{   \Meth  \addspace{\SpaceVal}+ \MNN-matching}{
								\ifthenelse{\value{DTLrowi}<9}{ \Meth   }{\NoteF $~+$ \Ours{\NoteM}   }   }   & 
							\textit{$\Pubs$} &
							\BoldUndLineLARGK{\Matc}{1500}{4000} &   \BoldUndLineLARG{\Ini}{0.805}{0.86}  
							&\BoldUndLineSMALLESTMK{\Para}{1000000}{500000} &  \ifthenelse{\value{DTLrowi} = 11 }{  \hspace{-7pt}\BoldUndLineSMALLEST{\Rtim}{69}{35} }{ \ifthenelse{\value{DTLrowi} = 3 }{\hspace{1pt} \BoldUndLineSMALLEST{\Rtim}{69}{35}}{\BoldUndLineSMALLEST{\Rtim}{69}{35}} 
								 }    }  \\  
						\bottomrule   
					\end{tabular}
					\begin{tablenotes}
						\item[$\ddagger$] Total Time = Feature Extraction Time + Sparse/Dense Matching Time.  \vspace{-0.2cm} 
					\end{tablenotes}
				\end{threeparttable} 
			}  	  
		}
		
	\end{minipage}   
	\caption{Image matching. Our method versus SOTAs--- local features, dense matching, and sparse matching---on HPatches dataset~\cite{HPatch}. We report  \textbf{\textit{MMA}} across error thresholds (1-10 $\mathbf{\textit{px}}$), the number of matches~(\NumMatches), averaged ratios of the inliers (\InlierRat),  the number of learnable parameters (\NumParam), and \TotalTime$^{\ddagger}$.  } 
	\label{Fig:HPatch}  
	\vspace*{-0.2cm}  
\end{figure*}

\begin{table} [t]  	
	\vspace{-0.2cm} 
	\begin{threeparttable} 
		\renewrobustcmd{\bfseries}{\fontseries{b}\selectfont}   
		\sisetup{detect-weight,mode=text,round-mode=places, round-precision=2}
		{   { \caption{Impact of each component in our network  (\Fig{Fig:Network}).  
					\vspace{-0.2cm}  }	 
				\label{tab:AbationI}  
				\fontsize{6.8}{9}\selectfont   
				\setlength\tabcolsep{1pt}  	\renewcommand{\arraystretch}{1.2} 
				\begin{tabular}{c@{\hskip2pt} l c@{\hskip1pt}c@{\hskip2pt}c@{\hskip2pt}c@{\hskip2pt} c@{\hskip2pt} c@{\hskip1pt} !{\vrule width 0.5pt} p{0.6cm} p{0.5cm} p{0.50cm}  p{0.675cm} }  
					\toprule
					\bfseries  \textit{No.} & \bfseries  Methods	&\multicolumn{6}{c}{\bfseries Configurations} & \multicolumn{4}{c}{ \bfseries Accuracy @ 0.25m,~\degc{2} }\\ \cmidrule(lr){3-8}  \cmidrule(lr){9-12}
					& & \bfseries LA & \bfseries PN & \bfseries DM &  \bfseries \textit{Filt.} &  \bfseries \textit{\#dim}  &  \bfseries \textit{size}   &  \quad 1k & \hspace{2pt} 2k & \hspace{1pt} 3k & \hspace{1pt} 4k     
					\DTLforeach{Ablation1}{ \Meth=Method, \OneK=OneK, \TwoK=TwoK, \ThreeK=ThreeK, \FourK=FourK}{
						\ifthenelse{\value{DTLrowi}=1 }{\tabularnewline \midrule }{  
							\ifthenelse{\value{DTLrowi}=2}{\tabularnewline \cdashlinelr{1-12} }{  
								\ifthenelse{\value{DTLrowi}=4 \OR \value{DTLrowi}=6}{\tabularnewline \cdashlinelr{1-12} }{ \tabularnewline} } }
						\textit{\arabic{DTLrowi}} &\ifthenelse{\value{DTLrowi}=1}{AdaLAM~\cite{AdaLAM2020}}{\ifthenelse{\value{DTLrowi}=2}{\Ours{\Meth}}{\Ours{\Meth}}}   & \OursLin{\Meth} & \OursPair{\Meth} & \IsMatcher{\Meth}  & \IsFilter{\Meth}  & \IsFeat{\Meth} & \IsSize{\Meth}		& \hspace{1pt} \BoldUndLineLARGSing{\OneK}{63}{66} & \BoldUndLineLARGSing{\TwoK}{73.5}{74} & \BoldUndLineLARGSing{\ThreeK}{78}{79} &  \BoldUndLineLARGSing{\FourK}{78}{80} }  \\
					\bottomrule   
				\end{tabular} 
			}  	  
		}
		\begin{tablenotes} 
			\item \footnotesize{\textbf{LA}: \LinearAttenLayerCap,  \textbf{PN}: \PairwiseAttenCapLayer},  \textbf{DM}: Distance Matching,  \textbf{\textit{\#dim}}: Feature dimension $C'$,  \textbf{\textit{size}}: Network size  large(L) / small(S). 
		\end{tablenotes} 
	\end{threeparttable}
	\renewrobustcmd{\bfseries}{\fontseries{b}\selectfont}   
	\sisetup{detect-weight,mode=text,round-mode=places, round-precision=2}
	{	
		\caption{Configurations of local neighborhood selection. }	 
		\label{tab:AbationII}   
		\vspace{-0.2cm}
		\fontsize{7.0}{9}\selectfont   
		{  \setlength\tabcolsep{2pt}  	\renewcommand{\arraystretch}{1.2} 
			\begin{tabular}{l@{\hskip2pt} c@{\hskip2pt}  c@{\hskip1pt} !{\vrule width 0.5pt} p{0.65cm} p{0.60cm} p{0.50cm}  p{0.675cm} }  %
				\toprule  
				\bfseries Methods  &  	\multicolumn{2}{c}{\bfseries Config. of Local Neigh. Selection} &  \multicolumn{4}{c}{\bfseries Accuracy @ 0.25m,~\degc{2} }\\ \cmidrule(lr){2-3} \cmidrule(lr){4-7}
				&   Inputs &  Seed separation  \eqreff{Equation:HypothesisSeeds}  & \hspace{4pt} 1k & \hspace{2pt} 2k & \hspace{1pt} 3k & \hspace{1pt} 4k     
				\DTLforeach{Ablation2}{ \Meth=Method, \OneK=OneK, \TwoK=TwoK, \ThreeK=ThreeK, \FourK=FourK}{
					\ifthenelse{\value{DTLrowi}=1 }{\tabularnewline \midrule }{ 
						\ifthenelse{\value{DTLrowi}=3 }{\tabularnewline \cdashlinelr{1-7} }{ \tabularnewline}
					}
					\ours{\Meth}  &  \OursInputs{\Meth} & \OursSeps{\Meth} & \hspace{1pt} \BoldUndLineLARGSing{\OneK}{58}{59} & \BoldUndLineLARGSing{\TwoK}{72}{75} & \BoldUndLineLARGSing{\ThreeK}{74}{78} &  \BoldUndLineLARGSing{\FourK}{78}{80} }  \\
				\bottomrule
			\end{tabular} 
		}  	  
	}
	\vspace{-0.2cm}
\end{table}

\subsection{Overall Scalability}
\label{Section:Experiment:Scalability} 
We confirm the overall performance of our work on time and memory cost when running inference in \Fig{Fig:Scalability}. All the reported results are based on the official settings and run in real-time on a Titan RTX. In the official SuperGlue and SGMNet, the Sinkhorn iteration is set to 100. We also compare against SuperGlue-10 and SGMNet-10 where Sinkhorn iteration set to 10. We also report our large-size model (\Ours{1}-L), with the same settings as \No{6} in \Table{tab:AbationI}.  
 
\smallskip \noindent 
\textbf{Time Cost.} From \Fig{Fig:MatchTime}, our time cost is remarkably lower than SuperGlue and SGMNet and is \textit{linear} with the number of keypoints (\textit{\#kpt}). Specifically, at 16k keypoints,  our method is about 28 and 9 times faster than the official SuperGlue and SGMNet and is about 21 and 3 times faster than SuperGlue-10 and SGMNet-10. Our large model has higher runtime yet is much faster than the SOTAs.  

\smallskip \noindent
\textbf{Memory Cost.} In \Fig{Fig:Memory}, we measure the memory cost using the peak of memory consumption similar to~\cite{SGMNet_ICCV2021}. Our method consumes lower memory than SuperGlue and SGMNet even when  the number of keypoints is as low as 1k. When the number of keypoints $\geq$ 4k, our GPU memory cost is 50\% and 20\% lower than SuperGlue and SGMNet, respectively. Our large-size model consumes slightly higher memory, which resonates with the advantage of linear attention~\cite{ICML2020_LinearAttention}. 

\smallskip \noindent
\textbf{Accuracy vs. Keypoints.}	\Fig{Fig:Accuracy_kp} demonstrates the impact on visual localization accuracy (0.25m,~\ang{2}) as the number of keypoints increases. For our work, the impact on visual localization accuracy is more obvious as the keypoints increase. Meanwhile, SuperGlue and SGMNet only slightly improve with the number of keypoints. Our work outperforms both when the number of keypoints is $\geq$ 3k.       

\smallskip \noindent 
\textbf{Runtime of Individual Parts.} \Table{tab:IndividualPart} provides the time cost of the individual parts: (a) Transformer and (b) matching.  Our runtime increases with a much lower rate for both parts. Our large-size model (L) behaves similarly. This confirms the superior efficiency of our linear attention against the regular softmax attention of the SOTAs, as well as the faster speed of our distance-based matching and filtering over Sinkhorn used in SuperGlue and SGMNet.

\begin{table}     
	\caption{ Individual runtime: ~(a) Transformer \& (b)  Matching. }
	\vspace{-0.2cm}
	\renewrobustcmd{\bfseries}{\fontseries{b}\selectfont}   
	\sisetup{detect-weight,mode=text,round-mode=places, round-precision=2}
	\begin{threeparttable}
		{\fontsize{7.25}{9.2}\selectfont   
			{  \setlength\tabcolsep{1pt}  	\renewcommand{\arraystretch}{1.2} 
				\begin{tabular}{l@{\hskip1pt} P{0.5cm}    P{1.3cm}  P{0.35cm} P{0.4cm} P{0.4cm} P{0.55cm}   !{\vrule width 0.4pt} P{0.7cm}P{0.35cm}   P{0.4cm} P{0.4cm}  P{0.75cm}  }  %
					\toprule  	 
					\bfseries Methods &   & \multicolumn{5}{c}{\bfseries (a)~Transformer (ms) }   &   \multicolumn{5}{c}{\bfseries (b)~ Matching (ms) } \\  
					\cmidrule(rl){3-7} \cmidrule(rl){8-12} & \#Param
					& \textit{\textbf{Cplx.}} &    2k &    4k & 8k & 16k    &  Type &    2k &  4k & 8k & 16k    
					\DTLforeach{DiffTime}{\Meth=Method,\OneK=OneK,\TwoK=TwoK,\FourK=FourK,\EightK=EightK,\SixTK=SixTK,\NetOne=NetOne,\NetTwo=NetTwo,\NetFour=NetFour,\NetEight=NetEight,\NetSix=NetSix}{
						\ifthenelse{\value{DTLrowi}=1 }{\tabularnewline \midrule }{ 
							\ifthenelse{\value{DTLrowi}=3 }{\tabularnewline \cdashlinelr{1-12} }{ \tabularnewline} 	}
						\ifthenelse{\value{DTLrowi}<3 }{ \Meth    }{ \Ours{\Meth}  } & 
						\ifthenelse{\value{DTLrowi}<3 }{ \ifthenelse{\value{DTLrowi}<2}{12M}{30M}    }{ \ifthenelse{\value{DTLrowi}<4}{0.8M}{12M} } &
						\ifthenelse{\value{DTLrowi}<3 }{ \ifthenelse{\value{DTLrowi}<2}{$N^2C$}{$NKC + K^2C$}    }{ $\approx NC'^2$ } & 
						\BoldUndLineSMALLESTINT{\NetTwo}{51}{34} & 
						\BoldUndLineSMALLESTINT{\NetFour}{46}{36} & 
						\BoldUndLineSMALLESTINT{\NetEight}{97}{45}  &  
						\BoldUndLineSMALLESTINT{\NetSix}{250}{101} 	  &   \hspace{2pt}
						\ifthenelse{\value{DTLrowi}<3 }{\ifthenelse{\value{DTLrowi}<2}{Sink.}{Sink.}}{DF} & 
						\BoldUndLineSMALLESTINT{\TwoK}{7}{5.5} & 
						\BoldUndLineSMALLESTINT{\FourK}{10}{9.8} &  
						\BoldUndLineSMALLESTINT{\EightK}{27}{23}  &  
						\BoldUndLineSMALLESTINT{\SixTK}{88}{68}  }
					\\
					\bottomrule
				\end{tabular} 
			}  	  
		}
		\begin{tablenotes} 
			\item \footnotesize{Sink: Sinkhorn,~~~~DF: Distance Matching \& Filtering. 
				\item \Ours{1}-L: the setting is similar to \No{6} in Table~\ref{tab:AbationI}. }
		\end{tablenotes} 
	\end{threeparttable}
	
	\label{tab:IndividualPart}    
\end{table}

\subsection{Image Matching} 
\label{Section:Experiment:ImageMatching}
This section we compare the image matching performance between our method against the SOTA \textit{local features}, \textit{dense matching}, and \textit{sparse matching} on  HPatches~\cite{HPatch} following the protocol of~\cite{CVPR2019:D2Net}.  The additional visual results are provided in ~\suppl{\ref{Supp:Section:Matching-Visual}}.  

\smallskip
\noindent\textbf{Local Features}. In~\Fig{Fig:HPatch}, our \MMA~curves surpass all the SOTA local features D2-Net, R2D2, and SuperPoint with \MNN-matching (baseline). Notice that our \TotalTime~also includes the feature extraction runtime of SuperPoint.    

\smallskip
\noindent\textbf{Dense Matching}. In overall and viewpoint changes, our methods achieve similar \MMA~curves to LoFTR.  LoFTR also offers the highest \InlierRat~and \NumMatches. However, \TotalTime~of LoFTR  is about three times that of our method. Patch2Pix also provides high \MMA, but the \TotalTime~is about six times of ours.  

\smallskip
\noindent\textbf{Sparse Matching}. Our methods offer slightly lower \MMA~ and \NumMatches, and a similar \InlierRat~compared to SuperGlue and SGMNet. However, we can realize this performance with 1/10 \NumParam~and half \TotalTime. Our study on scalability (Section~\ref{Section:Experiment:Scalability}) reveals the efficiency improvement by our methods.


\begin{table}[t] 
	\caption{Small-size ETH. Our methods versus the official Superglue and SGMNet, Superglue-10 and SGMet-10. }
	\vspace{-0.3cm}
	\begin{threeparttable}[b]
		\renewrobustcmd{\bfseries}{\fontseries{b}\selectfont} 
		\sisetup{detect-weight,mode=text,round-mode=places, round-precision=2}
		{\fontsize{7.0}{7.0}\selectfont 
			\setlength\tabcolsep{0.1pt}  
			\setlength{\extrarowheight}{2pt} 
			\begin{tabular}{p{1.2cm}  l@{\hskip3pt}   p{0.8cm} p{0.8cm} P{0.95cm} P{0.9cm}  P{1.8cm} }   
				\toprule   \bfseries Datasets & \bfseries Methods     &  \bfseries \textit{\Track}  & \bfseries \textit{\Reproj} & \bfseries \textit{Sparse Points} &  \bfseries \textit{Dense Points} &  \bfseries \textit{Match. Time(sec)} 
				\DTLforeach{SfM:Herzjesu}{  \numreg=num_reg_images,   \numsparse=num_sparse_points_round,  \numsparseunit=num_sparse_points_unit, \tracklen=mean_track_length, \reprojerr=mean_reproj_error, \numdense=num_dense_points_round, \numdenseunit=num_dense_points_unit,\numdense=num_dense_points_round,  \numdenseunit=num_dense_points_unit,  \Mtime=MatchingTime, \MtimeV=MatchingRunTime, \note=note, \re=ref, \fnote=fnote}%
				{  	 \ifthenelse{\value{DTLrowi}=1}{\tabularnewline \midrule    }{   	\ifthenelse{\value{DTLrowi}=5}{\tabularnewline \cmidrule(lr){2-7} }{
							\ifthenelse{\value{DTLrowi}=3}{\tabularnewline \cdashlinelr{2-7} }{ \tabularnewline } 	}} 
					\ifthenelse{\value{DTLrowi}=1}
					{  \multirow{2}{11mm}{ \bfseries Herzjesu 8 images   }  & \RefC{\note}{\re}\fnote & 
						\BoldUndLineLARGDB{\tracklen}{4.52}{4.53} &   
						\BoldUndLineSMALLESTThree{\reprojerr}{0.875}{0.873}   &  
						\BoldUndLineLARGSingU{\numsparse}{8.0}{9.7}{\numsparseunit} &   
						\BoldUndLineLARGDBU{\numdense}{1.140}{1.145}{\numdenseunit}    & 
						\BoldUndLineSMALLESTKK{\MtimeV}{25}{23}   
					} 
					{  
						\ifthenelse{\value{DTLrowi} < 5}
						{&  \RefC{\note}{\re}\fnote & 	
							\BoldUndLineLARGDB{\tracklen}{4.52}{4.53} &  
							\BoldUndLineSMALLESTThree{\reprojerr}{0.885}{0.873}  &   
							\BoldUndLineLARGSingU{\numsparse}{8.0}{9.7}{\numsparseunit} &   
							\BoldUndLineLARGDBU{\numdense}{1.140}{1.145}{\numdenseunit}    &  
							\BoldUndLineSMALLESTKK{\MtimeV}{25}{23}    
						}
						{   
							& \Ours{\note}  & 	
							\BoldUndLineLARGDB{\tracklen}{4.52}{4.53} &  
							\BoldUndLineSMALLESTThree{\reprojerr}{0.885}{0.873}   &  
							\BoldUndLineLARGSingU{\numsparse}{8.0}{9.7}{\numsparseunit} &   
							\BoldUndLineLARGDBU{\numdense}{1.140}{1.145}{\numdenseunit}    & 
							\BoldUndLineSMALLESTKK{\MtimeV}{25}{23}   
						}
					}
				}       
				\DTLforeach{SfM:Fountain}{  \numreg=num_reg_images,    \numsparse=num_sparse_points_round,  \numsparseunit=num_sparse_points_unit, \tracklen=mean_track_length, \reprojerr=mean_reproj_error,\numobs=num_observations_round,  \numdense=num_dense_points_round,  \numdenseunit=num_dense_points_unit,  \numdenseunit=num_dense_points_unit, \Mtime=MatchingTime, \MtimeV=MatchingRunTime, \note=note, \re=ref, \fnote=fnote}%
				{\ifthenelse{\value{DTLrowi}=1}{\tabularnewline \toprule    }{   \ifthenelse{\value{DTLrowi}=5}{\tabularnewline \cmidrule(lr){2-7} }{
							\ifthenelse{\value{DTLrowi}=3}{\tabularnewline \cdashlinelr{2-7} }{ \tabularnewline } 	}} 
					\ifthenelse{\value{DTLrowi}=1}
					{  \multirow{2}{11mm}{   \bfseries Fountain 11 images }  &   \RefC{\note}{\re}\fnote &  
						\BoldUndLineLARGDB{\tracklen}{5.14}{5.16} &  
						\BoldUndLineSMALLESTThree{\reprojerr}{0.910}{0.904}   &  
						\BoldUndLineLARGSingU{\numsparse}{11.3}{11.5}{\numsparseunit} & 
						\BoldUndLineLARGDBU{\numdense}{1.835}{1.835}{\numdenseunit}  &  
						\BoldUndLineSMALLESTKK{\MtimeV}{42}{41}     
					} 
					{ \ifthenelse{\value{DTLrowi} < 5}{
							& \RefC{\note}{\re}\fnote  & 	
							\BoldUndLineLARGDB{\tracklen}{5.14}{5.16} &  
							\BoldUndLineSMALLESTThree{\reprojerr}{0.910}{0.904}     &  
							\BoldUndLineLARGSingU{\numsparse}{11.3}{11.5}{\numsparseunit} & 
							\BoldUndLineLARGDBU{\numdense}{1.835}{1.835}{\numdenseunit}  &  
							\BoldUndLineSMALLESTKK{\MtimeV}{42}{41}     
						}
						{&  \Ours{\note} & 	
							\BoldUndLineLARGDB{\tracklen}{5.14}{5.16} &  
							\BoldUndLineSMALLESTThree{\reprojerr}{0.910}{0.904}   &  
							\BoldUndLineLARGSingU{\numsparse}{11.3}{11.5}{\numsparseunit} & 
							\BoldUndLineLARGDBU{\numdense}{1.835}{1.835}{\numdenseunit}  &  
							\BoldUndLineSMALLESTKK{\MtimeV}{42}{41}     
					}   }
				} 
				\DTLforeach{SfM:SouthBuilding}{  \numreg=num_reg_images,    \numsparse=num_sparse_points_round,  \numsparseunit=num_sparse_points_unit, \tracklen=mean_track_length, \reprojerr=mean_reproj_error,\numobs=num_observations_round,  \numdense=num_dense_points_round, \MtimeV=MatchingRunTime,  \numdenseunit=num_dense_points_unit,  \numdenseunit=num_dense_points_unit,   \note=note, \re=ref, \fnote=fnote}%
				{\ifthenelse{\value{DTLrowi}=1}{\tabularnewline \toprule }{      	\ifthenelse{\value{DTLrowi}=5}{\tabularnewline \cmidrule(lr){2-7} }{
							\ifthenelse{\value{DTLrowi}=3}{\tabularnewline \cdashlinelr{2-7} }{ \tabularnewline } 	}} 
					\ifthenelse{\value{DTLrowi}=1}
					{  	\multirow{2}{11mm}{ \bfseries South-Building 128 images }  &   \RefC{\note}{\re}\fnote &  
						\BoldUndLineLARGDB{\tracklen}{8.2}{8.36} &  
						\BoldUndLineSMALLESTThree{\reprojerr}{0.837}{0.833}   &  
						\BoldUndLineLARGSingU{\numsparse}{114.7}{132}{\numsparseunit} & 
						\BoldUndLineLARGDBU{\numdense}{12.45}{12.52}{\numdenseunit}  &  
						\BoldUndLineSMALLESTKK{\MtimeV}{1500}{1360}
					} 
					{ \ifthenelse{\value{DTLrowi} < 5}{
							& \RefC{\note}{\re}\fnote  & 	
							\BoldUndLineLARGDB{\tracklen}{8.2}{8.3} &  
							\BoldUndLineSMALLESTThree{\reprojerr}{0.837}{0.833}   &  
							\BoldUndLineLARGSingU{\numsparse}{114.7}{132}{\numsparseunit} & 
							\BoldUndLineLARGDBU{\numdense}{12.45}{12.52}{\numdenseunit}  &  
							\BoldUndLineSMALLESTKK{\MtimeV}{1500}{1360}
						}
						{&  \Ours{\note} & 	
							\BoldUndLineLARGDB{\tracklen}{8.2}{8.3} &  
							\BoldUndLineSMALLESTThree{\reprojerr}{0.837}{0.833}    &  
							\BoldUndLineLARGSingU{\numsparse}{114.7}{132}{\numsparseunit} & 
							\BoldUndLineLARGDBU{\numdense}{12.45}{12.52}{\numdenseunit}  &  
							\BoldUndLineSMALLESTKK{\MtimeV}{1500}{1360}
					}   }
				}
				\\ 	\bottomrule
			\end{tabular}  
		}
		\begin{tablenotes}
			\item[\textdagger] Superglue with its official setting (Sinkhorn iter. = 100).  
			\item[\S] SGMNet with its official setting (Sinkhorn iter. = 100, num. seeds =128).
		\end{tablenotes} 
	\end{threeparttable}	 
	\label{Table:3DRecon-Small}	 
	\vspace{-0.5cm}
\end{table}

\subsection{3D Reconstruction}
\label{Section:Experiment:3DRecon} 
\noindent
\textbf{Evaluation.} 3D reconstruction is a keypoint-consuming application; thus, we report the matching runtime (\MatchTime) to indicate the efficiency alongside other indicators. We follow the  ETH evaluation~\cite{Eval:3DReconETH} where sparse and dense reconstruction are performed by the SfM and MVS  from  COLMAP~\cite{BM:CVPR2016:Colmap}. The dense points are from the dense reconstruction. Detailed settings and visual results are provided in ~\suppl{\ref{Supp:Section:3DReconSettings}}  and~\suppl{\ref{Supp:Section:3DRecon-Visual}}.

\smallskip 
\noindent
Our method is compared against the official SuperGlue and SGMNet and SuperGlue-10 and SGMNet-10 in ~\Table{Table:3DRecon-Small}.  Because the official implementations take too much runtime on the medium-size datasets, we compare our method against SuperGlue-10 and SGMNet-10 in \Table{Table:3DRecon-Medium}. We also report AdaLAM and \MNN+Lowe's Threshold~\cite{LowesThresholding}.   

\smallskip
\noindent 
\textbf{Results on Small-size ETH.} From~\Table{Table:3DRecon-Small}, our methods provide  the longest \Track, with lower \Reproj, and comparable \DensePoint~to SuperGlue, SuperGlue-10, SGMNet, and SGMNet-10. Our \MatchTime~is about  \textbf{\textit{10 times}} and \textbf{\textit{3 times} faster} than SuperGlue-10 and SGMNet-10, respectively. Compared with SuperGlue and SGMNet with the official settings, the efficiency gap of our work becomes larger. Our \MatchTime~is at least \textbf{\textit{20 times}} and about~\textbf{\textit{8 times} faster} than SuperGlue and SGMNet, respectively.   

\smallskip\noindent
\textbf{Results on Medium-size ETH.} From~\Table{Table:3DRecon-Medium}, our method provides the longest \Track~ and low \Reproj~in most cases. Our method offers moderate  \DensePoint~ with lower runtime than SuperGlue-10 and SGMNet-10. The baseline~\cite{LowesThresholding} provides the lowest reprojection error. However, our methods provide longer tracking length and higher \RegImg~to AdaLAM and the baseline in most cases. Our \DensePoint~is also higher than these two approaches and is comparable with SuperGlue-10 and SGMNet-10, suggesting the similar visual quality of the 3D reconstruction. Our \MatchTime~is about \textbf{\textit{3 times}} and \textbf{\textit{twice faster}} than SuperGlue-10 and SGMNet-10,  due to the lower detected keypoints by SuperPoint.  


\begin{table}[t]  
	\caption{Medium-size ETH. Our method versus \MNN+Lowe's Thresholding, AdaLAM,  Superglue-10, and SGMNet-10.}
	\label{Table:3DRecon-Medium} 
	\vspace{-0.3cm}
	\renewrobustcmd{\bfseries}{\fontseries{b}\selectfont} 
	\sisetup{detect-weight,mode=text,round-mode=places, round-precision=2}
	{\fontsize{6.8}{7.0}\selectfont 
		\setlength\tabcolsep{1pt} 
		\setlength{\extrarowheight}{2pt}
		\begin{tabular}{p{1cm}    l@{\hskip2pt}   P{0.75cm} P{0.8cm} P{0.7cm} P{0.9cm}    P{1.4cm}  }  
			\toprule   \bfseries Datasets & \bfseries Methods    &  \bfseries \textit{\Track}  & \bfseries \textit{\Reproj}   &  \bfseries \RegImg    &  \bfseries \textit{Dense Points} &  \bfseries \textit{Match. Time (H:M:S)}
			\DTLforeach{SfM:Madrid}{  \numreg=num_reg_images,   \numsparse=num_sparse_points_round,  \numsparseunit=num_sparse_points_unit, \tracklen=mean_track_length, \reprojerr=mean_reproj_error, \numdense=num_dense_points_round, \numdenseunit=num_dense_points_unit,\numdense=num_dense_points_round,  \numdenseunit=num_dense_points_unit,  \Mtime=MatchingTime, \MtimeV=MatchingRunTime,  \note=note, \re=ref, \fnote=fnote}%
			{  	 \ifthenelse{\value{DTLrowi}=1}{\tabularnewline \midrule  }{\ifthenelse{\value{DTLrowi}=5  \OR \value{DTLrowi}=2 \OR \value{DTLrowi}=3}{\tabularnewline \cdashlinelr{2-7} }{\tabularnewline}} 
				\ifthenelse{\value{DTLrowi}=1}
				{  \multirow{2}{10mm}{  \bfseries Madrid Metropolis 1344 images  }  & \RefC{\note}{\re}\fnote & 
					\BoldUndLineLARGDB{\tracklen}{8.2}{8.6} &  
					\BoldUndLineSMALLESTThree{\reprojerr}{1.1135}{1.10}   & 
					\BoldUndLineLARGINT{\numreg}{560}{750}   & 
					\BoldUndLineLARGDBU{\numdense}{3.03}{3.3}{\numdenseunit}    & 
					\BoldTextSmall{\MtimeV}{4700}{0400}{\Mtime} 
				} 
				{  
					\ifthenelse{\value{DTLrowi} < 5}
					{ &  \RefC{\note}{\re}\fnote & 	
						\BoldUndLineLARGDB{\tracklen}{8.2}{8.6} & 
						\BoldUndLineSMALLESTThree{\reprojerr}{1.1135}{1.10}   & 
						\BoldUndLineLARGINT{\numreg}{560}{750}   & 
						\BoldUndLineLARGDBU{\numdense}{3.03}{3.3}{\numdenseunit}    & 
						\BoldTextSmall{\MtimeV}{4700}{0400}{\Mtime} 
					}
					{   
						& \Ours{\note}  & 	
						\BoldUndLineLARGDB{\tracklen}{8.2}{8.6} & 
						\BoldUndLineSMALLESTThree{\reprojerr}{1.1135}{1.10}   & 
						\BoldUndLineLARGINT{\numreg}{560}{750}   & 
						\BoldUndLineLARGDBU{\numdense}{3.03}{3.3}{\numdenseunit}    & 
						\BoldTextSmall{\MtimeV}{4700}{0400}{\Mtime} 
					}
				}
			}       
			\DTLforeach{SfM:Gendark}{  \numreg=num_reg_images,    \numsparse=num_sparse_points_round,  \numsparseunit=num_sparse_points_unit, \tracklen=mean_track_length, \reprojerr=mean_reproj_error,\numobs=num_observations_round,  \numdense=num_dense_points_round,  \numdenseunit=num_dense_points_unit,  \numdenseunit=num_dense_points_unit, \Mtime=MatchingTime, \MtimeV=MatchingRunTime,
				\note=note, \re=ref, \fnote=fnote}%
			{\ifthenelse{\value{DTLrowi}=1}{\tabularnewline \toprule}{\ifthenelse{\value{DTLrowi}=5  \OR \value{DTLrowi}=2 \OR \value{DTLrowi}=3 }{\tabularnewline \cdashlinelr{2-7}  }{\tabularnewline}}  
				\ifthenelse{\value{DTLrowi}=1}
				{  	\multirow{2}{10mm}{ \bfseries Gendarmenmarkt 1463 images}  &   \RefC{\note}{\re}\fnote &  
					\BoldUndLineLARGDB{\tracklen}{7.92}{8.3} &  
					\BoldUndLineSMALLESTThree{\reprojerr}{1.117}{1.10}   &  
					\BoldUndLineLARGINT{\numreg}{1040}{1100}   & 
					\BoldUndLineLARGDBU{\numdense}{7.05}{7.2}{\numdenseunit}  &  
					\BoldTextSmall{\MtimeV}{8000}{0400}{\Mtime} 
				} 
				{ \ifthenelse{\value{DTLrowi} < 5}{
						& \RefC{\note}{\re}\fnote  & 	
						\BoldUndLineLARGDB{\tracklen}{7.92}{8.3} &  
						\BoldUndLineSMALLESTThree{\reprojerr}{1.117}{1.10}   &  
						\BoldUndLineLARGINT{\numreg}{1040}{1100}   & 
						\BoldUndLineLARGDBU{\numdense}{7.05}{7.2}{\numdenseunit}  &  
						\BoldTextSmall{\MtimeV}{8000}{0400}{\Mtime} 
					}
					{&  \Ours{\note} & 	 
						\BoldUndLineLARGDB{\tracklen}{7.92}{8.3} &  
						\BoldUndLineSMALLESTThree{\reprojerr}{1.117}{1.10}   &  
						\BoldUndLineLARGINT{\numreg}{1040}{1100}   & 
						\BoldUndLineLARGDBU{\numdense}{7.05}{7.2}{\numdenseunit}  &  
						\BoldTextSmall{\MtimeV}{8000}{0400}{\Mtime} 
				}   }
			} 
			\DTLforeach{SfM:Tower}{  \numreg=num_reg_images,    \numsparse=num_sparse_points_round,  \numsparseunit=num_sparse_points_unit, \tracklen=mean_track_length, \reprojerr=mean_reproj_error,\numobs=num_observations_round,  \numdense=num_dense_points_round, \Mtime=MatchingTime, \MtimeV=MatchingRunTime,  \numdenseunit=num_dense_points_unit,  \numdenseunit=num_dense_points_unit,    \note=note, \re=ref, \fnote=fnote}%
			{\ifthenelse{\value{DTLrowi}=1}{\tabularnewline \toprule}{\ifthenelse{\value{DTLrowi}=5  \OR \value{DTLrowi}=2 \OR \value{DTLrowi}=3}{\tabularnewline \cdashlinelr{2-7}  }{\tabularnewline}}  
				\ifthenelse{\value{DTLrowi}=1}
				{  	\multirow{2}{10mm}{ \bfseries Tower of London  1576 images}  &   \RefC{\note}{\re}\fnote &  
					\BoldUndLineLARGDB{\tracklen}{8.47}{8.6} &  
					\BoldUndLineSMALLESTThree{\reprojerr}{1.04}{1.02}   &  
					\BoldUndLineLARGINT{\numreg}{778}{930}   & 
					\BoldUndLineLARGDBU{\numdense}{5.5}{5.9}{\numdenseunit}  &  
					\BoldTextSmall{\MtimeV}{7000}{0400}{\Mtime} 
				} 
				{ \ifthenelse{\value{DTLrowi} < 5}{
						& \RefC{\note}{\re}\fnote  & 	
						\BoldUndLineLARGDB{\tracklen}{8.47}{8.6} &  
						\BoldUndLineSMALLESTThree{\reprojerr}{1.04}{1.02}   &  
						\BoldUndLineLARGINT{\numreg}{778}{930}   & 
						\BoldUndLineLARGDBU{\numdense}{5.5}{5.9}{\numdenseunit}  &  
						\BoldTextSmall{\MtimeV}{7000}{0400}{\Mtime} 
					}
					{&  \Ours{\note} & 	
						\BoldUndLineLARGDB{\tracklen}{8.47}{8.6} &  
						\BoldUndLineSMALLESTThree{\reprojerr}{1.04}{1.02}   &  
						\BoldUndLineLARGINT{\numreg}{778}{930}   & 
						\BoldUndLineLARGDBU{\numdense}{5.5}{5.9}{\numdenseunit}  &  
						\BoldTextSmall{\MtimeV}{7000}{0400}{\Mtime} 
				}   }
			}
			\\ 	\bottomrule
		\end{tabular}  
	} 	 
\end{table} 

\subsection{Visual Localization}
\label{Section:Experiment:VisualLocalization}   

\smallskip\noindent
\textbf{Evaluation.}
We employ the Aachen Day-Night~\cite{Aachen2012BMVC,Aachen2018CVPR} to demonstrate the effect on visual localization. We follow the evaluation protocols of~\textit{\href{https://www.visuallocalization.net/}{Visual Localization Benchmark}}\footnote[2]{\label{VisuallocalNet}More details about settings are in~\suppl{\ref{Supp:Section:VisualLocSettings}}.} and report the percent of successfully localized images. The  full results of our works with different configurations are provided in Table~\ref{Supp:Table:AachenDayNight}.

\begin{table}[t]
	\caption{ Visual localization on the
		Aachen Day-Night against SOTA local features, keypoint filtering, and sparse matching. }
	\label{Table:Aachen_DayNight}
	\vspace{-0.3cm}
	\makegapedcells
	\renewrobustcmd{\bfseries}{\fontseries{b}\selectfont} 
	{\fontsize{6.5}{7.2}\selectfont  
		\setlength\tabcolsep{2pt} 
		\setlength{\extrarowheight}{3pt}
		\begin{tabular} { l@{\hskip1pt} P{0.4cm} P{0.5cm} P{0.7cm} P{0.75cm} P{1.1cm} !{\vrule width 0.5pt} P{0.8cm} P{0.8cm} P{0.7cm} }   
			\toprule  
			 \bfseries Methods   & \bfseries	\textit{\#kpts}  & \bfseries  \textit{\#dim}   & \bfseries \textit{\#Param.} &  \bfseries \textit{Pub.} & \bfseries \textit{Cplx.} &   \bfseries 0.25m,~\degc{2}  & \bfseries  0.5m,~\degc{5}    & \bfseries 5m,~\degc{10}  \\  
			\midrule  
			 	D2-Net~\cite{CVPR2019:D2Net}              & 19K & 512 & 15M & \textit{CVPR'19} & -  &  74.5    &  86.7    & \bfseries \bfseries \underline{100.0}   \\ 
			  	ASLFeat~\cite{CVPR2020:ASLFeat} v.2       &   10K &   128 & \bfseries \underline{0.4M}   & \textit{CVPR'20} & -  & \bfseries \underline{81.6}    &  87.8    & \bfseries \underline{100.0}   \\
			   	R2D2~\cite{NIPS2019:R2D2} $N=8$  &  40K &   128 & 1.0M &  \textit{NIPS'19} &-  &  76.5    & \bfseries  \underline{90.8}    &  \bfseries \underline{100.0}     \\  
			   	SPTD2~\cite{IJCV21:SPTD2}   &  10K &   128 & - &  \textit{IJCV'21} &-  &  78.8    & \bfseries  89.3   &  \bfseries 99.0    \\ \hline
			 \multicolumn{5}{l}{SuperPoint~\cite{CVPR2018:SuperPoint} + SOTA Filtering/Matching}    &  & & \\   \cdashlinelr{1-9}
			   $\LshR$+ Baseline MNN & \bfseries \underline{4K} & 256 & - & \textit{CVPR'18} & - &  71.4    &  78.6    &  87.8   \\   \cdashlinelr{1-9}  
			  $\LshR$~+~OANet  ~\cite{OANet_ICCV2019}     &  \bfseries \underline{4K} & 256  &  4.8M  &  \textit{ICCV'19} &- &  77.6   & \bfseries \underline{ 90.8}  & \bfseries \underline{100.0} \\  
			  $\LshR$~+~AdaLAM  ~\cite{AdaLAM2020}   &  \bfseries \underline{4K} & 256 & -  &  \textit{ECCV'20} & - &  76.5    &  86.7    &  95.9   \\  \cdashlinelr{1-9}  
			  $\LshR$~+~SuperGlue~\cite{SuperGlue_CVPR2020}    & \bfseries \underline{4K} & 256 &  12M  &  \textit{CVPR'20} & $N^2C$ &  79.6    & \bfseries \underline{ 90.8}    &  \bfseries \underline{100.0}   \\   
			    $\LshR$~+~SGMNet~\cite{SGMNet_ICCV2021}       & \bfseries \underline{4K} & 256 &  31M  &  \textit{ICCV'21} & ${NKC\!+\!K^2C}$ &  77.6    &    88.8    & \bfseries 99.0   \\    
			\cdashlinelr{1-9} 
			   $\LshR$~+~\Ours{0}   & \bfseries \underline{4K} & \bfseries \underline{64}  & \bfseries   0.8M &  -  & $\approx NC'^2$  &   78.6 & 86.7 & 95.9      \\  
			  $\LshR$~+~\Ours{1}   & \bfseries \underline{4K} & \bfseries \underline{64}  & \bfseries  0.8M  & -  & $\approx N C'^2$ & \bfseries 80.6 & 86.7 & 95.9      \\  
			  $\LshR$~+~\Ours{1}-L  &  \bfseries \underline{4K} &   256  &   12M  & - & $\approx NC^2$ &   78.6 & 87.8 & 96.9      \\    \bottomrule
		\end{tabular}
	}  
\end{table}

\smallskip\noindent
\textbf{Results.}  From Table~\ref{Table:Aachen_DayNight}, our method  gives the competitive accuracy at (0.25m, \ang{2}): our \ours{1}\footnote[3]{Official result is provided on  \href{https://www.visuallocalization.net/details/46890/}{visuallocalization.net as EffLinAtt+Superpoint-4K}} gives the highest accuracy among  the methods that employ SuperPoint as input features, \ie,  the sparse matching (SuperGlue, SGMNet) and the keypoint filtering (OANet, AdaLAM). Meanwhile, our \ours{0} offers higher accuracy than SGMNet but lower than SuperGlue. Our performance drops as the error threshold becomes less restrictive and is  on par with AdaLAM. 
 This suggests that our method is more accurate but less robust, as our works tend to provide less matches than SuperGlue and SGMNet. Nevertheless, our methods can achieve this with a much lower \textit{\#Param.} and \textit{\#dim.} Compared to the SOTA local features, we use only 4k keypoints but give the closest performance  to ASLFeat.

\section{Summary} 		
To improve the efficiency of existing SOTA Transformers in sparse matching applications, we propose efficient attention that offers linear time complexity and high accuracy by aggregating the local and global formation. To keep the high efficiency, we proposed to train the Transformer with the joint learning of the sparse matching and description optimized based on the feature distance. This enables the use of feature distance-based matching and filtering that is simpler and faster than Sinkhorn, which results in high accuracy and extremely low runtime. Extensive experiments indicate a significant improvement in efficiency against the bigger SOTAs. 


\appendix 

\section{Parameter settings} 
\label{Section:App:Implement}
In the first layer, we set  $W'_Q , W'_K$, and $W'_V$ to linearly project from high to low dimensional space. Given that the dimensionality of SuperPoint is 256, the linear projection maps from $256\rightarrow 64$, and for any subsequent layer, $D, D_Q, D_K, D_V$ = 64.  The encoded descriptors with 64 dimensions are reshaped to $8\times8$ for multi-head attention (the number of heads = 8). For the local neighborhood selection, we set $\theta=1.0$ for Lowes'Thresholding. For \eqreff{Equation:Neighborhood}, we use $\lambda = 2$ for image matching and 3D reconstruction and $\lambda = 3$ for localization,  where $\mathcal{R}, \mathcal{R}_s, \mathcal{R}_t = \sqrt{\frac{H\times W}{100 \pi}}$.

\section{Training datasets} 
\label{Section:App:Implement:Training}
We train the proposed model with Megadepth~\cite{MegadepthCVPR2018} datasets using the same image scenes as~\cite{SuperGlue_CVPR2020}. For each epoch, we sample 100 pairs per scene and select the pair with overlapping scores in
range [0.5,1]. Given an image pair, we extract the local features using SuperPoint~\cite{CVPR2018:SuperPoint} and sample 1024 keypoints per image.
To generate the ground truth correspondence, we use the camera poses with depth maps corresponding to the two images to project the keypoints. The reprojection distances of the keypoints is used to determine ground truth matches and unmatchable points. Following~\cite{SuperGlue_CVPR2020}, a pair of keypoints are considered ground truth matches if they are mutual nearest with a reprojection distance lower than 3 pixels; otherwise, it is labeled as unmatchable. We further filter out pairs if the ground truth matches are fewer than 50. Our data generation produces around 200k training pairs in total. 

 \begin{table}[H]   
	\caption{Sinkhorn vs. distance matching and filtering on  \#matches and inlier ratio.} 
	\vspace{-0.3cm} 
	\label{Supp:Fig:HPatch2}  
	\renewrobustcmd{\bfseries}{\fontseries{b}\selectfont}   
	\sisetup{detect-weight,mode=text,round-mode=places, round-precision=2}
	{\fontsize{7.05}{8.75}\selectfont   
		{  \setlength\tabcolsep{5pt}  	\renewcommand{\arraystretch}{1.2}  
			\begin{tabular}{  l@{\hskip5pt}  P{1.3cm} P{1.00cm}   }  %
				\toprule  
				\bfseries	Methods     & \bfseries \textit{\#Matches} & \bfseries \textit{Inl.Ratio}    
				\DTLforeach{HPatchSuppSink}{ \Meth=Meth, \Matc=Matc, \Ini=Inli, \Rtim=Rtim, \Para=Para, \SpaceVal=SpaceVal, \NoteF=NoteF, \NoteM=NoteM, \Pubs=Pub, \File=File}{
					\ifthenelse{\value{DTLrowi}=1}{\tabularnewline \midrule   }{\tabularnewline}   
					\ifthenelse{\value{DTLrowi}<2}{  \hspace{0.25pt} \Meth}{
						\ifthenelse{\value{DTLrowi}<9}{ \Meth   }{  \NoteM  }   }   &  
					\BoldUndLineLARGK{\Matc}{1500}{4000} &   \BoldUndLineLARG{\Ini}{0.805}{0.86}  
				}  \\  
				\bottomrule   
			\end{tabular}  
		}  	  
	} 
\end{table} 
\begin{figure}[H]  
	\vspace{-0.70cm} 
	{\fontsize{7.05}{8.0}\selectfont 
	\begin{tabular}{c@{\hskip1pt} c@{\hskip1pt} }
		\multicolumn{2}{c}{\includegraphics[trim=0 0 0 0,clip,width=1\columnwidth]{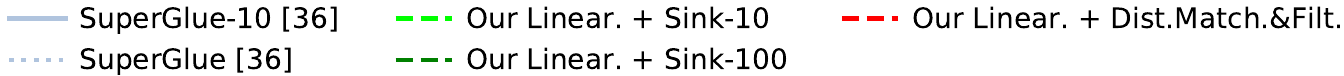}}\\	
		\includegraphics[trim=0 0 0 0,clip,width=0.45\columnwidth]{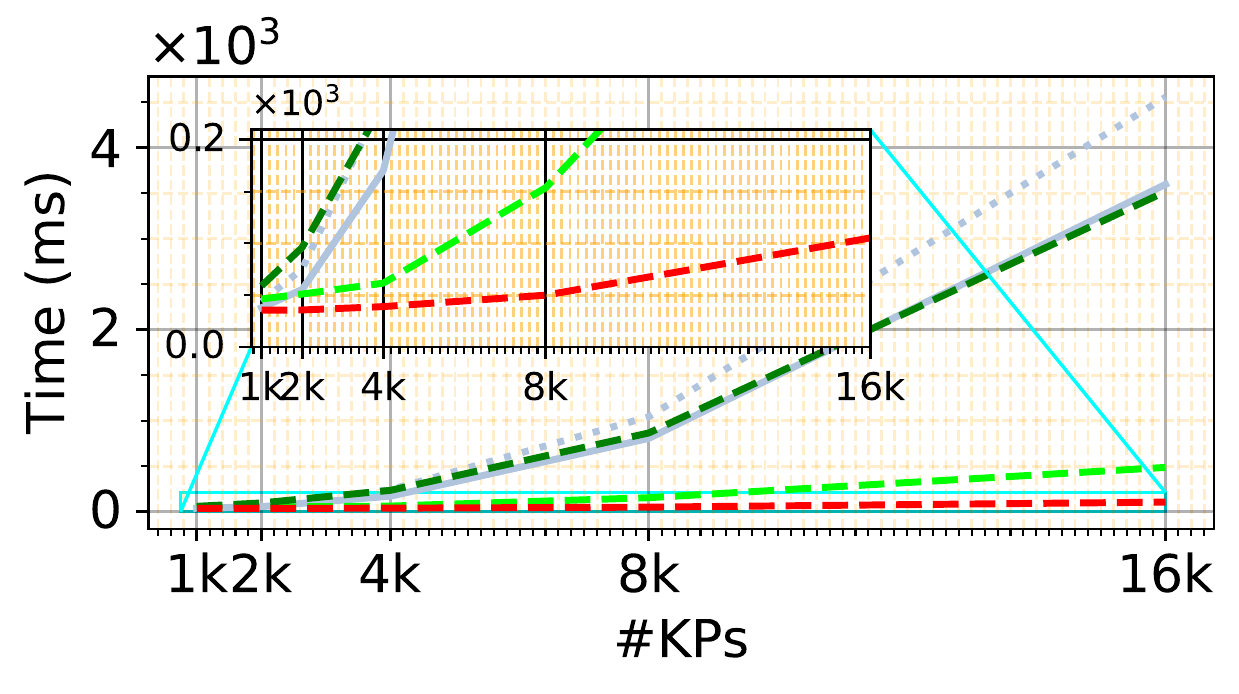} & 	
		\includegraphics[trim=0 0 0 0,clip,width=0.45\columnwidth]{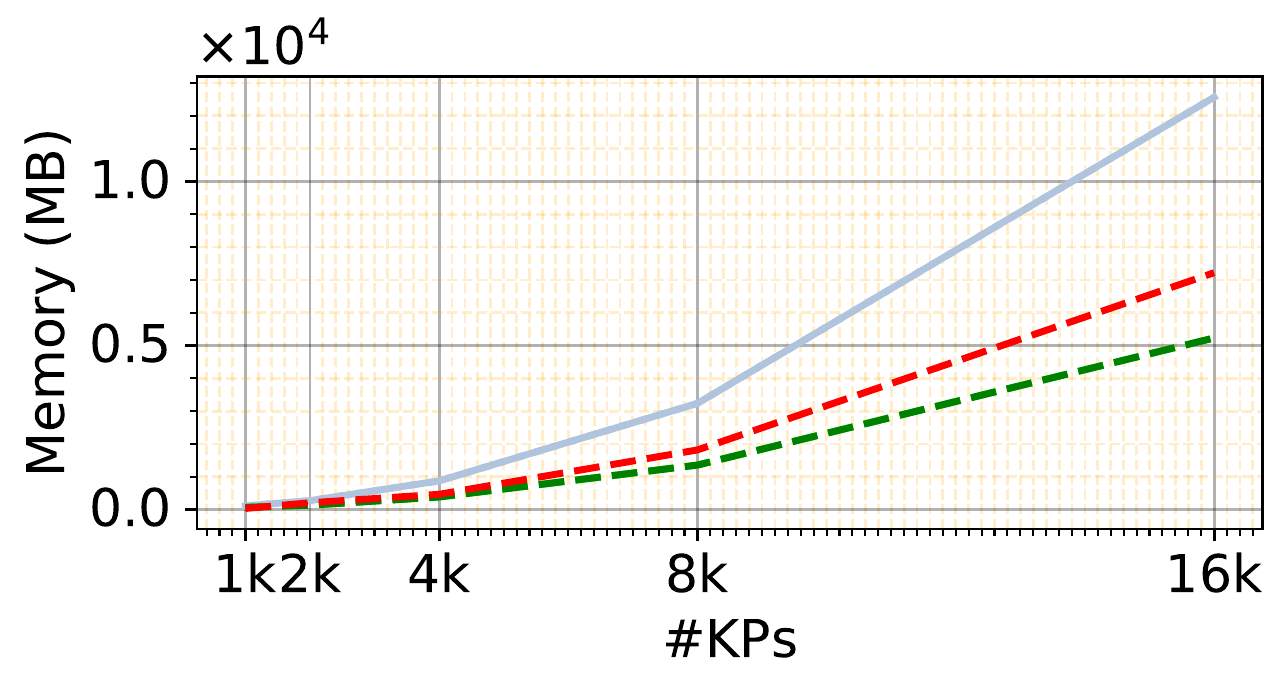} \\
		\bfseries (a) Time  & \bfseries (b) Memory 
		\vspace{-0.15cm} 
	\end{tabular}
	}
	\vspace{-0.25cm} 
	\caption{Comparison on (a) time and (b) memory cost.}	 
	\label{Supp:Fig:SinkTime}  
	\vspace{-0.25cm} 
\end{figure}

\noindent \textbf{Learning.} We use Adam optimizer with learning rate of $10^{-3}$ with exponential decay rate of 0.99992. We train for 10 epochs.

\section{Evaluation protocols \& settings}

\subsection{3D Reconstruction}
\label{Supp:Section:3DReconSettings}
Exhaustive matching that matches the global information between all possible images is used to retrieve images for the small datasets, Herzjesu and Fountain. Meanwhile, NetVLAD~\cite{NetVLAD} is used to retrieve the top 20 nearby images from South-Building, Madrid Metropolis, Gendarmenmarkt, and Tower of London. Sparse and dense reconstruction are performed by the SfM and MVS  from  COLMAP~\cite{BM:CVPR2016:Colmap}.

\subsection{Visual Localization}
\label{Supp:Section:VisualLocSettings}
According to the protocols of~\textit{Visual Localization Benchmark\footnote{\label{VisuallocalNet2} \href{https://www.visuallocalization.net/}{https://www.visuallocalization.net/} }}, we provided the costumed features and performed image registration with COLMAP~\cite{BM:CVPR2016:Colmap}; then, the localization is performed. We use the Aachen Day-Night datasets~\cite{Aachen2012BMVC,Aachen2018CVPR} whose goal is to match images with extreme day-night changes for 98 queries.

\section{Sinkhorn vs. Distance Matching \& Filtering}
\label{Section:App:Sinkhorn}
Table~\ref{Supp:Fig:HPatch2} provides the comparison between using Sinkhorn versus distance matching \& filtering with the linear transformer. Following~\cite{SuperGlue_CVPR2020}, we have trained the linear transformer with Sinkhorn with optimal transport loss (similar settings to Section~\ref{Section:App:Implement:Training}). Using Sinkhorn does not provide higher \#matches nor inlier ratios, yet Sinkhorn requires much higher time cost  in~\Fig{Supp:Fig:SinkTime}.  
 
\begin{table*} [t]  
	{		
		\fontsize{8}{9.5}\selectfont    
		\setlength\tabcolsep{5.0pt}  
		\caption{Impact of components in the proposed network (\Fig{Fig:Network}) on localization accuracy. \vspace*{-0.2cm}}	 
		\captionsetup[subfigure]{justification=centering }  
		
		\renewrobustcmd{\bfseries}{\fontseries{b}\selectfont}   
		\sisetup{detect-weight,mode=text,round-mode=places, round-precision=2}
		
		\label{Supp:tab:AbationAll}
		\begin{threeparttable}  
			
			\begin{tabular}{ c@{\hskip1pt} c@{\hskip1pt} c@{\hskip1pt}  c@{\hskip1pt}}     
				\begin{subtable}{0.42\linewidth} 
					{   {   
							\fontsize{7}{9}\selectfont   
							\setlength\tabcolsep{2pt}  	\renewcommand{\arraystretch}{1.2} 
							\begin{tabular}{p{0.3cm} p{2.75cm}  p{0.65cm} p{0.65cm} p{0.45cm}  p{0.45cm} p{0.5cm}  p{0.5cm}  }  
								\toprule
								\bfseries  \textit{No.} & \bfseries  Methods & \multicolumn{6}{c}{\bfseries Network Architecture}  \\  \cmidrule(lr){3-8}  
								& & \bfseries LA  & \bfseries PA & \bfseries \textit{DM} &  \bfseries \textit{Filt.} &  \bfseries \textit{\#dim}  &  \bfseries \textit{size}   
								\DTLforeach{Ablation1Supp}{ \Meth=Method, \OneK=OneK, \TwoK=TwoK, \ThreeK=ThreeK, \FourK=FourK}{
									\ifthenelse{\value{DTLrowi}=1 }{\tabularnewline \midrule }{  
										\ifthenelse{\value{DTLrowi}=2}{\tabularnewline  \cdashlinelr{1-8} }{  
											\ifthenelse{\value{DTLrowi}=4 \OR \value{DTLrowi}=6}{\tabularnewline \cdashlinelr{1-8} }{ \tabularnewline} } }
									\textit{\arabic{DTLrowi}} &\ifthenelse{\value{DTLrowi}=1}{AdaLAM~\cite{AdaLAM2020}}{\ifthenelse{\value{DTLrowi}=2}{\Ours{\Meth}}{\Ours{\Meth}}}   &  \OursLin{\Meth} & \OursPair{\Meth} & \IsMatcher{\Meth}  & \IsFilter{\Meth}  & \IsFeat{\Meth} & \IsSize{\Meth}   }  \\
								\bottomrule   
							\end{tabular} 
						}  	  
					}
				\end{subtable}
				& 
				\begin{subtable}{0.18\linewidth} 
					{   {   
							\fontsize{7}{9}\selectfont   
							\setlength\tabcolsep{2pt}  	\renewcommand{\arraystretch}{1.2} 
							\begin{tabular}{  p{0.65cm} p{0.65cm} p{0.65cm}  p{0.65cm} }  
								\toprule
								\multicolumn{4}{c}{ \bfseries Accuracy @ 0.25m,~\degc{2} }\\  \cmidrule(lr){1-4}  
								\quad 1k & \hspace{2pt} 2k & \hspace{1pt} 3k & \hspace{1pt} 4k     
								\DTLforeach{Ablation1Supp}{ \Meth=Method, \OneK=OneK, \TwoK=TwoK, \ThreeK=ThreeK, \FourK=FourK}{
									\ifthenelse{\value{DTLrowi}=1 }{\tabularnewline \midrule }{  
										\ifthenelse{\value{DTLrowi}=2}{\tabularnewline  \cdashlinelr{1-4} }{  
											\ifthenelse{\value{DTLrowi}=4 \OR \value{DTLrowi}=6}{\tabularnewline \cdashlinelr{1-4} }{ \tabularnewline} } } 
									\hspace{1pt} \BoldUndLineLARGSing{\OneK}{58}{63} & \BoldUndLineLARGSing{\TwoK}{72.0}{73} & \BoldUndLineLARGSing{\ThreeK}{78}{79} &  \BoldUndLineLARGSing{\FourK}{78}{80} }  \\
								\bottomrule   
							\end{tabular} 
						}  	  
					}
				\end{subtable}
				& 
				\begin{subtable}{0.18\linewidth} 
					{   {   
							\fontsize{7}{9}\selectfont   
							\setlength\tabcolsep{2pt}  	\renewcommand{\arraystretch}{1.2} 
							\begin{tabular}{ p{0.65cm} p{0.65cm} p{0.65cm}  p{0.65cm} }  
								\toprule
								\multicolumn{4}{c}{ \bfseries Accuracy @ 0.5m,~\degc{5} }\\ \cmidrule(lr){1-4}  
								\quad 1k & \hspace{2pt} 2k & \hspace{1pt} 3k & \hspace{1pt} 4k          
								\DTLforeach{Ablation2Supp}{ \Meth=Method, \OneK=OneK, \TwoK=TwoK, \ThreeK=ThreeK, \FourK=FourK}{
									\ifthenelse{\value{DTLrowi}=1 }{\tabularnewline \midrule }{  
										\ifthenelse{\value{DTLrowi}=2}{\tabularnewline \cdashlinelr{1-4} }{  
											\ifthenelse{\value{DTLrowi}=4 \OR \value{DTLrowi}=6}{\tabularnewline \cdashlinelr{1-4}  }{ \tabularnewline} } }
									\hspace{1pt} \BoldUndLineLARGSing{\OneK}{65}{73} & \BoldUndLineLARGSing{\TwoK}{82.5}{85} & \BoldUndLineLARGSing{\ThreeK}{86}{87} &  \BoldUndLineLARGSing{\FourK}{86}{87} }  \\
								\bottomrule   
							\end{tabular} 
						}  	  
					}
				\end{subtable} 
				& 
				\begin{subtable}{0.18\linewidth} 
					{   { 	   
							\fontsize{7}{9}\selectfont   
							\setlength\tabcolsep{2pt}  	\renewcommand{\arraystretch}{1.2} 
							\begin{tabular}{ p{0.65cm} p{0.65cm} p{0.65cm}  p{0.65cm}  }  
								\toprule
								\multicolumn{4}{c}{ \bfseries Accuracy @ 5m,~\degc{10} }\\ \cmidrule(lr){1-4}  
								\quad 1k & \hspace{2pt} 2k & \hspace{1pt} 3k & \hspace{1pt} 4k      
								\DTLforeach{Ablation3Supp}{ \Meth=Method, \OneK=OneK, \TwoK=TwoK, \ThreeK=ThreeK, \FourK=FourK}{
									\ifthenelse{\value{DTLrowi}=1 }{\tabularnewline \midrule }{  
										\ifthenelse{\value{DTLrowi}=2}{\tabularnewline \cdashlinelr{1-4} }{  
											\ifthenelse{\value{DTLrowi}=4 \OR \value{DTLrowi}=6}{\tabularnewline \cdashlinelr{1-4} }{ \tabularnewline} } } 
									\hspace{1pt} \BoldUndLineLARGSing{\OneK}{73}{76} & \BoldUndLineLARGSing{\TwoK}{93.5}{94} & \BoldUndLineLARGSing{\ThreeK}{92}{95} &  \BoldUndLineLARGSing{\FourK}{96}{98} }  \\
								\bottomrule   
							\end{tabular} 
						}  	  
					}
				\end{subtable} 
			\end{tabular}  
			\begin{tablenotes} 
				\item \footnotesize{\textbf{LA}: \LinearAttenLayerCap,  \textbf{PN}: \PairwiseAttenCapLayer},  \textbf{DM}: Distance Matching, \textbf{Filt}: Filtering process,  \textbf{\textit{\#dim}}: Encoded feature dimension, $C'$,   
				\item  \textbf{\textit{size}}: Network size, large (L) or small (S). 
			\end{tablenotes} 
		\end{threeparttable}
		
	}
	\addtocounter{table}{-1}    
\end{table*}

\begin{figure*}[t]     
	{\fontsize{6.8}{7.0}\selectfont    
		\setlength\tabcolsep{5pt}  
		\renewcommand{\arraystretch}{1.5}		
		\renewrobustcmd{\bfseries}{\fontseries{b}\selectfont}   
		\begin{tabular}{c@{\hskip1pt}   c@{\hskip1pt} | c@{\hskip1pt}     c@{\hskip1pt}     }   
			\bfseries   Our \textit{Linear.} (No Filt.)  & \bfseries	Our  \textit{Pair.Neigh.} (No Filt.)  & \bfseries  Our \textit{Linear.}  &  \bfseries	 Our \textit{Pair.Neigh.}  \\  
			\begin{minipage}{0.5\columnwidth}   
				\includegraphics [width=\textwidth, trim={0.0 0 0.0cm 0.0cm},clip]{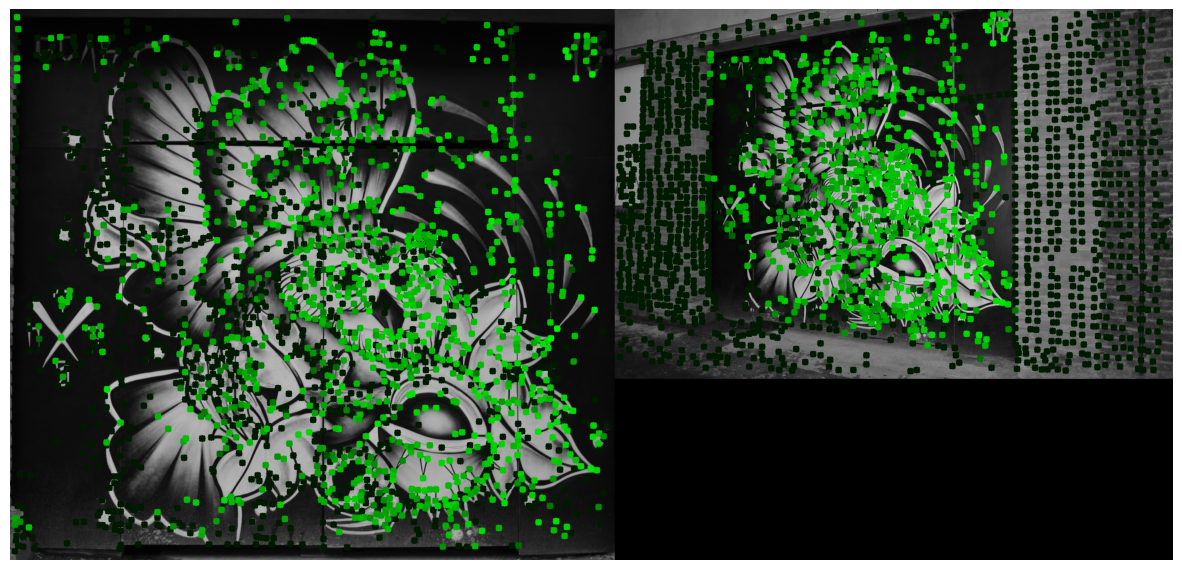} 
			\end{minipage}
			&
			\begin{minipage}{0.5\columnwidth}    
				\includegraphics [width=\textwidth, trim={0.0 0 0.0cm 0.0cm},clip] {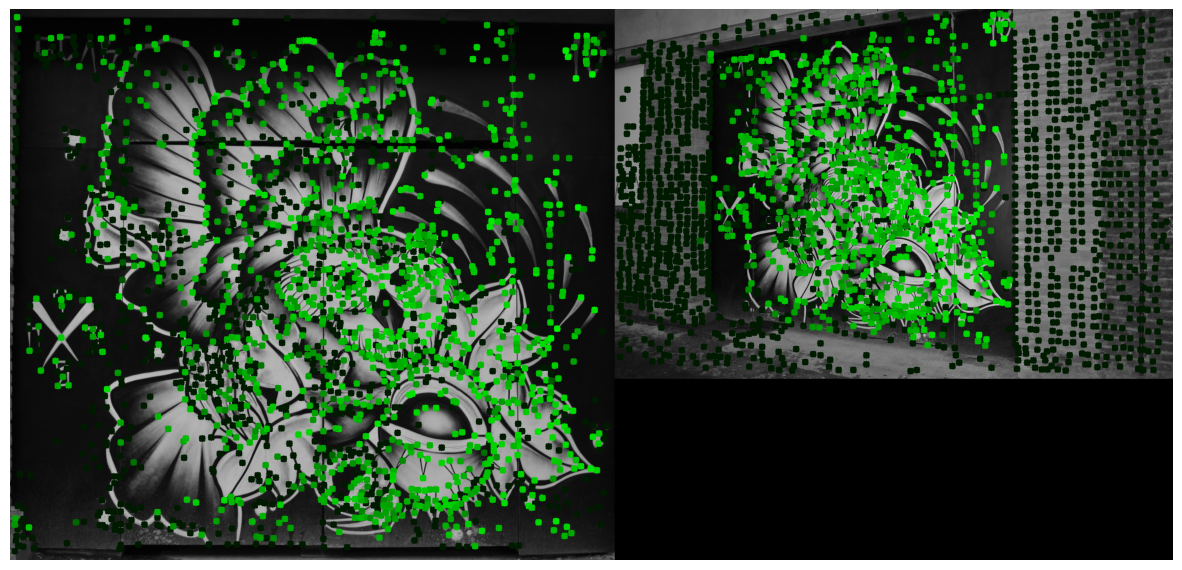} 
			\end{minipage}
			&
			\begin{minipage}{0.5\columnwidth}   
				\includegraphics [width=\textwidth, trim={0.0 0 0.0cm 0.0cm},clip]{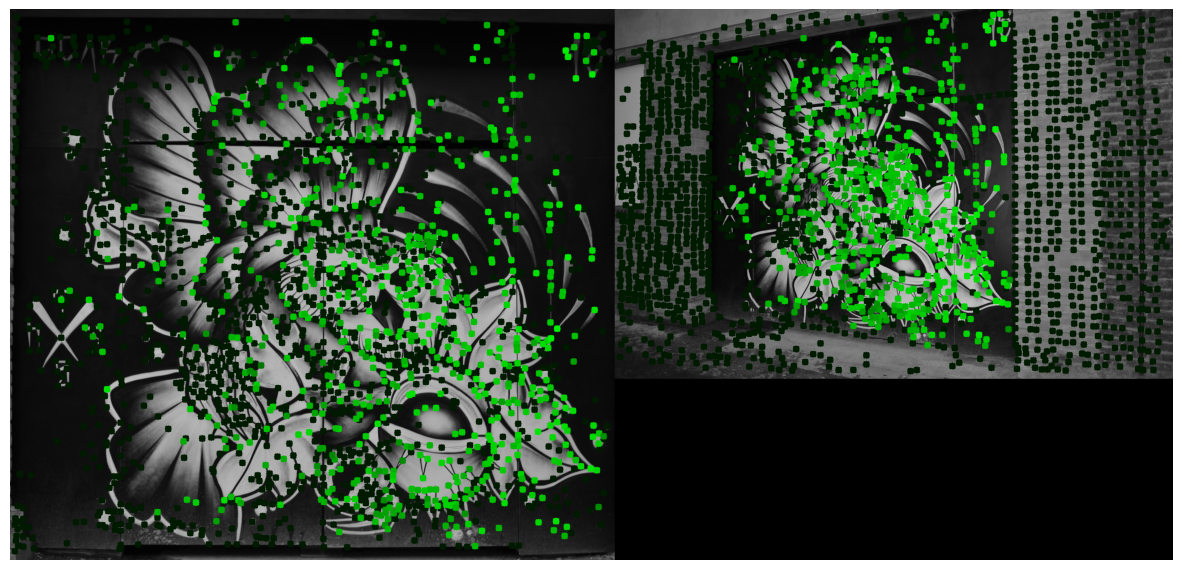} 
			\end{minipage}
			&
			\begin{minipage}{0.5\columnwidth}    
				\includegraphics [width=\textwidth, trim={0.0 0 0.0cm 0.0cm},clip] {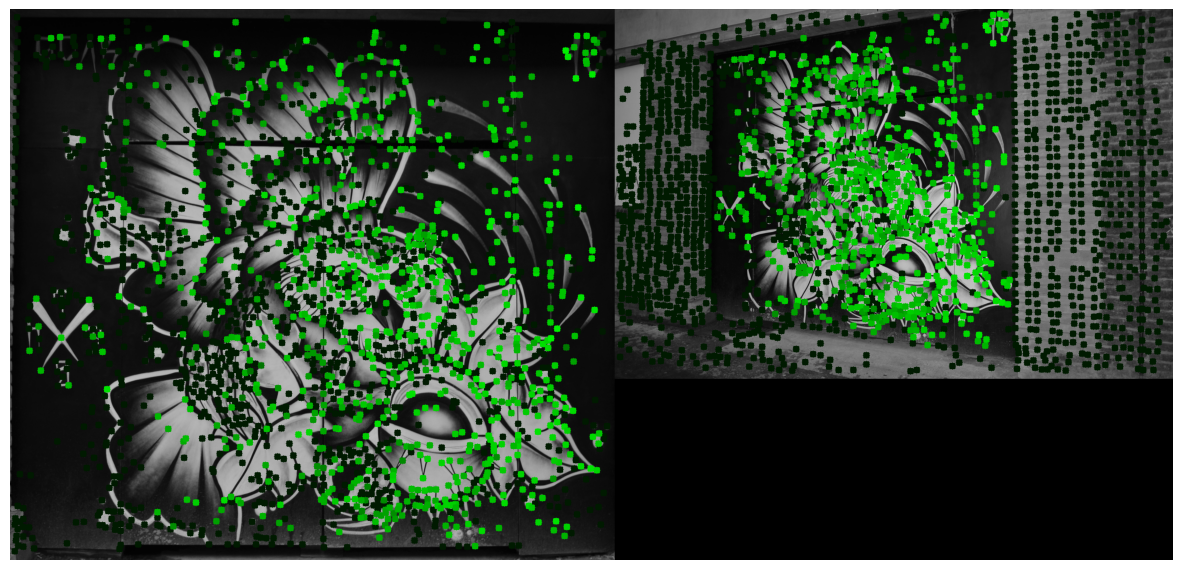} 
			\end{minipage} 
			\\ 
			\#Matches=1039 & \bfseries \#Matches=1189 	& \#Matches=692 &  \bfseries\#Matches=835  \\
			\begin{minipage}{0.5\columnwidth}   
				\includegraphics [width=\textwidth, trim={0.0 0 0.0cm 0.0cm},clip]{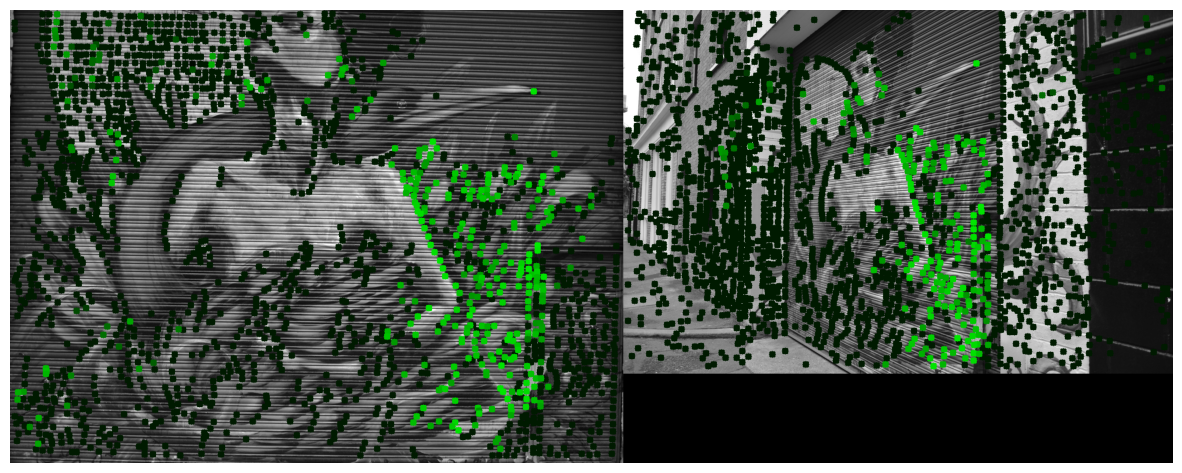} 
			\end{minipage}
			&
			\begin{minipage}{0.5\columnwidth}    
				\includegraphics [width=\textwidth, trim={0.0 0 0.0cm 0.0cm},clip] {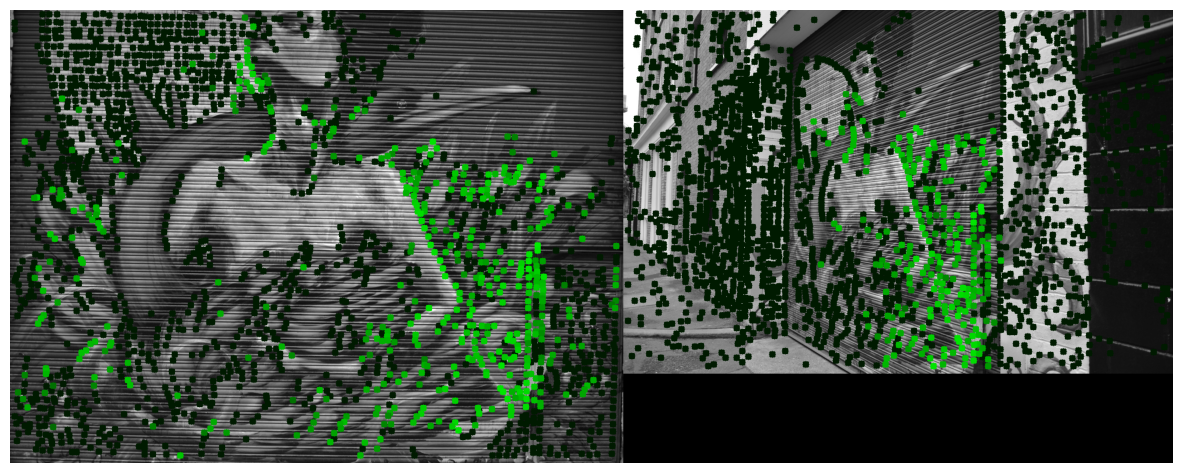} 
			\end{minipage}
			&
			\begin{minipage}{0.5\columnwidth}    
				\includegraphics [width=\textwidth, trim={0.0 0 0.0cm 0.0cm},clip] {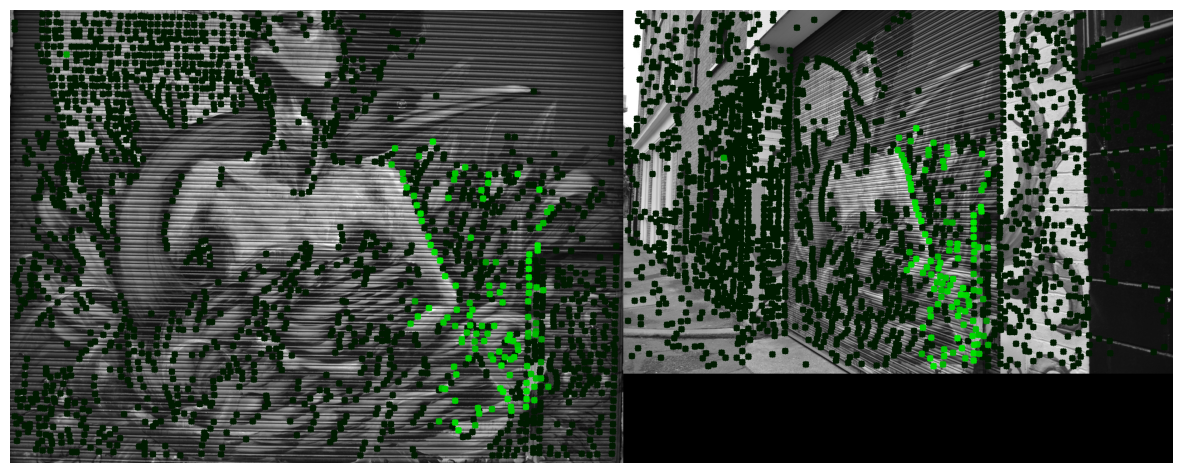} 
			\end{minipage} 
			&
			\begin{minipage}{0.5\columnwidth}   
				\includegraphics [width=\textwidth, trim={0.0 0 0.0cm 0.0cm},clip]{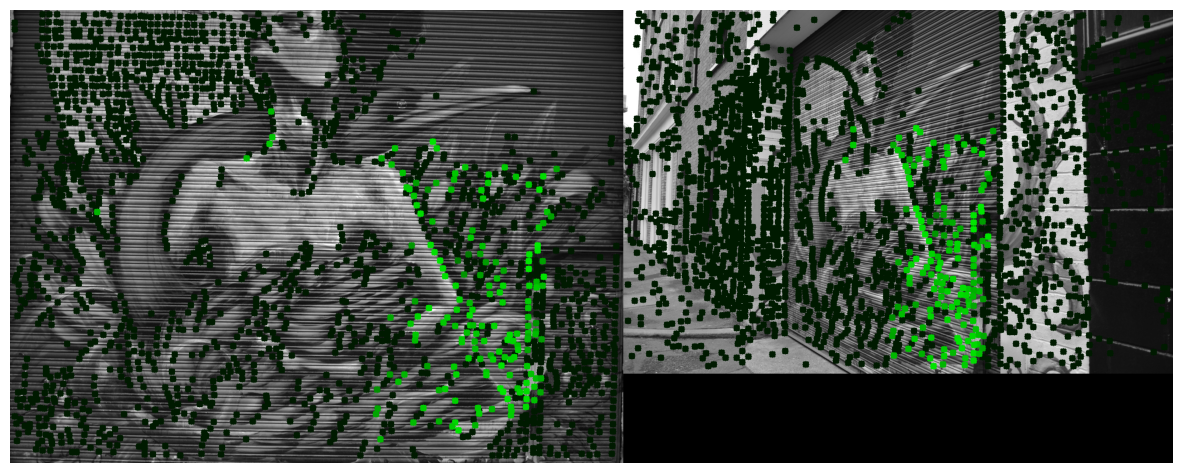} 
			\end{minipage}
			\\ 
			\#Matches=356 & \bfseries \#Matches=474	& \#Matches=123 & \bfseries \#Matches=171  \\ 
			\begin{minipage}{0.5\columnwidth}   
				\includegraphics [width=\textwidth, trim={0.0 0 0.0cm 0.0cm},clip]{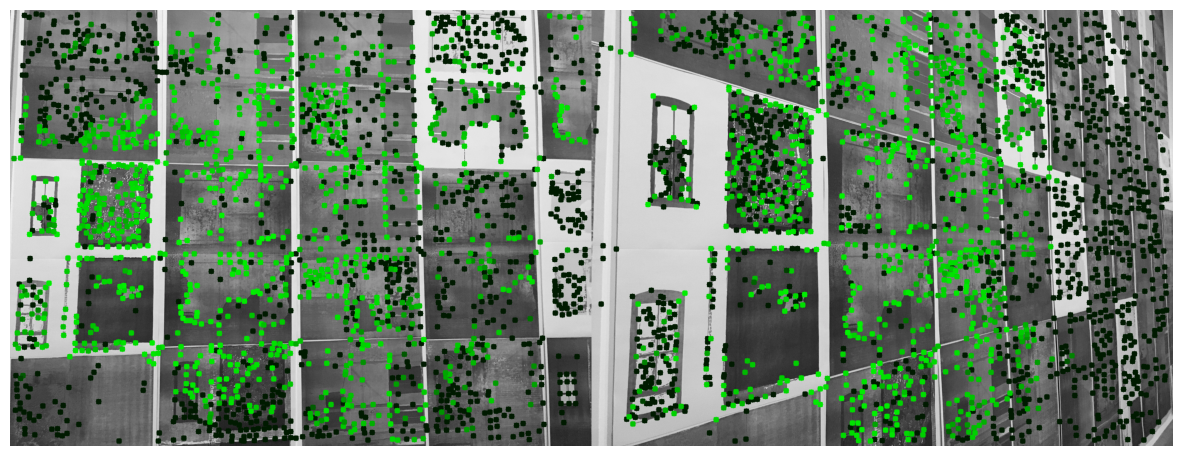} 
			\end{minipage}
			&
			\begin{minipage}{0.5\columnwidth}    
				\includegraphics [width=\textwidth, trim={0.0 0 0.0cm 0.0cm},clip] {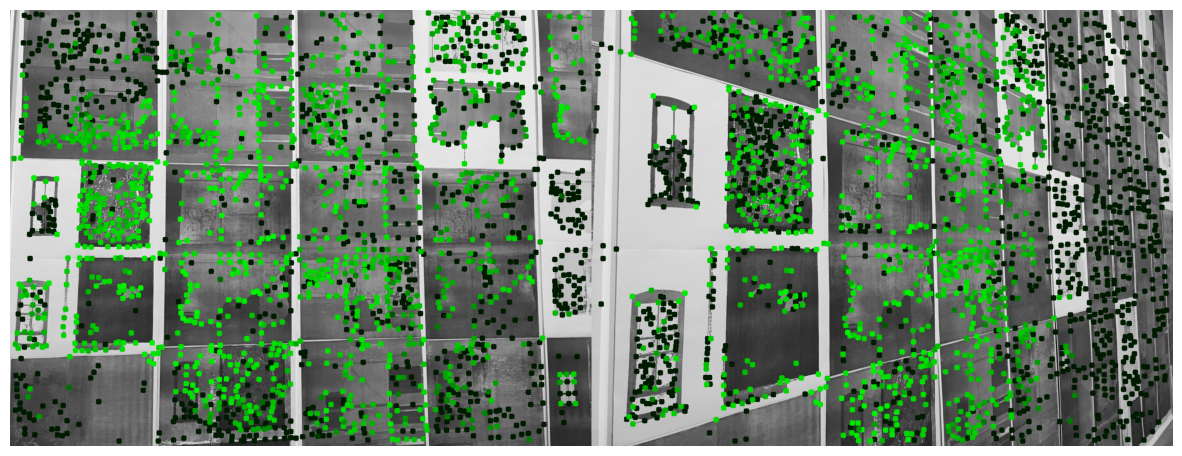} 
			\end{minipage}
			&
			\begin{minipage}{0.5\columnwidth}    
				\includegraphics [width=\textwidth, trim={0.0 0 0.0cm 0.0cm},clip] {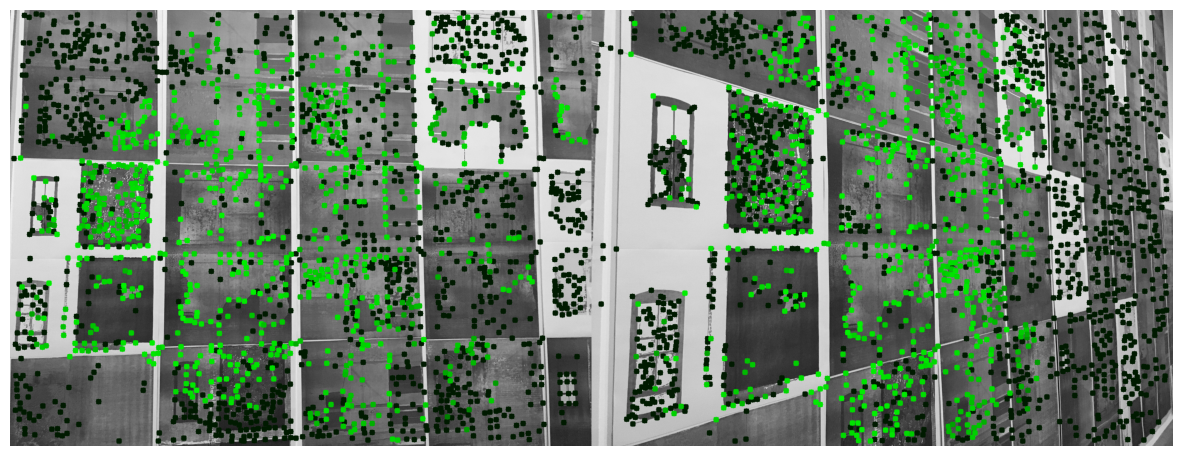} 
			\end{minipage} 
			&
			\begin{minipage}{0.5\columnwidth}   
				\includegraphics [width=\textwidth, trim={0.0 0 0.0cm  0.0cm},clip]{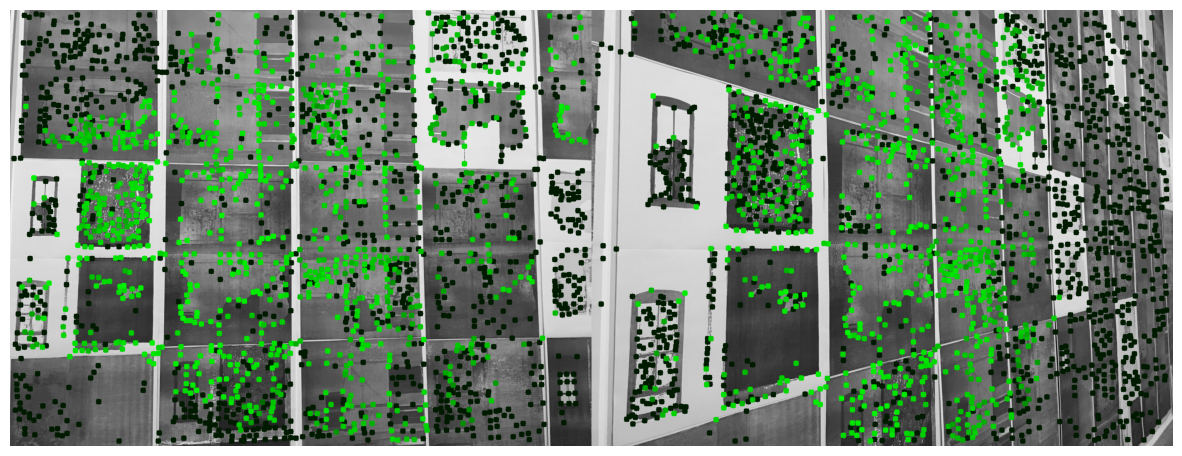} 
			\end{minipage}
			\\ 
			\#Matches=1049	& \bfseries \#Matches=1217	 & \#Matches=834 & \bfseries \#Matches=960 \\  
			\begin{minipage}{0.5\columnwidth}   
				\includegraphics [width=\textwidth, trim={0.0 0 0.0cm 0.0cm},clip]{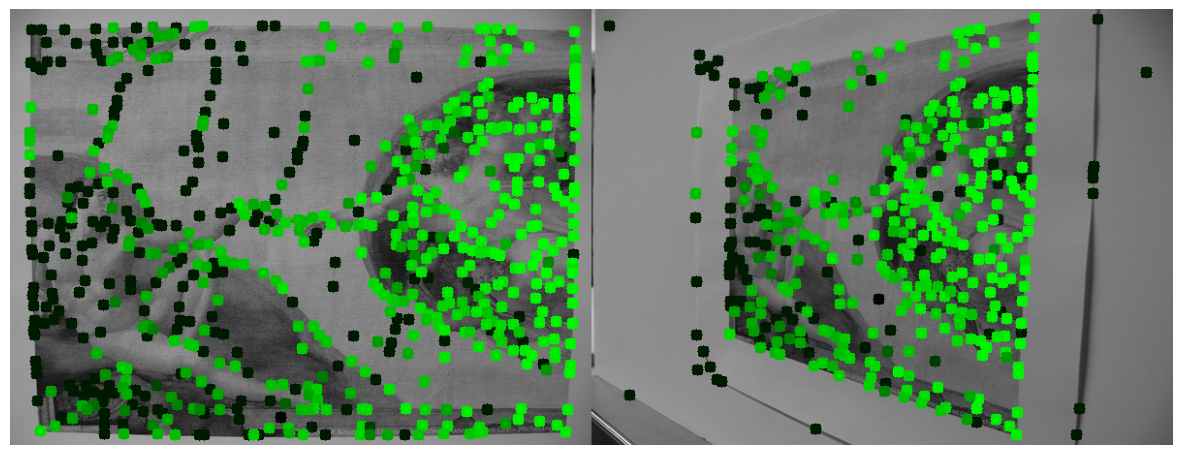} 
			\end{minipage}
			&
			\begin{minipage}{0.5\columnwidth}    
				\includegraphics [width=\textwidth, trim={0.0 0 0.0cm 0.0cm},clip] {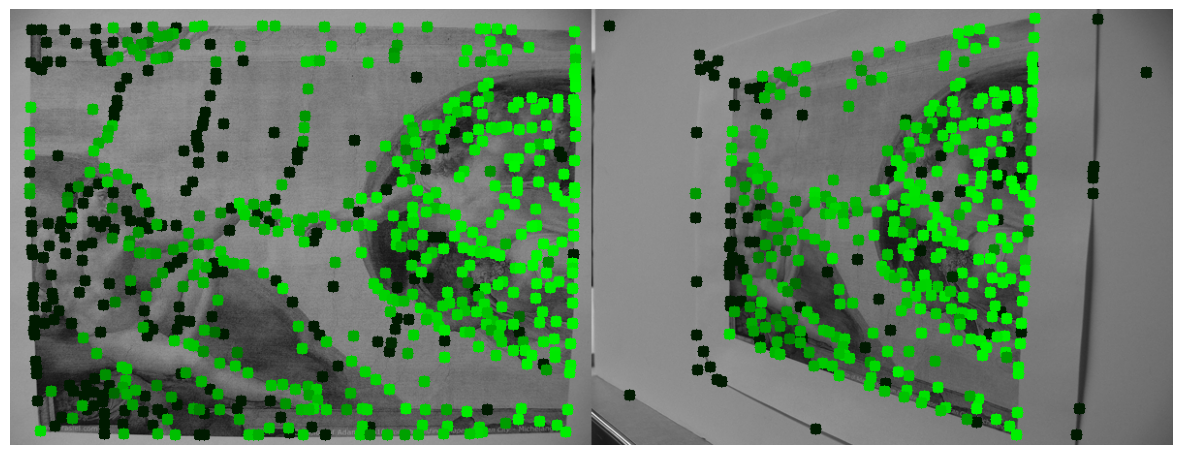} 
			\end{minipage}
			&
			\begin{minipage}{0.5\columnwidth}    
				\includegraphics [width=\textwidth, trim={0.0 0 0.0cm 0.0cm},clip] {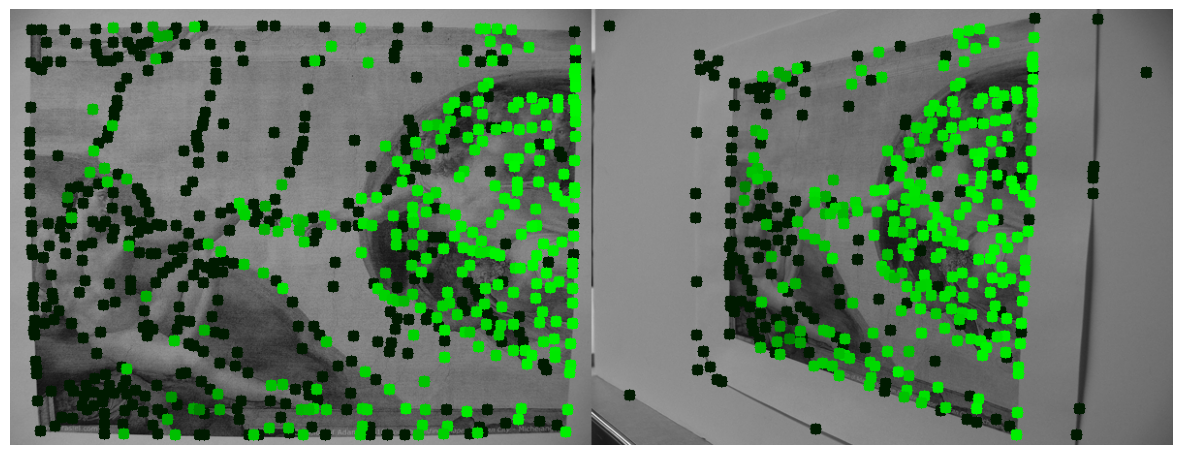} 
			\end{minipage} 
			& 
			\begin{minipage}{0.5\columnwidth}   
				\includegraphics [width=\textwidth, trim={0.0 0 0.0cm 0.0cm},clip]{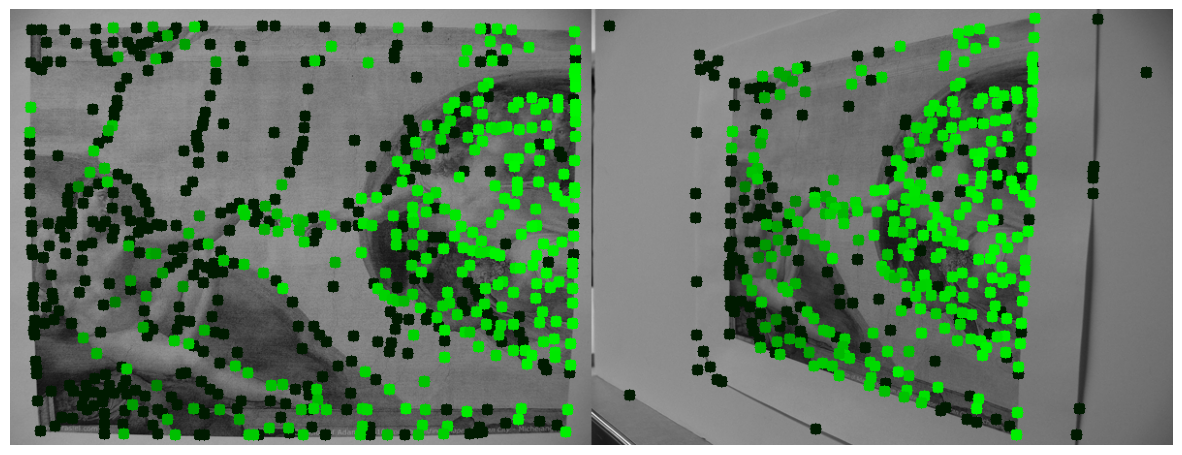} 
			\end{minipage}
			\\  
			\#Matches=418	& \bfseries \#Matches=438	 &  \#Matches=261 & \bfseries \#Matches=277 \\   
			\begin{minipage}{0.5\columnwidth}   
				\includegraphics [width=\textwidth, trim={0.0 0 0.0cm 0.0cm},clip]{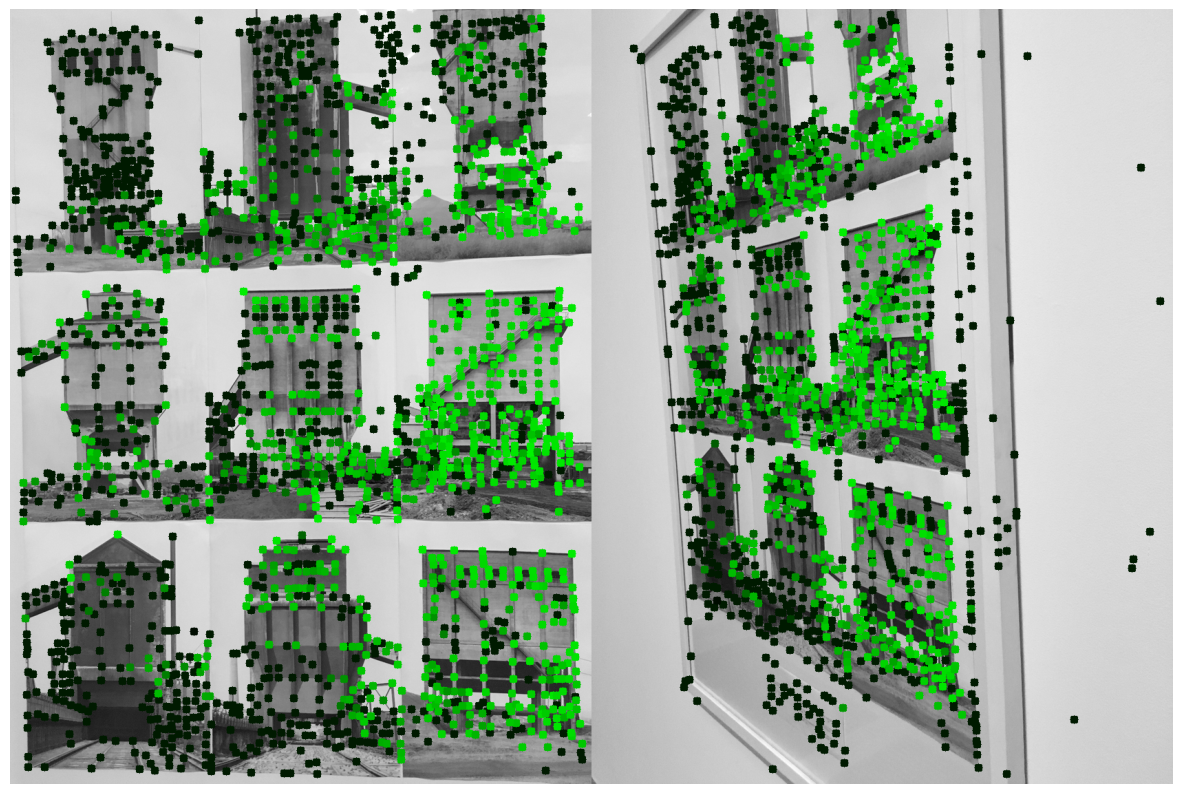} 
			\end{minipage}
			&
			\begin{minipage}{0.5\columnwidth}    
				\includegraphics [width=\textwidth, trim={0.0 0 0.0cm 0.0cm},clip] {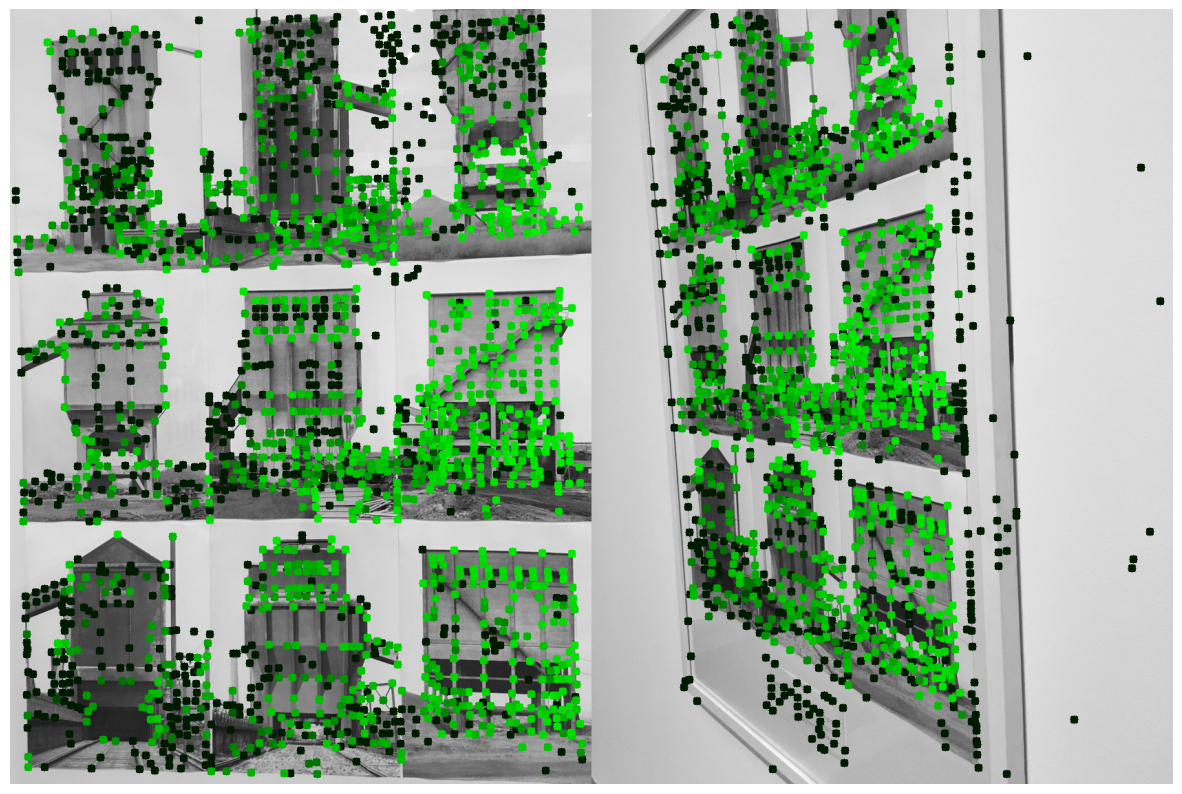} 
			\end{minipage} 
			&
			\begin{minipage}{0.5\columnwidth}    
				\includegraphics [width=\textwidth, trim={0.0 0 0.0cm 0.0cm},clip] {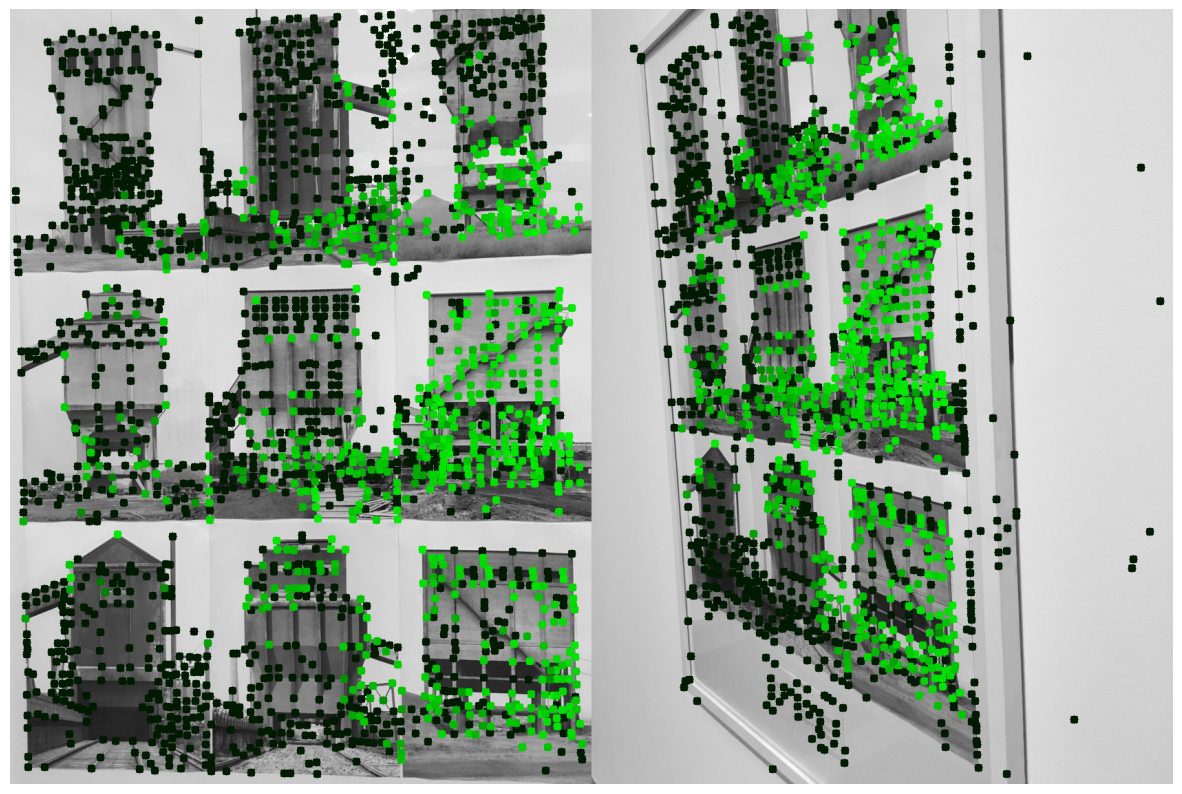} 
			\end{minipage} 
			&
			\begin{minipage}{0.5\columnwidth}   
				\includegraphics [width=\textwidth, trim={0.0 0 0.0cm 0.0cm},clip]{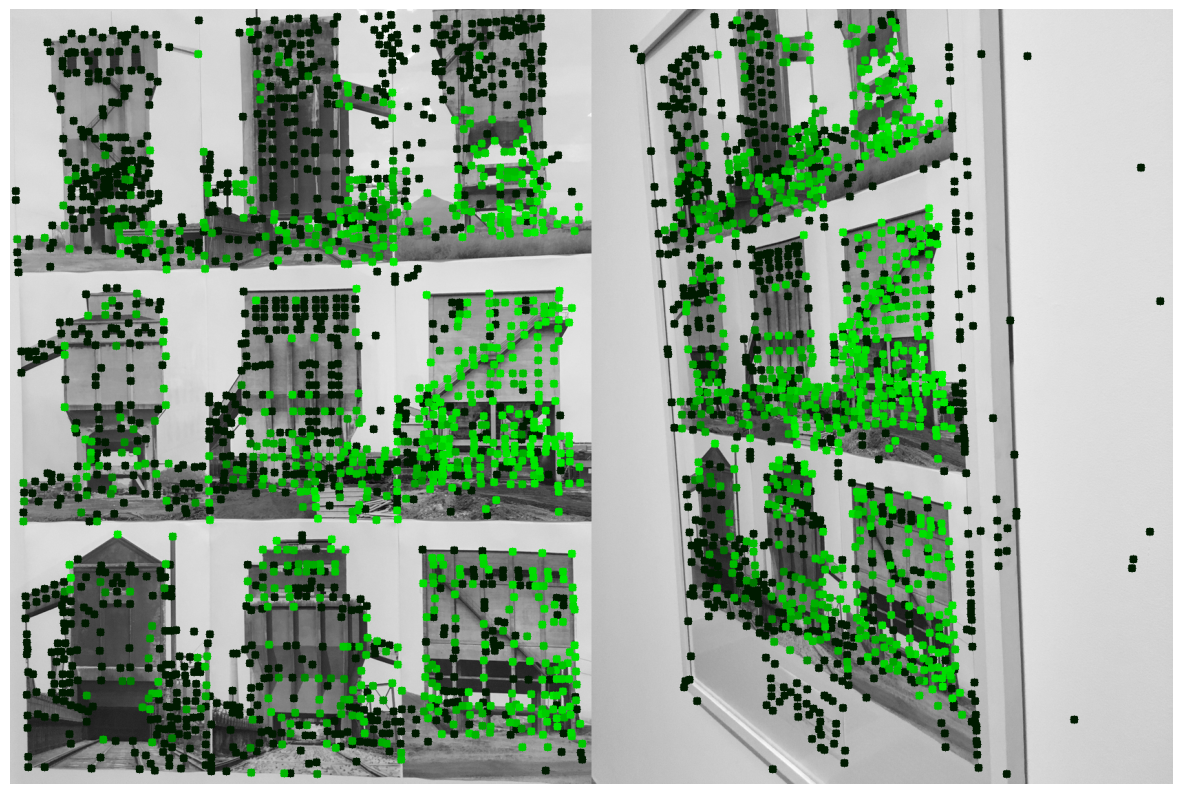} 
			\end{minipage}
			\\  
			\#Matches=976	&  \bfseries \#Matches=1232	& \#Matches=698 & \bfseries  \#Matches=853 \\      
	\end{tabular}  	}  
	\vspace{-0.2cm}
	\caption{The impact of pairwise neighborhood attention and linear attention layers on keypoint matching. We provide the keypoint matching samples by our \textit{Linear.}, \textit{Linear.} (\textbf{No Filt.}), \textit{Pair.Neigh.}, and \textit{Pair.Neigh.} (\textbf{No Filt.}). Our  \textit{Pair.Neigh.} and \textit{Pair.Neigh.} (\textbf{No Filt.}) can match more and cover more areas than  \textit{Linear.} and \textit{Linear.} (\textbf{No Filt.}) that uses only linear attention layers.  }\label{Supp:Fig:Attention} 
\end{figure*}

\begin{table*} [t]  
	{		
		\fontsize{8}{9.5}\selectfont    
		\setlength\tabcolsep{5.0pt}  
		\caption{Impact of configurations in Local Neighborhood Selection (Section~\ref{Section:Neighborhood}) on Localization Accuracy. \vspace*{-0.2cm}}	 
		\captionsetup[subfigure]{justification=centering }  
		
		\renewrobustcmd{\bfseries}{\fontseries{b}\selectfont}   
		\sisetup{detect-weight,mode=text,round-mode=places, round-precision=2}
		
		\label{Supp:tab:AbationAll2} 
		
		\begin{tabular}{ c@{\hskip1pt} c@{\hskip1pt} c@{\hskip1pt}  c@{\hskip1pt}}     
			\begin{subtable}{0.42\linewidth} 
				{   {   
						\fontsize{7}{9}\selectfont   
						\setlength\tabcolsep{2pt}  	\renewcommand{\arraystretch}{1.2} 
						\begin{tabular}{p{0.3cm} p{2.6cm}  P{1.5cm} P{2.5cm}  }  
							\toprule
							\bfseries  \textit{No.} & \bfseries  Methods & \multicolumn{2}{c}{\bfseries Local Neigh. Selection}  \\  \cmidrule(lr){3-4}  
							& & \bfseries Input &  \bfseries Seed Sep.~\eqreff{Equation:HypothesisSeeds} 
							\DTLforeach{Ablation21Supp}{ \Meth=Method, \OneK=OneK, \TwoK=TwoK, \ThreeK=ThreeK, \FourK=FourK}{
								\ifthenelse{\value{DTLrowi}=1 }{\tabularnewline \midrule }{  
									\ifthenelse{\value{DTLrowi}=3}{\tabularnewline  \cdashlinelr{1-4} }{  
										\ifthenelse{\value{DTLrowi}=4 \OR \value{DTLrowi}=6}{\tabularnewline \cdashlinelr{1-8} }{ \tabularnewline} } }
								\textit{\arabic{DTLrowi}} & \Ours{\Meth}  &  \OursInputs{\Meth} & \OursSeps{\Meth}      }  \\
							\bottomrule   
						\end{tabular} 
					}  	  
				}
			\end{subtable}
			& 
			\begin{subtable}{0.18\linewidth} 
				{   {   
						\fontsize{7}{9}\selectfont   
						\setlength\tabcolsep{2pt}  	\renewcommand{\arraystretch}{1.2} 
						\begin{tabular}{  p{0.65cm} p{0.65cm} p{0.65cm}  p{0.65cm} }  
							\toprule
							\multicolumn{4}{c}{ \bfseries Accuracy @ 0.25m,~\degc{2} }\\  \cmidrule(lr){1-4}  
							\quad 1k & \hspace{2pt} 2k & \hspace{1pt} 3k & \hspace{1pt} 4k     
							\DTLforeach{Ablation21Supp}{ \Meth=Method, \OneK=OneK, \TwoK=TwoK, \ThreeK=ThreeK, \FourK=FourK}{
								\ifthenelse{\value{DTLrowi}=1 }{\tabularnewline \midrule }{  
									\ifthenelse{\value{DTLrowi}=2}{\tabularnewline  \cdashlinelr{1-4} }{  
										\ifthenelse{\value{DTLrowi}=4 \OR \value{DTLrowi}=6}{\tabularnewline \cdashlinelr{1-4} }{ \tabularnewline} } } 
								\hspace{1pt} \BoldUndLineLARGSing{\OneK}{58}{59} & \BoldUndLineLARGSing{\TwoK}{72.0}{73} & \BoldUndLineLARGSing{\ThreeK}{74}{78} &  \BoldUndLineLARGSing{\FourK}{78}{80} }  \\
							\bottomrule   
						\end{tabular} 
					}  	  
				}
			\end{subtable}
			& 
			\begin{subtable}{0.18\linewidth} 
				{   {   
						\fontsize{7}{9}\selectfont   
						\setlength\tabcolsep{2pt}  	\renewcommand{\arraystretch}{1.2} 
						\begin{tabular}{ p{0.65cm} p{0.65cm} p{0.65cm}  p{0.65cm} }  
							\toprule
							\multicolumn{4}{c}{ \bfseries Accuracy @ 0.5m,~\degc{5} }\\ \cmidrule(lr){1-4}  
							\quad 1k & \hspace{2pt} 2k & \hspace{1pt} 3k & \hspace{1pt} 4k          
							\DTLforeach{Ablation22Supp}{ \Meth=Method, \OneK=OneK, \TwoK=TwoK, \ThreeK=ThreeK, \FourK=FourK}{
								\ifthenelse{\value{DTLrowi}=1 }{\tabularnewline \midrule }{  
									\ifthenelse{\value{DTLrowi}=2}{\tabularnewline \cdashlinelr{1-4} }{  
										\ifthenelse{\value{DTLrowi}=4 \OR \value{DTLrowi}=6}{\tabularnewline \cdashlinelr{1-4}  }{ \tabularnewline} } }
								\hspace{1pt} \BoldUndLineLARGSing{\OneK}{65}{69} & \BoldUndLineLARGSing{\TwoK}{84.5}{85} & \BoldUndLineLARGSing{\ThreeK}{84}{86} &  \BoldUndLineLARGSing{\FourK}{84}{86} }  \\
							\bottomrule   
						\end{tabular} 
					}  	  
				}
			\end{subtable} 
			& 
			\begin{subtable}{0.18\linewidth} 
				{   { 	   
						\fontsize{7}{9}\selectfont   
						\setlength\tabcolsep{2pt}  	\renewcommand{\arraystretch}{1.2} 
						\begin{tabular}{ p{0.65cm} p{0.65cm} p{0.65cm}  p{0.65cm}  }  
							\toprule
							\multicolumn{4}{c}{ \bfseries Accuracy @ 5m,~\degc{10} }\\ \cmidrule(lr){1-4}  
							\quad 1k & \hspace{2pt} 2k & \hspace{1pt} 3k & \hspace{1pt} 4k      
							\DTLforeach{Ablation23Supp}{ \Meth=Method, \OneK=OneK, \TwoK=TwoK, \ThreeK=ThreeK, \FourK=FourK}{
								\ifthenelse{\value{DTLrowi}=1 }{\tabularnewline \midrule }{  
									\ifthenelse{\value{DTLrowi}=2}{\tabularnewline \cdashlinelr{1-4} }{  
										\ifthenelse{\value{DTLrowi}=4 \OR \value{DTLrowi}=6}{\tabularnewline \cdashlinelr{1-4} }{ \tabularnewline} } } 
								\hspace{1pt} \BoldUndLineLARGSing{\OneK}{73}{75} & \BoldUndLineLARGSing{\TwoK}{93.5}{93.5} & \BoldUndLineLARGSing{\ThreeK}{92}{95} &  \BoldUndLineLARGSing{\FourK}{94}{95} }  \\
							\bottomrule   
						\end{tabular} 
					}  	  
				}
			\end{subtable} 
		\end{tabular}    
		
	} 	
\end{table*}

\begin{figure*}[t]     
	{\fontsize{6.8}{7.0}\selectfont    
		\setlength\tabcolsep{5pt}  
		\renewcommand{\arraystretch}{1.5}		
		\renewrobustcmd{\bfseries}{\fontseries{b}\selectfont}   
		\begin{tabular}{c@{\hskip1pt}   c@{\hskip1pt}  c@{\hskip1pt}     }   
			\bfseries Our \textit{Pair.-w/oSep-Inp.}   & \bfseries	 Our \textit{Pair.-w/oSep.}  & \bfseries Our \textit{Pair.Neigh.}   \\  
			\begin{minipage}{0.33\linewidth}   
				\includegraphics [width=\textwidth, trim={0.0 0 0.0cm 0.0cm},clip]{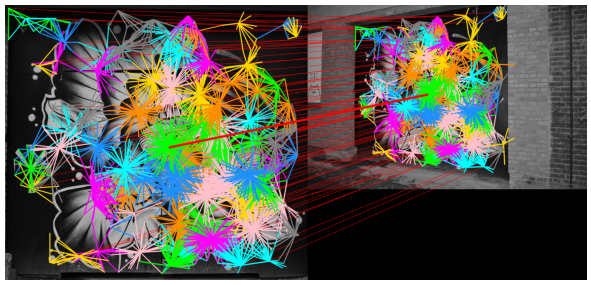} 
			\end{minipage}
			&
			\begin{minipage}{0.33\linewidth}    
				\includegraphics [width=\textwidth, trim={0.0 0 0.0cm 0.0cm},clip] {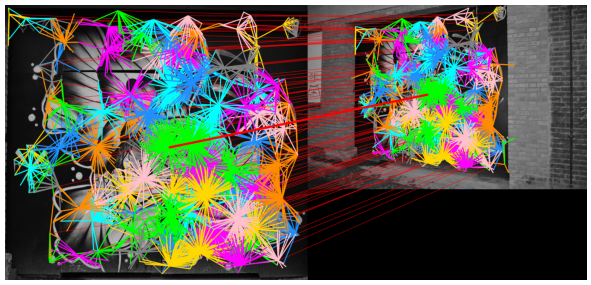} 
			\end{minipage}
			&
			\begin{minipage}{0.33\linewidth}   
				\includegraphics [width=\textwidth, trim={0.0 0 0.0cm 0.0cm},clip]{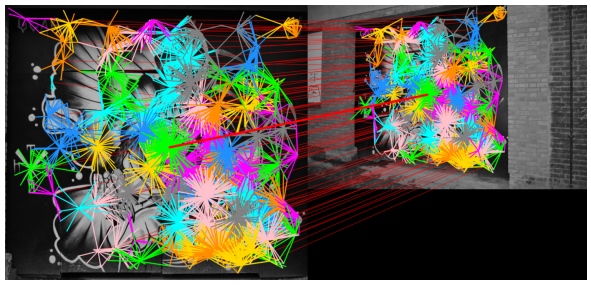} 
			\end{minipage} 
			\\ 
			\#Matching seeds=85 & \bfseries \#Matching seeds=93 	& \#Matching seeds=80   \\
			\begin{minipage}{0.33\linewidth}   
				\includegraphics [width=\textwidth, trim={0.0 0 0.0cm 0.0cm},clip]{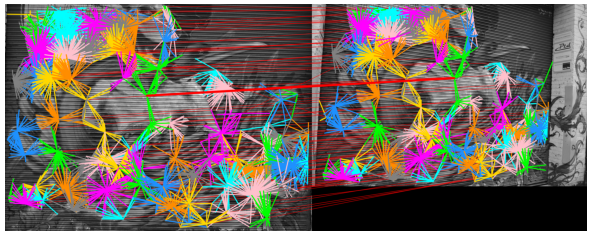} 
			\end{minipage}
			&
			\begin{minipage}{0.33\linewidth}    
				\includegraphics [width=\textwidth, trim={0.0 0 0.0cm 0.0cm},clip] {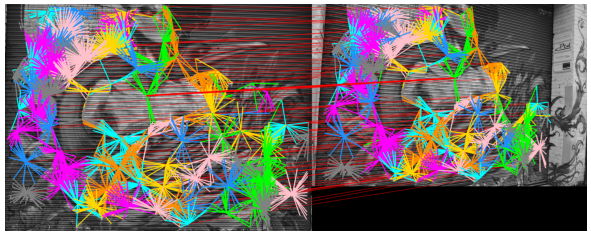} 
			\end{minipage}
			&
			\begin{minipage}{0.33\linewidth}   
				\includegraphics [width=\textwidth, trim={0.0 0 0.0cm 0.0cm},clip]{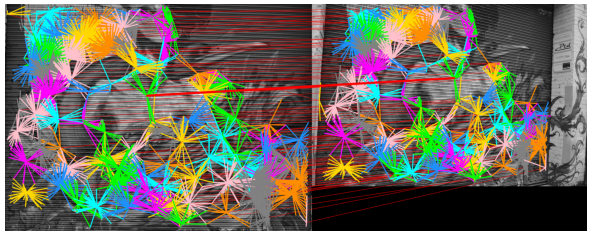} 
			\end{minipage}
			\\ 
			\# Matching seeds=82	& \bfseries \# Matching seeds=87 &  \#Matching seeds=80  \\     
			\begin{minipage}{0.33\linewidth}   
				\includegraphics [width=\textwidth, trim={0.0 0 0.0cm 0.0cm},clip]{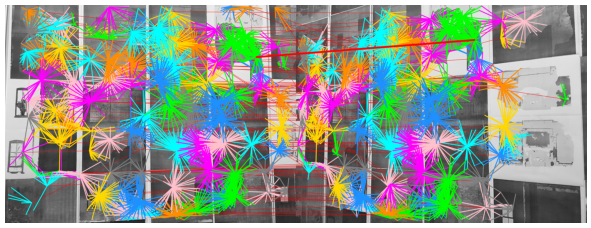} 
			\end{minipage}
			&
			\begin{minipage}{0.33\linewidth}    
				\includegraphics [width=\textwidth, trim={0.0 0 0.0cm 0.0cm},clip] {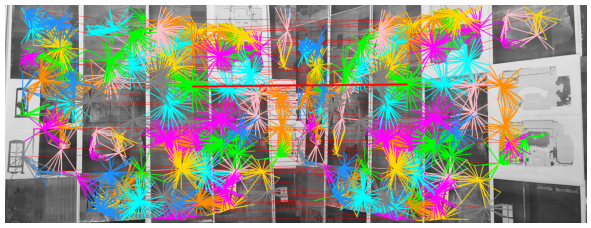} 
			\end{minipage}
			&
			\begin{minipage}{0.33\linewidth}   
				\includegraphics [width=\textwidth, trim={0.0 0 0.0cm 0.0cm},clip]{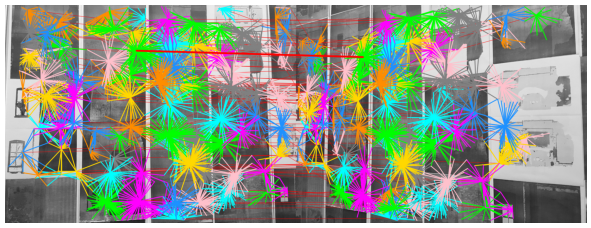} 
			\end{minipage}
			\\ 
			\# Matching seeds=99	& \bfseries \# Matching seeds=88 &  \# Matching seeds=85  \\     
	\end{tabular}  	}  
	\vspace{-0.2cm}
	\caption{The output samples from Local Neighborhood Selection (Section~\ref{Section:Neighborhood})  configured according to our \textit{Pair.-w/oSep-Input.},  \textit{Pair.-w/oSep.},  and  \textit{Pair.Neigh.} The local neighborhood selection of \textit{Pair.-w/oSep-Input.} depends on the input descriptors $\vect{x^s}, \vect{x^t}$. However, the local neighborhood selection of our  \textit{Pair.-w/oSep.}  and \textit{Pair.Neigh.} depends on $\vect{\hat{f}^s}, \vect{\hat{f}^t}$.  Nevertheless, \textit{Pair.-w/oSep.} ignores the separation condition; thus, it can collect more matching seeds than \textit{Pair.Neigh.}.  }\label{Supp:Fig:Attention2} 
\end{figure*}

\section{Additional Ablation Studies} 
\label{Section:App:Abblation}

In this section, we provide the additional results to confirm our conclusion in Section~\ref{Section:Experiment:Ablation}. We provide the results of the localization accuracy across all the three error tolerances, \ie,  (0.25m,~\ang{2}),  (0.5m,~\ang{5}), (5m,~\ang{10})  on Aachen Day-Night~\cite{Aachen2012BMVC,Aachen2018CVPR}. 

\subsection{Components in the Proposed Network.}  

 \Table{Supp:tab:AbationAll} demonstrates the impact of components in the proposed network (\Fig{Fig:Network}) on the localization accuracy across all the three error tolerances. Our \ours{4}, without any filtering  process, offers higher accuracy than \textit{Linear.} when the number of keypoints is low. This could be due to the combination of pairwise neighborhood (\textbf{PN}) and linear attention (\textbf{LA}). Meanwhile, our \textit{Linear.}  uses the {\LinearAttenLayer}s only.  Employing the filtering process (\textit{\textbf{Filt}}) can improve the performance further, yet the performance gain is more obvious with \ours{1}. The large-size model (\ours{10}) provides the highest accuracy in most cases. \Fig{Supp:Fig:Attention} demonstrates the combination of pairwise neighborhood and linear attention layers versus using linear attention layers only on the keypoint matching. We provide  the output samples of keypoint matching resulted from our \textit{Linear.}  (\textbf{No Filt.}), \textit{Pair.Neigh} (\textbf{No Filt.}), and \textit{Linear.} and \textit{Pair.Neigh}. The matched keypoints are highlighted with the green, and the unmatched keypoints are in black. The brighter color indicates the higher similarity score between the encoded features from Transformers.  Generally, \ours{1}  can match more keypoints and cover more areas than \ours{5}.

\begin{figure*}[t]  
	\hspace{-0.75cm}  
	\begin{minipage}[b]{0.45\linewidth}
		\includegraphics[trim=10 0 0 0,clip,width=1.0\textwidth]{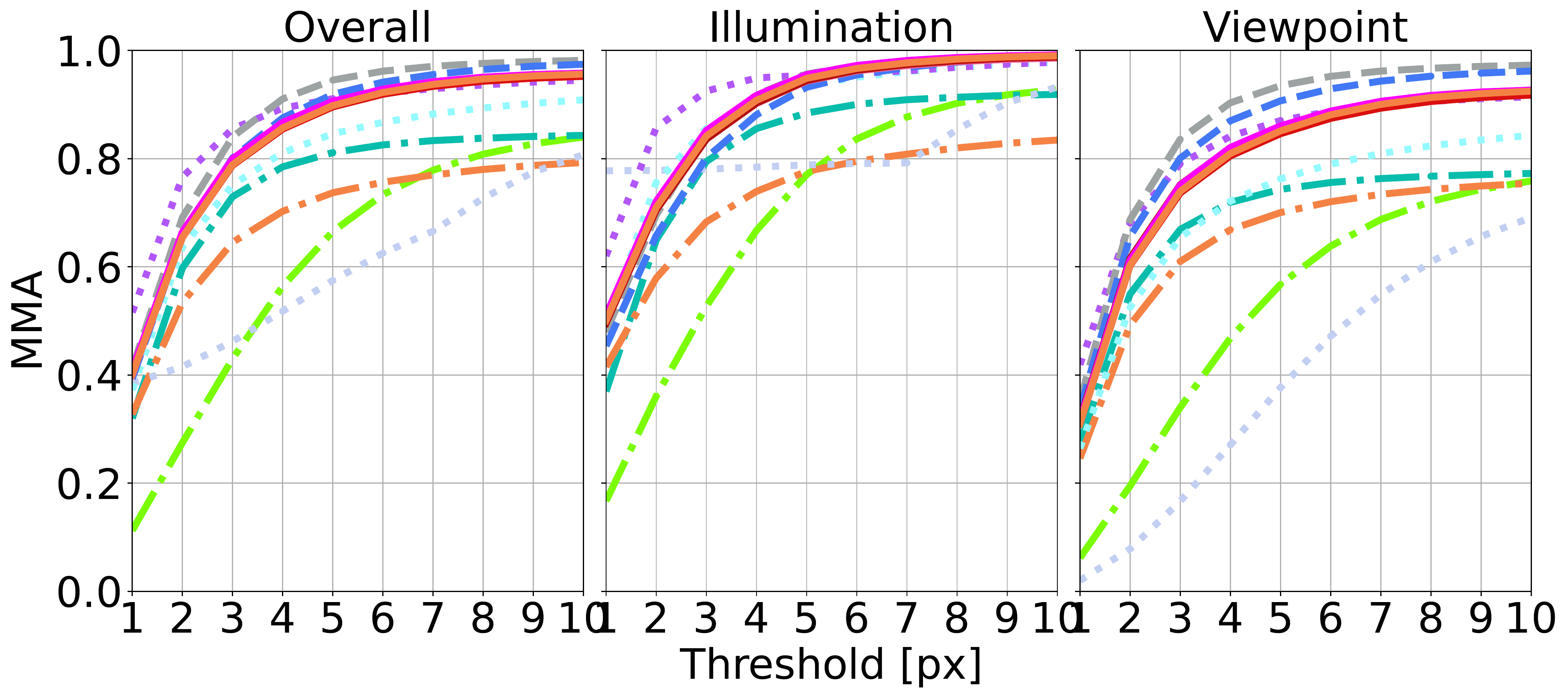}  \\ \vspace{0.5cm}
		\includegraphics[trim=30 1060 30 10,clip,width=1\textwidth]{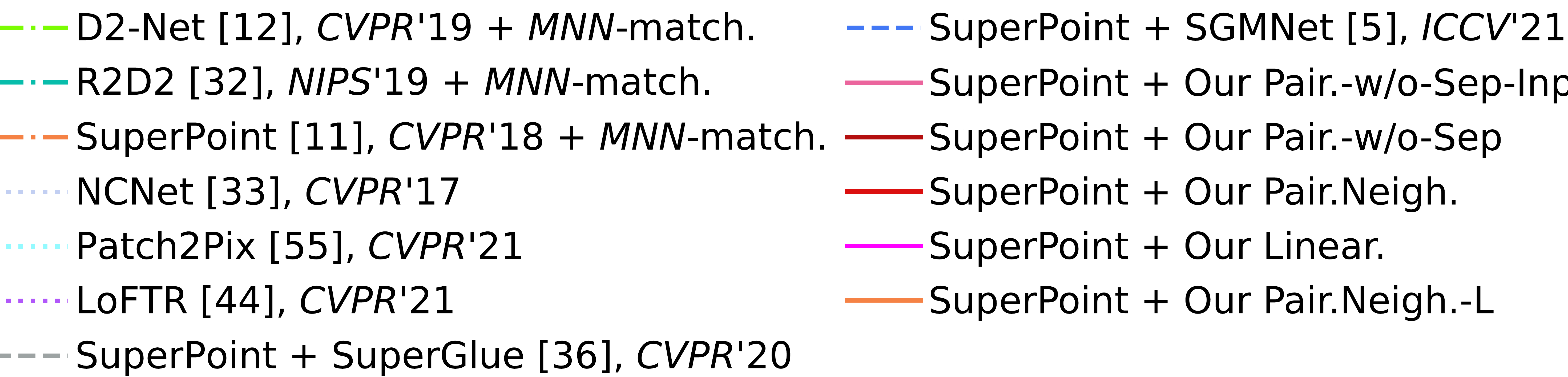}  
	\end{minipage}%
	\hspace{-0.01cm}    
	\begin{minipage}[b]{0.5\linewidth}   
		\renewrobustcmd{\bfseries}{\fontseries{b}\selectfont}   
		\sisetup{detect-weight,mode=text,round-mode=places, round-precision=2}
		{\fontsize{7.05}{8.75}\selectfont   
			{  \setlength\tabcolsep{1pt}  	\renewcommand{\arraystretch}{1.2} 
				\begin{threeparttable} 
					\begin{tabular}{  l@{\hskip2pt} P{1.0cm} P{1.3cm} P{1.00cm} P{1.1cm} P{1.7cm} }  %
						\toprule  
						\bfseries	Methods &  \bfseries	 Pub.   & \bfseries \textit{\#Matches} & \bfseries \textit{Inl.Ratio}    & \bfseries \textit{\#Param.}   & \bfseries \textit{Total Time(ms)\tnote{$\ddagger$}}   
						\DTLforeach{HPatchSupp}{ \Meth=Meth, \Matc=Matc, \Ini=Inli, \Rtim=Rtim, \Para=Para, \SpaceVal=SpaceVal, \NoteF=NoteF, \NoteM=NoteM, \Pubs=Pub}{
							\ifthenelse{\value{DTLrowi}=1}{\tabularnewline \midrule   }{ 
								\ifthenelse{\value{DTLrowi}=4 \OR \value{DTLrowi}=7}{ \tabularnewline \cdashlinelr{1-6}}{ 
									\ifthenelse{\value{DTLrowi}=12}{ \tabularnewline \cdashlinelr{1-6}}{	
										\ifthenelse{\value{DTLrowi}=9}{ \tabularnewline \midrule}{ \tabularnewline } } } 
							} %
							\ifthenelse{\value{DTLrowi}<4}{   \Meth  \addspace{\SpaceVal}+ \MNN-matching}{
								\ifthenelse{\value{DTLrowi}<9}{ \Meth   }{\NoteF $~+$ \Ours{\NoteM}   }   }   & 
							\textit{$\Pubs$} &
							\BoldUndLineLARGK{\Matc}{1500}{4000} &   \BoldUndLineLARG{\Ini}{0.805}{0.86}  
							&\BoldUndLineSMALLESTMK{\Para}{1000000}{500000} &  \ifthenelse{\value{DTLrowi} = 13 }{  \hspace{-1pt}\BoldUndLineSMALLEST{\Rtim}{69}{35} }{ \ifthenelse{ \value{DTLrowi} > 8  }{\hspace{3pt} \BoldUndLineSMALLEST{\Rtim}{69}{35}}{\BoldUndLineSMALLEST{\Rtim}{69}{35}} 
						}    }  \\  
						\bottomrule   
					\end{tabular}
					\begin{tablenotes}
						\item[$\ddagger$] Total Time = Feature Extraction Time + Sparse/Dense Matching Time.  \vspace{-0.2cm} 
					\end{tablenotes}
				\end{threeparttable} 
			}  	  
		}
		
	\end{minipage}   
	\caption{Image matching. Our method versus SOTAs --- local features, dense matching, and sparse matching --- on HPatches~\cite{HPatch}.   } 
	\label{Supp:Fig:HPatch}   
\end{figure*}

\subsection{Local Neighborhood Selection.}    
\Table{Supp:tab:AbationAll2} provides the impact on localization accuracy due to the configuration of Local Neighborhood Selection (Section~\ref{Section:Neighborhood}). ~\Ours{2} offers higher robustness when number of keypoints are low. Meanwhile, \ours{1} and \ours{0} offer the highest accuracy when the number of keypoints are high, as $\vect{\hat{f}^s}, \vect{\hat{f}^t}$ are resulted from the aggregation of information. Using both \PairwiseAtten~and {\LinearAttenLayer}s~offers higher accuracy than using only \LinearAttenLayer~in most cases. 

\Fig{Supp:Fig:Attention2} demonstrates the output samples from the  Local Neighborhood Selection by our~\ours{2}, \ours{0} and \ours{1} at 2k keypoints.  The local neighborhood selection of ~\ours{2} depends on the input descriptors $\vect{x^s}, \vect{x^t}$. Meanwhile, the local neighborhood selection of \ours{0} and \ours{1} depends on $\vect{\hat{f}^s}, \vect{\hat{f}^t}$.  Since \ours{0} ignores the separation condition, it can collect more matching seeds than \ours{1} Despite having less matching seeds, \ours{1}  employs the seed separation condition which enforces the matching seeds to be spreading across the images. The spreading of matching seeds shows to be an important factor in gaining high accuracy in localization according to \Table{Supp:tab:AbationAll2}.

\section{Additional SOTA Comparisons}  
\label{Section:App:Visual}
This section provides the additional results of our work on \ours{11} and \ours{10}, where  Table~\ref{Supp:Table:Configs} provides the summary of all the configuration and settings of our proposed method:    

\begin{itemize}
	\item 	\Ours{11} \hspace{1.3cm} Configuration \No{4} in \Table{Supp:Table:Configs};  
	\item 	\Ours{10} \hspace{0.5cm} Configuration \No{5} in \Table{Supp:Table:Configs}.  
\end{itemize} 

\noindent The results are provided in the last two rows in addition to those  results reported in the main paper, including our \ours{1}, \ours{2}, and \ours{0}, which correspond to Configuration \No{1}, \No{2}, \No{3} in \Table{Supp:Table:Configs}. We compare with: \textbf{$\bullet$  Sparse matching:} SuperGlue~\cite{SuperGlue_CVPR2020} and SGMNet~\cite{SGMNet_ICCV2021};  \textbf{ $\bullet$  Dense matching:}  LoFTR~\cite{LOFTR_CVPR2021}, Patch2Pix~\cite{Patch2pix_CVPR2021}, NCNet~\cite{NCNet_CVPR2017};  \textbf{ $\bullet$  Local features:} SuperPoint~\cite{CVPR2018:SuperPoint}, R2D2~\cite{NIPS2019:R2D2},   D2-Net~\cite{CVPR2019:D2Net}, and  ASLFeat~\cite{CVPR2020:ASLFeat}, with the standard matching;  \textbf{ $\bullet$ Keypoint filtering:} AdaLAM~\cite{AdaLAM2020} and OANet~\cite{OANet_ICCV2019}.

\subsection{Image Matching  \& Visual Results}  
\label{Supp:Section:Matching-Visual}
This section provides the additional numerical results of our \ours{11} and \ours{10} and the visual results of our method (\Ours{1}) on image matching task on HPatches~\cite{HPatch}.
 
\smallskip 
\noindent\textbf{Numerical Results.} From~\Fig{Supp:Fig:HPatch}, \ours{11} and \ours{10} offers similar MMA curve to the other configuration of our work, \Ours{11} offers higher \InlierRat, but lower matches. Meanwhile, \ours{10} offers higher matches, but also 10-20 ms more runtime than other configurations of our works.     

\smallskip 
\noindent\textbf{Visual Results.} \Fig{tab:Supp:Vis1}  and \Fig{tab:Supp:Vis3}  provide the selected visual results of image matching  on illumination and viewpoint changes between our method (\ours{1}) versus  SuperGlue and SGMNet. The correct and incorrect matches are denoted with green and red color, respectively. From the results on illumination changes in~\Fig{tab:Supp:Vis1},  our method  provides the highest MMA with less incorrect matches on the illumination changes, which corresponds to the results in~\Fig{Supp:Fig:HPatch}. Meanwhile, on viewpoint changes in~\Fig{tab:Supp:Vis3}, our work provides the accurate matches, but it achieves slightly lower performance due to the lower number of matches.

\begin{table} [t]   
	\caption{Summary of the settings \& configurations. }	  
	\renewrobustcmd{\bfseries}{\fontseries{b}\selectfont}   
	\sisetup{detect-weight,mode=text,round-mode=places, round-precision=2}
	{   {  	  
			
			\fontsize{7}{9}\selectfont   
			\setlength\tabcolsep{5pt}  	\renewcommand{\arraystretch}{1.2} 
			\begin{tabular}{c@{\hskip5pt}   l   c@{\hskip2pt}c@{\hskip2pt}c@{\hskip2pt}c@{\hskip3pt} c@{\hskip2pt} c@{\hskip3pt}  | c@{\hskip1pt} c@{\hskip1pt}  }    
				\toprule	
				\bfseries  No.  & \bfseries  Name & \multicolumn{6}{c}{\bfseries Network Architecture} &  	\multicolumn{2}{c}{\bfseries Local Neigh. Sel.}	\\   \cmidrule(lr){3-8}  \cmidrule(lr){9-10} 
				&   & \bfseries LA  & \bfseries PA & \bfseries \textit{DM} &  \bfseries \textit{Filt.} &  \bfseries \textit{\#dim}  &  \bfseries \textit{size}   &  \bfseries Inp. &  \bfseries Sep.~\eqreff{Equation:HypothesisSeeds} \\  \midrule 
				\textit{1}	& \Ours{2} & \OursLin{10} & \OursPair{10} & \IsMatcher{10}  & \IsFilter{10}  & \IsFeat{6} & \IsSize{6} & \OursInputs{2} & \OursSeps{2}  \\   
				\textit{2}	& \Ours{0} & \OursLin{10} & \OursPair{10} & \IsMatcher{10}  & \IsFilter{10}  & \IsFeat{6} & \IsSize{6}& \OursInputs{0} & \OursSeps{0} \\   
				\textit{3}	& \Ours{1} & \OursLin{10} & \OursPair{10} & \IsMatcher{10}  & \IsFilter{10}  & \IsFeat{6} & \IsSize{6} & \OursInputs{1} & \OursSeps{1}  \\  \cdashlinelr{1-10}
				\textit{4}	& \Ours{11} & \OursLin{5} & \OursPair{5} & \IsMatcher{5}  & \IsFilter{5}  & \IsFeat{5} & \IsSize{5} & N/A  & N/A \\    
				\textit{5}	& \Ours{10} & \OursLin{10} & \OursPair{10} & \IsMatcher{10}  & \IsFilter{10}  & \IsFeat{10} & \IsSize{10} & \OursInputs{1} & \OursSeps{1} \\    \bottomrule
			\end{tabular} 
		}  	  
	}
	\label{Supp:Table:Configs}
\end{table} 

\subsection{3D Reconstruction  \& Visual Results}  
\label{Supp:Section:3DRecon-Visual}

This section we provide the additional numerical results of \ours{11} and \ours{10} on 3D reconstruction using ETH small- and medium-size datasets~\cite{Eval:3DReconETH}. We also provide the visual results of  our \ours{1} in comparison with the SOTAs, SuperGlue and SGMNet.  

\smallskip 
\noindent\textbf{Numerical Results.}  \Table{Supp:Table:3DRecon-Small} and \Table{Supp:Table:3DRecon-Medium} provide the results on the small-size datasets---Fountain, Herzjesu, and South-Building---and the medium-size datasets--- Madrid Metropolis, Gendarmenmarkt, and Tower of London. From \Table{Supp:Table:3DRecon-Small}, our \ours{11} and \ours{10} offer similar performance to the other configurations of our work, \ie,  long \Track, lower \Reproj, and comparable \DensePoint~to SuperGlue and SGMNet, SuperGlue-10, SGMNet-10. Despite our \ours{10} being a larger model, its \MatchTime~ only slightly higher than the other configurations of our work, which is about  \textbf{\textit{7 times}} and \textbf{\textit{1.6  times} faster} than SuperGlue-10 and SGMNet-10, respectively.  From \Table{Supp:Table:3DRecon-Medium}, our \ours{11} and \ours{10} provide long \Track~, low \Reproj, and moderate  \DensePoint, in most cases. Both \ours{11} and \ours{10} achieve these results with similar runtime to the other configurations of our work, which is lower than the runtime of SuperGlue-10 and SGMNet-10. That is, the \MatchTime~is about \textbf{\textit{3 times}} and \textbf{\textit{twice faster}} than SuperGlue-10 and SGMNet-10,  due to the lower detected keypoints by SuperPoint.

 \begin{table}[t] 
 	\caption{Small-size ETH. Our methods versus the official Superglue and SGMNet, Superglue-10 and SGMet-10. }
 	\label{Supp:Table:3DRecon-Small}
 	\vspace{-0.3cm}
 	\begin{threeparttable}[b]
 		\renewrobustcmd{\bfseries}{\fontseries{b}\selectfont} 
 		\sisetup{detect-weight,mode=text,round-mode=places, round-precision=2}
 		{\fontsize{7.0}{7.0}\selectfont 
 			\setlength\tabcolsep{0.1pt}  
 			\setlength{\extrarowheight}{2pt} 
 			\begin{tabular}{p{1.2cm}  l@{\hskip3pt}   p{0.8cm} p{0.8cm} P{0.95cm} P{0.9cm}  P{1.8cm} }   
 				\toprule   \bfseries Datasets & \bfseries Methods     &  \bfseries \textit{\Track}  & \bfseries \textit{\Reproj} & \bfseries \textit{Sparse Points} &  \bfseries \textit{Dense Points} &  \bfseries \textit{Match. Time(sec)} 
 				\DTLforeach{SfMSupp:Herzjesu}{  \numreg=num_reg_images,   \numsparse=num_sparse_points_round,  \numsparseunit=num_sparse_points_unit, \tracklen=mean_track_length, \reprojerr=mean_reproj_error, \numdense=num_dense_points_round, \numdenseunit=num_dense_points_unit,\numdense=num_dense_points_round,  \numdenseunit=num_dense_points_unit,  \Mtime=MatchingTime, \MtimeV=MatchingRunTime, \note=note, \re=ref, \fnote=fnote}%
 				{  	 \ifthenelse{\value{DTLrowi}=1}{\tabularnewline \midrule    }{   	\ifthenelse{\value{DTLrowi}=5   }{\tabularnewline \cdashlinelr{2-7} }{
 							\ifthenelse{\value{DTLrowi}=3 \OR  \value{DTLrowi}=8}{\tabularnewline \cdashlinelr{2-7} }{ \tabularnewline } 	}} 
 					\ifthenelse{\value{DTLrowi}=1}
 					{  \multirow{2}{11mm}{ \bfseries Herzjesu 8 images   }  & \RefC{\note}{\re}\fnote & 
 						\BoldUndLineLARGDB{\tracklen}{4.525}{4.54} &   
 						\BoldUndLineSMALLESTThree{\reprojerr}{0.874}{0.858}   &  
 						\BoldUndLineLARGSingU{\numsparse}{8.4}{9.7}{\numsparseunit} &   
 						\BoldUndLineLARGDBU{\numdense}{1.145}{1.145}{\numdenseunit}    & 
 						\BoldUndLineSMALLESTKK{\MtimeV}{24.4}{23}   
 					} 
 					{  
 						\ifthenelse{\value{DTLrowi} < 5}
 						{&  \RefC{\note}{\re}\fnote & 	
 							\BoldUndLineLARGDB{\tracklen}{4.525}{4.54} &  
 							\BoldUndLineSMALLESTThree{\reprojerr}{0.874}{0.858}  &   
 							\BoldUndLineLARGSingU{\numsparse}{8.4}{9.7}{\numsparseunit} &   
 							\BoldUndLineLARGDBU{\numdense}{1.145}{1.145}{\numdenseunit}    &  
 							\BoldUndLineSMALLESTKK{\MtimeV}{24.4}{23}    
 						}
 						{   
 							& \Ours{\note}  & 	
 							\BoldUndLineLARGDB{\tracklen}{4.525}{4.54} &  
 							\BoldUndLineSMALLESTThree{\reprojerr}{0.874}{0.858}   &  
 							\BoldUndLineLARGSingU{\numsparse}{8.4}{9.7}{\numsparseunit} &   
 							\BoldUndLineLARGDBU{\numdense}{1.145}{1.145}{\numdenseunit}    & 
 							\BoldUndLineSMALLESTKK{\MtimeV}{24.4}{23}   
 						}
 					}
 				}       
 				\DTLforeach{SfMSupp:Fountain}{  \numreg=num_reg_images,    \numsparse=num_sparse_points_round,  \numsparseunit=num_sparse_points_unit, \tracklen=mean_track_length, \reprojerr=mean_reproj_error,\numobs=num_observations_round,  \numdense=num_dense_points_round,  \numdenseunit=num_dense_points_unit,  \numdenseunit=num_dense_points_unit, \Mtime=MatchingTime, \MtimeV=MatchingRunTime, \note=note, \re=ref, \fnote=fnote}%
 				{\ifthenelse{\value{DTLrowi}=1}{\tabularnewline \toprule    }{   \ifthenelse{\value{DTLrowi}=5   }{\tabularnewline \cdashlinelr{2-7} }{    
 							\ifthenelse{\value{DTLrowi}=3 \OR  \value{DTLrowi}=8}{\tabularnewline \cdashlinelr{2-7} }{ \tabularnewline } 	}} 
 					\ifthenelse{\value{DTLrowi}=1}
 					{  \multirow{2}{11mm}{   \bfseries Fountain 11 images }  &   \RefC{\note}{\re}\fnote &  
 						\BoldUndLineLARGDB{\tracklen}{5.14}{5.16} &  
 						\BoldUndLineSMALLESTThree{\reprojerr}{0.9035}{0.90}   &  
 						\BoldUndLineLARGSingU{\numsparse}{11.3}{11.5}{\numsparseunit} & 
 						\BoldUndLineLARGDBU{\numdense}{1.835}{1.835}{\numdenseunit}  &  
 						\BoldUndLineSMALLESTKK{\MtimeV}{42.5}{41}     
 					} 
 					{ \ifthenelse{\value{DTLrowi} < 5}{
 							& \RefC{\note}{\re}\fnote  & 	
 							\BoldUndLineLARGDB{\tracklen}{5.14}{5.16} &  
 							\BoldUndLineSMALLESTThree{\reprojerr}{0.9035}{0.90}     &  
 							\BoldUndLineLARGSingU{\numsparse}{11.3}{11.5}{\numsparseunit} & 
 							\BoldUndLineLARGDBU{\numdense}{1.835}{1.835}{\numdenseunit}  &  
 							\BoldUndLineSMALLESTKK{\MtimeV}{42.5}{41}     
 						}
 						{&  \Ours{\note} & 	
 							\BoldUndLineLARGDB{\tracklen}{5.14}{5.16} &  
 							\BoldUndLineSMALLESTThree{\reprojerr}{0.9035}{0.90}   &  
 							\BoldUndLineLARGSingU{\numsparse}{11.3}{11.5}{\numsparseunit} & 
 							\BoldUndLineLARGDBU{\numdense}{1.835}{1.835}{\numdenseunit}  &  
 							\BoldUndLineSMALLESTKK{\MtimeV}{42.5}{41}     
 					}   }
 				} 
 				\DTLforeach{SfMSupp:SouthBuilding}{  \numreg=num_reg_images,    \numsparse=num_sparse_points_round,  \numsparseunit=num_sparse_points_unit, \tracklen=mean_track_length, \reprojerr=mean_reproj_error,\numobs=num_observations_round,  \numdense=num_dense_points_round, \MtimeV=MatchingRunTime,  \numdenseunit=num_dense_points_unit,  \numdenseunit=num_dense_points_unit,   \note=note, \re=ref, \fnote=fnote}%
 				{\ifthenelse{\value{DTLrowi}=1}{\tabularnewline \toprule }{    \ifthenelse{\value{DTLrowi}=5}{\tabularnewline \cdashlinelr{2-7} }{
 							\ifthenelse{\value{DTLrowi}=3 \OR  \value{DTLrowi}=8}{\tabularnewline \cdashlinelr{2-7} }{ \tabularnewline } 	}  } 
 					\ifthenelse{\value{DTLrowi}=1}
 					{  	\multirow{2}{11mm}{ \bfseries South-Building 128 images }  &   \RefC{\note}{\re}\fnote &  
 						\BoldUndLineLARGDB{\tracklen}{8.31}{8.32} &  
 						\BoldUndLineSMALLESTThree{\reprojerr}{0.8365}{0.823}   &  
 						\BoldUndLineLARGSingU{\numsparse}{114.7}{132}{\numsparseunit} & 
 						\BoldUndLineLARGDBU{\numdense}{12.44}{12.52}{\numdenseunit}  &  
 						\BoldUndLineSMALLESTKK{\MtimeV}{1450}{1360}
 					} 
 					{ \ifthenelse{\value{DTLrowi} < 5}{
 							& \RefC{\note}{\re}\fnote  & 	
 							\BoldUndLineLARGDB{\tracklen}{8.31}{8.32} &  
 							\BoldUndLineSMALLESTThree{\reprojerr}{0.8365}{0.823}   &  
 							\BoldUndLineLARGSingU{\numsparse}{114.7}{132}{\numsparseunit} & 
 							\BoldUndLineLARGDBU{\numdense}{12.44}{12.52}{\numdenseunit}  &  
 							\BoldUndLineSMALLESTKK{\MtimeV}{1450}{1360}
 						}
 						{&  \Ours{\note} & 	
 							\BoldUndLineLARGDB{\tracklen}{8.31}{8.32} &  
 							\BoldUndLineSMALLESTThree{\reprojerr}{0.8365}{0.823}    &  
 							\BoldUndLineLARGSingU{\numsparse}{114.7}{132}{\numsparseunit} & 
 							\BoldUndLineLARGDBU{\numdense}{12.44}{12.52}{\numdenseunit}  &  
 							\BoldUndLineSMALLESTKK{\MtimeV}{1450}{1360}
 					}   }
 				}
 				\\ 	\bottomrule
 			\end{tabular}  
 		}
 		\begin{tablenotes}
 			\item[\textdagger] Superglue with its official setting (Sinkhorn iter. = 100).  
 			\item[\S] SGMNet with its official setting (Sinkhorn iter. = 100, num. seeds =128).
 		\end{tablenotes} 
 	\end{threeparttable}	 
 	
 \end{table}

 \begin{table}[t]  
 	\caption{Medium-size ETH. Our method versus \MNN+Lowe's Thresholding, AdaLAM,  Superglue-10, and SGMNet-10.}
 	\label{Supp:Table:3DRecon-Medium}	  
 	\vspace{-0.3cm}
 	\renewrobustcmd{\bfseries}{\fontseries{b}\selectfont} 
 	\sisetup{detect-weight,mode=text,round-mode=places, round-precision=2}
 	{\fontsize{6.8}{7.0}\selectfont 
 		\setlength\tabcolsep{1pt} 
 		\setlength{\extrarowheight}{2pt}
 		\begin{tabular}{p{1cm}    l@{\hskip2pt}   P{0.75cm} P{0.8cm} P{0.7cm} P{0.9cm}    P{1.4cm}  }  
 			\toprule   \bfseries Datasets & \bfseries Methods    &  \bfseries \textit{\Track}  & \bfseries \textit{\Reproj}   &  \bfseries \RegImg    &  \bfseries \textit{Dense Points} &  \bfseries \textit{Match. Time (H:M:S)}
 			\DTLforeach{SfMSupp:Madrid}{  \numreg=num_reg_images,   \numsparse=num_sparse_points_round,  \numsparseunit=num_sparse_points_unit, \tracklen=mean_track_length, \reprojerr=mean_reproj_error, \numdense=num_dense_points_round, \numdenseunit=num_dense_points_unit,\numdense=num_dense_points_round,  \numdenseunit=num_dense_points_unit,  \Mtime=MatchingTime, \MtimeV=MatchingRunTime,  \note=note, \re=ref, \fnote=fnote}%
 			{  	 \ifthenelse{\value{DTLrowi}=1}{\tabularnewline \midrule  }{\ifthenelse{\value{DTLrowi}=5  \OR \value{DTLrowi}=2 \OR \value{DTLrowi}=3  \OR  \value{DTLrowi}=8}{\tabularnewline \cdashlinelr{2-7} }{\tabularnewline}} 
 				\ifthenelse{\value{DTLrowi}=1}
 				{  \multirow{2}{10mm}{  \bfseries Madrid Metropolis 1344 images  }  & \RefC{\note}{\re}\fnote & 
 					\BoldUndLineLARGDB{\tracklen}{8.33}{8.6} &  
 					\BoldUndLineSMALLESTThree{\reprojerr}{1.10}{1.070}   & 
 					\BoldUndLineLARGINT{\numreg}{563}{750}   & 
 					\BoldUndLineLARGDBU{\numdense}{3.3}{3.4}{\numdenseunit}    & 
 					\BoldTextSmall{\MtimeV}{4500}{0400}{\Mtime} 
 				} 
 				{  
 					\ifthenelse{\value{DTLrowi} < 5}
 					{ &  \RefC{\note}{\re}\fnote & 	
 						\BoldUndLineLARGDB{\tracklen}{8.33}{8.6} & 
 						\BoldUndLineSMALLESTThree{\reprojerr}{1.110}{1.070}   & 
 						\BoldUndLineLARGINT{\numreg}{563}{750}   & 
 						\BoldUndLineLARGDBU{\numdense}{3.03}{3.3}{\numdenseunit}    & 
 						\BoldTextSmall{\MtimeV}{4500}{0400}{\Mtime} 
 					}
 					{   
 						& \Ours{\note}  & 	
 						\BoldUndLineLARGDB{\tracklen}{8.33}{8.6} & 
 						\BoldUndLineSMALLESTThree{\reprojerr}{1.11}{1.070}   & 
 						\BoldUndLineLARGINT{\numreg}{563}{750}   & 
 						\BoldUndLineLARGDBU{\numdense}{3.03}{3.3}{\numdenseunit}    & 
 						\BoldTextSmall{\MtimeV}{4500}{0400}{\Mtime} 
 					}
 				}
 			}       
 			\DTLforeach{SfMSupp:Gendark}{  \numreg=num_reg_images,    \numsparse=num_sparse_points_round,  \numsparseunit=num_sparse_points_unit, \tracklen=mean_track_length, \reprojerr=mean_reproj_error,\numobs=num_observations_round,  \numdense=num_dense_points_round,  \numdenseunit=num_dense_points_unit,  \numdenseunit=num_dense_points_unit, \Mtime=MatchingTime, \MtimeV=MatchingRunTime,
 				\note=note, \re=ref, \fnote=fnote}%
 			{\ifthenelse{\value{DTLrowi}=1}{\tabularnewline \toprule}{\ifthenelse{\value{DTLrowi}=5  \OR \value{DTLrowi}=2 \OR \value{DTLrowi}=3  \OR  \value{DTLrowi}=8}{\tabularnewline \cdashlinelr{2-7}  }{\tabularnewline}}  
 				\ifthenelse{\value{DTLrowi}=1}
 				{  	\multirow{2}{10mm}{ \bfseries Gendarmenmarkt 1463 images}  &   \RefC{\note}{\re}\fnote &  
 					\BoldUndLineLARGDB{\tracklen}{8.0}{8.3} &  
 					\BoldUndLineSMALLESTThree{\reprojerr}{1.105}{1.10}   &  
 					\BoldUndLineLARGINT{\numreg}{1040}{1100}   & 
 					\BoldUndLineLARGDBU{\numdense}{7.05}{7.2}{\numdenseunit}  &  
 					\BoldTextSmall{\MtimeV}{7200}{0400}{\Mtime} 
 				} 
 				{ \ifthenelse{\value{DTLrowi} < 5}{
 						& \RefC{\note}{\re}\fnote  & 	
 						\BoldUndLineLARGDB{\tracklen}{8.0}{8.3} &  
 						\BoldUndLineSMALLESTThree{\reprojerr}{1.105}{1.10}   &  
 						\BoldUndLineLARGINT{\numreg}{1040}{1100}   & 
 						\BoldUndLineLARGDBU{\numdense}{7.05}{7.2}{\numdenseunit}  &  
 						\BoldTextSmall{\MtimeV}{7200}{0400}{\Mtime} 
 					}
 					{&  \Ours{\note} & 	 
 						\BoldUndLineLARGDB{\tracklen}{8.0}{8.3} &  
 						\BoldUndLineSMALLESTThree{\reprojerr}{1.105}{1.10}   &  
 						\BoldUndLineLARGINT{\numreg}{1040}{1100}   & 
 						\BoldUndLineLARGDBU{\numdense}{7.05}{7.2}{\numdenseunit}  &  
 						\BoldTextSmall{\MtimeV}{7200}{0400}{\Mtime} 
 				}   }
 			} 
 			\DTLforeach{SfMSupp:Tower}{  \numreg=num_reg_images,    \numsparse=num_sparse_points_round,  \numsparseunit=num_sparse_points_unit, \tracklen=mean_track_length, \reprojerr=mean_reproj_error,\numobs=num_observations_round,  \numdense=num_dense_points_round, \Mtime=MatchingTime, \MtimeV=MatchingRunTime,  \numdenseunit=num_dense_points_unit,  \numdenseunit=num_dense_points_unit,    \note=note, \re=ref, \fnote=fnote}%
 			{\ifthenelse{\value{DTLrowi}=1}{\tabularnewline \toprule}{\ifthenelse{\value{DTLrowi}=5  \OR \value{DTLrowi}=2 \OR \value{DTLrowi}=3  \OR  \value{DTLrowi}=8}{\tabularnewline \cdashlinelr{2-7}  }{\tabularnewline}}  
 				\ifthenelse{\value{DTLrowi}=1}
 				{  	\multirow{2}{10mm}{ \bfseries Tower of London  1576 images}  &   \RefC{\note}{\re}\fnote &  
 					\BoldUndLineLARGDB{\tracklen}{8.50}{8.6} &  
 					\BoldUndLineSMALLESTThree{\reprojerr}{1.039}{1.02}   &  
 					\BoldUndLineLARGINT{\numreg}{780}{930}   & 
 					\BoldUndLineLARGDBU{\numdense}{5.52}{5.9}{\numdenseunit}  &  
 					\BoldTextSmall{\MtimeV}{6000}{0400}{\Mtime} 
 				} 
 				{ \ifthenelse{\value{DTLrowi} < 5}{
 						& \RefC{\note}{\re}\fnote  & 	
 						\BoldUndLineLARGDB{\tracklen}{8.50}{8.6} &  
 						\BoldUndLineSMALLESTThree{\reprojerr}{1.039}{1.02}   &  
 						\BoldUndLineLARGINT{\numreg}{780}{930}   & 
 						\BoldUndLineLARGDBU{\numdense}{5.52}{5.9}{\numdenseunit}  &  
 						\BoldTextSmall{\MtimeV}{6000}{0400}{\Mtime} 
 					}
 					{&  \Ours{\note} & 	
 						\BoldUndLineLARGDB{\tracklen}{8.50}{8.6} &  
 						\BoldUndLineSMALLESTThree{\reprojerr}{1.039}{1.02}   &  
 						\BoldUndLineLARGINT{\numreg}{780}{930}   & 
 						\BoldUndLineLARGDBU{\numdense}{5.52}{5.9}{\numdenseunit}  &  
 						\BoldTextSmall{\MtimeV}{6000}{0400}{\Mtime} 
 				}   }
 			}
 			\\ 	\bottomrule
 		\end{tabular}  
 	} 	 
 \end{table} 

 \smallskip
\noindent\textbf{Visual Results} 
\Fig{tab:Supp:3DRecon} and \Fig{tab:Supp:3DReconMedium} provides the visual results of the 3D reconstruction on the small- and the medium-size datasets. From \Fig{tab:Supp:3DRecon}, our method (\ours{1}) provides the best visual results, the lowest \Reproj,~and the longest \Track~in most cases. Here, we enlarged the 3D point clouds for Fountain and Herzjesu, and we applied the error threshold of 1.2 in all cases.  On Fountain, our method provides similar visual results to SuperGlue and SGMNet. On Herzjesu, our method can capture slightly more details. Meanwhile, our work can provide overall denser 3D reconstruction on South-Building.    From \Fig{tab:Supp:3DReconMedium}, our \ours{1} produces the 3D reconstruction with the least artifacts, the lowest \Reproj, ~and the longest \Track, in most cases.   For example, on Genkarkmentmarkt,  our \ours{1} can capture the most of landmark with accurate 3D reconstruction. Meanwhile, SuperGlue-10 produces small inaccurate 3D point clouds in front of the concert hall between the two churches. SGMNet-10 produces the inaccurate 3D point clouds of the entire concert hall building. On Madrid Metropolis, our 3D reconstruction accurately captures the landmark but is more sparse than SuperGlue-10. Meanwhile, SGMNet-10 provides the inaccurate 3D point clouds on the opposite side to the Metropolis Building. However, on Tower of London, all of the methods struggle to provide the accurate result. Our 3D reconstruction is very sparse and misses parts of the castle on the tops. Meanwhile, SuperGlue and SGMNet produce many artifacts around the castle areas. This suggests the future improvement to maintain  high accuracy.

\begin{table}[t]
	\caption{ Visual localization on Aachen Day-Night. }
	\label{Supp:Table:AachenDayNight}  
	\vspace{-0.3cm}
	\makegapedcells
	\renewrobustcmd{\bfseries}{\fontseries{b}\selectfont} 
	{\fontsize{6.8}{7.0}\selectfont  
		\setlength\tabcolsep{0.5 pt} 
		\setlength{\extrarowheight}{3pt}
		\begin{tabular} { l@{\hskip2pt} l@{\hskip2pt} P{0.6cm} P{0.6cm} P{0.7cm}  P{1.3cm} !{\vrule width 0.5pt} P{1cm} P{0.9cm} P{0.7cm} }   
			\toprule  
			& \bfseries Methods   & \bfseries	\textit{\#kpts}  & \bfseries  \textit{\#dim}   & \bfseries \textit{\#Param.} &  \bfseries \textit{Cplx.} &   \bfseries 0.25m,~\degc{2}  & \bfseries  0.5m,~\degc{5}    & \bfseries 5m,~\degc{10}  \\  
			\midrule  
			&	D2-Net~\cite{CVPR2019:D2Net}              & 19K & 512 & 15M  & -  &  74.5    &  86.7    & \bfseries \bfseries \underline{100.0}   \\ 
			& 	ASLFeat~\cite{CVPR2020:ASLFeat} v.2       &   10K &   128 & \bfseries \underline{0.4M}   & -  & \bfseries \underline{81.6}    &  87.8    & \bfseries \underline{100.0}   \\
			&  	R2D2~\cite{NIPS2019:R2D2} $N=8$  &  40K &   128 & 1.0M & -  &  76.5    & \bfseries  \underline{90.8}    &  \bfseries \underline{100.0}     \\  
			& 	SPTD2~\cite{IJCV21:SPTD2}   &  10K &   128 & - &   -  &  78.8    & \bfseries  \underline{89.3}    &  \bfseries \underline{99.0}     \\ \hline
			& \multicolumn{5}{l}{SuperPoint~\cite{CVPR2018:SuperPoint} + SOTA Filtering/Matching}    &  & & \\   \cdashlinelr{1-9}
			&  $\LshR$+ Baseline MNN & \bfseries \underline{4K} & 256 & - & - &  71.4    &  78.6    &  87.8   \\   \cdashlinelr{1-9}  
			& $\LshR$~+~OANet  ~\cite{AdaLAM2020}     &  \bfseries \underline{4K} & 256  &  4.8M  & - &  77.6   & \bfseries \underline{ 90.8}  & \bfseries \underline{100.0} \\  
			& $\LshR$~+~AdaLAM  ~\cite{AdaLAM2020}   &  \bfseries \underline{4K} & 256 & -  & - &  76.5    &  86.7    &  95.9   \\  \cdashlinelr{1-9}  
			& $\LshR$~+~SuperGlue~\cite{SuperGlue_CVPR2020}    & \bfseries \underline{4K} & 256 &  12M  & $N^2C$ &  79.6    & \bfseries \underline{ 90.8}    &  \bfseries \underline{100.0}   \\   
			&   $\LshR$~+~SGMNet~\cite{SGMNet_ICCV2021}       & \bfseries \underline{4K} & 256 &  31M  & $NKC + K^2C$ &  77.6    &    88.8    & \bfseries 99.0   \\   
			\cdashlinelr{1-9}     
			&  $\LshR$~+~\Ours{2}   & \bfseries \underline{4K} & \bfseries \underline{64}  & \bfseries  0.8M & $\approx NC'^2$  &  77.6 &  84.7    & 94.9    \\   
			&  $\LshR$~+~\Ours{0}   & \bfseries \underline{4K} & \bfseries \underline{64}  & \bfseries   0.8M & $\approx NC'^2$  &   78.6 & 86.7 & 95.9      \\    
			& $\LshR$~+~\Ours{1}   & \bfseries \underline{4K} & \bfseries \underline{64}  & \bfseries  0.8M  & $\approx N C'^2$ & \bfseries 80.6 & 86.7 & 95.9      \\  
			\cdashlinelr{1-9}    
			& $\LshR$~+~\Ours{11}   &  \bfseries \underline{4K} &    \bfseries \underline{64}  & \bfseries  0.8M  & $\approx NC^2$ &   
			77.6 & 84.7 & 92.9      \\  
			& $\LshR$~+~\Ours{1}-L  &  \bfseries \underline{4K} &   256  &    12M  & $\approx NC^2$ &   78.6 & 87.8 & 96.9      \\  \bottomrule
		\end{tabular}
	}  
\end{table}
\subsection{Visual Localization}
This section we provide the additional numerical results of our \ours{11} on visual localization in~\Table{Supp:Table:AachenDayNight}. \Ours{11} offers lower performance than the other configurations of our work---\ours{2}, \ours{0},  and \ours{1}. This is because \ours{11} tends to provide the lower number of matches as shown in Figure~\ref{Supp:Fig:Attention} and Figure~\ref{Supp:Fig:HPatch}. On the other hand, our work such as our \ours{1}  provides the better localization accuracy as it  offers the higher number of accurate matches.  Although our \ours{1} does not match as much as the SOTAs, SuperGlue and SGMNet, it offers better 3D reconstruction as shown in \Table{Supp:Table:3DRecon-Small} and \Table{Supp:Table:3DRecon-Medium}, leading to the better accuracy than the SOTAs.     
\begin{figure*}[h!]
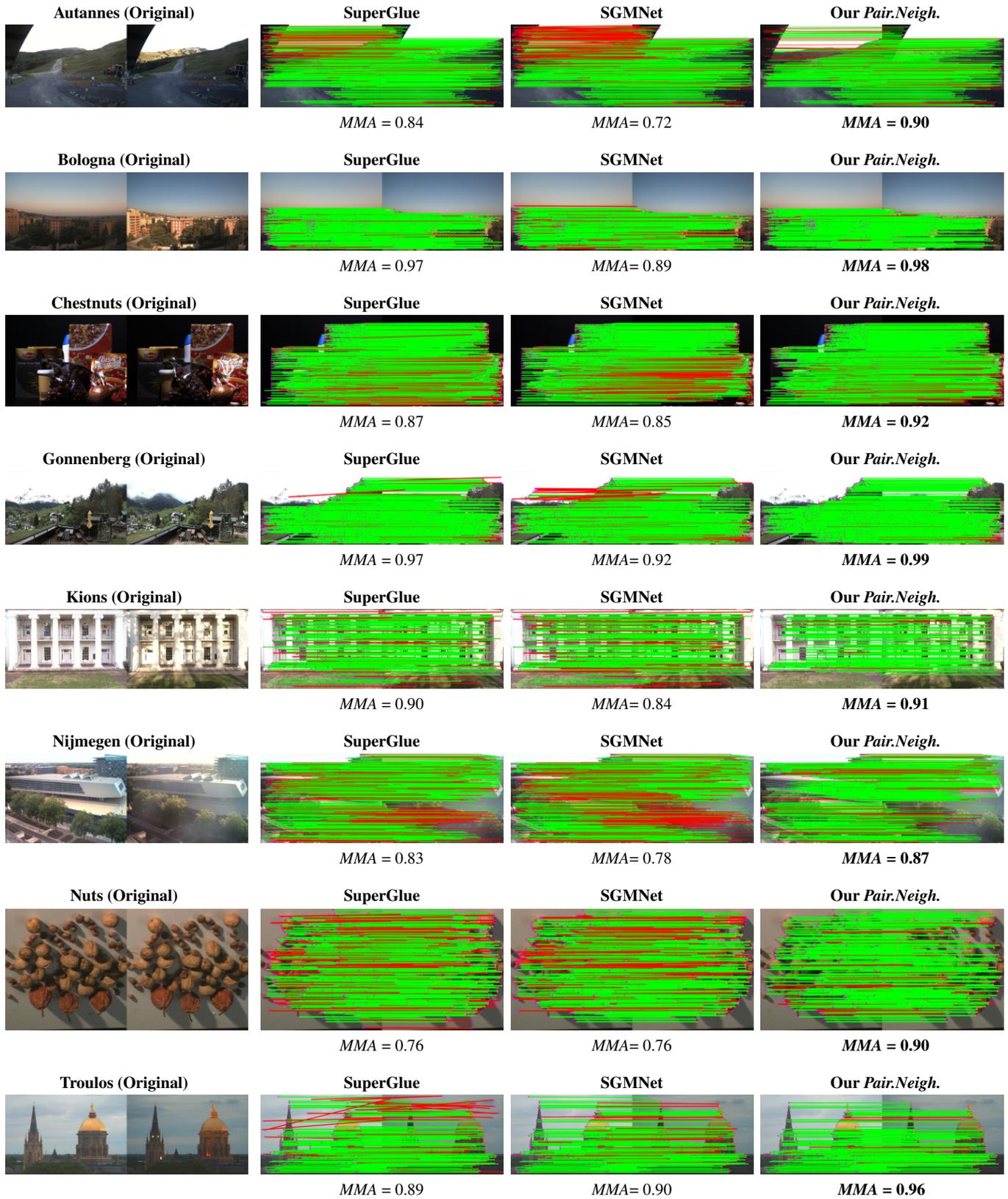
   	
	\renewrobustcmd{\bfseries}{\fontseries{b}\selectfont}   
	\sisetup{detect-weight,mode=text,round-mode=places, round-precision=2}
	{\fontsize{8.5}{8}\selectfont   
		{  \setlength\tabcolsep{2pt}  	\renewcommand{\arraystretch}{1.2} 
			\begin{tabular}{ c@{\hskip5pt} c@{\hskip2pt}  c@{\hskip2pt}   c@{\hskip2pt}    }  %
				\DTLforeach{App:VisualResultIup}{\Image=Image,\sg=SuperGlue,\sgm=SGMNet,\ours=Ours,\NoteV=NoteV,\NoteC=NoteC}{ 
					 							
					\ifthenelse{\value{DTLrowi}=1}{\tabularnewline  }{\tabularnewline}   
					\ifthenelse{\value{DTLrowi}=1 \OR \value{DTLrowi}=4\OR \value{DTLrowi}=7 \OR 	\value{DTLrowi}=10 \OR    \value{DTLrowi}=13 \OR  	\value{DTLrowi}=16 \OR 	\value{DTLrowi}=19 \OR  \value{DTLrowi}=22 }{ 
						\bfseries \Image 	&  \bfseries \sg  &  \bfseries  \sgm  & \bfseries \Ours{1} 
						
					}{ 
						\ifthenelse{ \value{DTLrowi}=2 \OR \value{DTLrowi}=5 \OR  	\value{DTLrowi}=8 \OR 	\value{DTLrowi}=11 \OR  \value{DTLrowi}=14    \OR  \value{DTLrowi}=17 \OR 	\value{DTLrowi}=20 \OR  \value{DTLrowi}=23  }{   
							\includegraphics [width=0.25\textwidth, trim={0cm 0 0.0cm 0.0cm},clip] {\Image}	& 	\includegraphics [width=0.25\textwidth, trim={0cm 0 0.0cm 0.0cm},clip] {\sg} & 
							\includegraphics [width=0.25\textwidth, trim={0cm 0 0.0cm 0.0cm},clip] {\sgm} & 
							\includegraphics [width=0.25\textwidth, trim={0cm 0 0.0cm 0.0cm},clip] {\ours}  }
						{  &  \textit{MMA} = \numDB{\sg} &  \textit{MMA}= \numDB{\sgm} &  \bfseries \textit{MMA} = \numDB{\ours}   } 
						
						\ifthenelse{ \value{DTLrowi}=3 \OR \value{DTLrowi}=6 \OR  	\value{DTLrowi}=9 \OR 	\value{DTLrowi}=12 \OR  \value{DTLrowi}=15    \OR  \value{DTLrowi}=18 \OR 	\value{DTLrowi}=21    }{
							\tabularnewline	 }{}
					}   
				
				}
			\end{tabular} 
		}  	  
	}
	\caption{Image matching against illumination changes on HPatches by SuperGlue, SGMNet, and our \textit{Pair.Neigh}.}	 
	\label{tab:Supp:Vis1}    
\end{figure*}

\begin{figure*}[b]
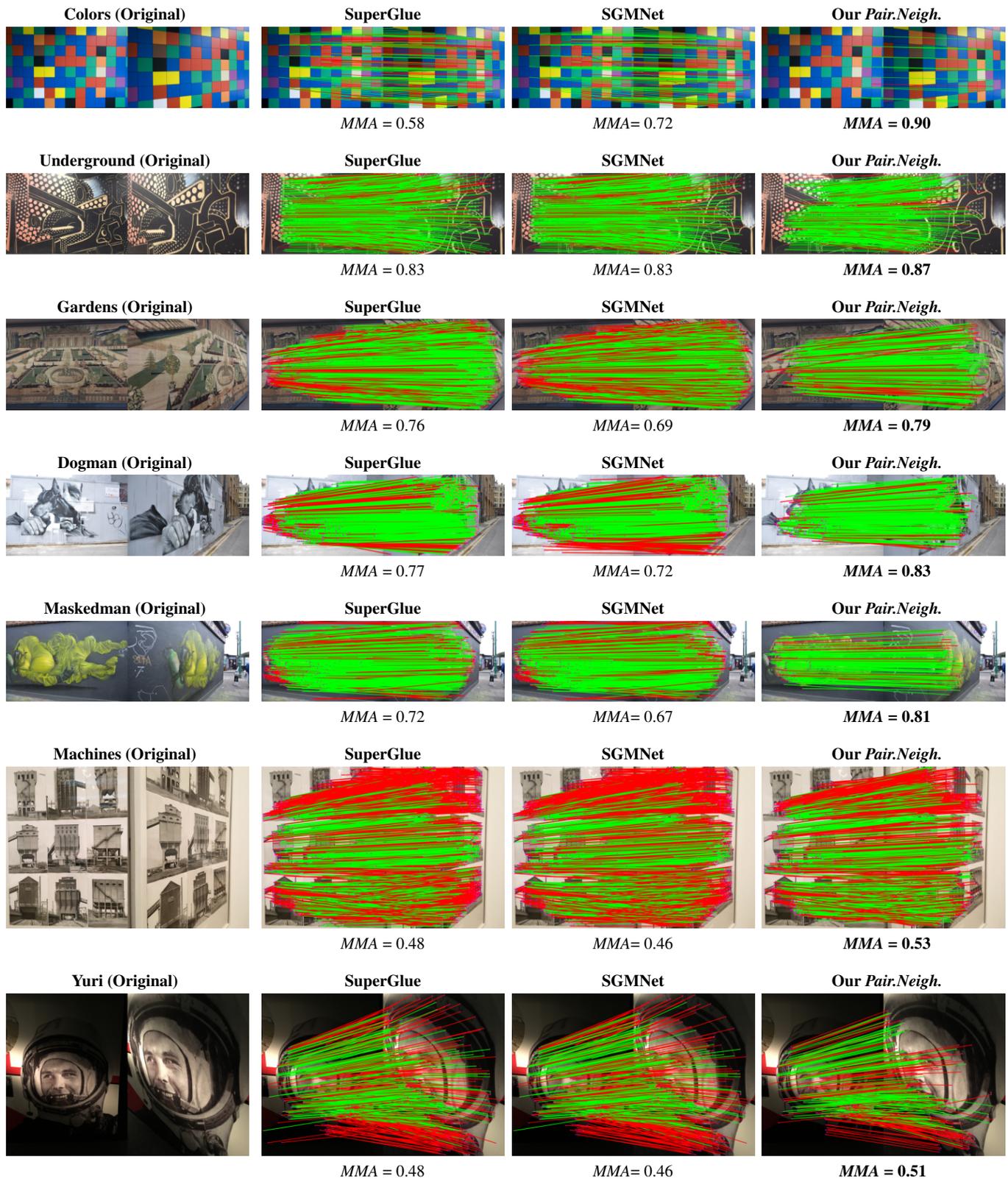
   	
	\renewrobustcmd{\bfseries}{\fontseries{b}\selectfont}   
	\sisetup{detect-weight,mode=text,round-mode=places, round-precision=2}
	{\fontsize{8.5}{8}\selectfont   
		{  \setlength\tabcolsep{2pt}  	\renewcommand{\arraystretch}{1.2} 
			\begin{tabular}{ c@{\hskip5pt} c@{\hskip2pt}  c@{\hskip2pt}   c@{\hskip2pt}    }  %
				\DTLforeach{App:VisualResultVup}{\Image=Image,\sg=SuperGlue,\sgm=SGMNet,\ours=Ours,\NoteV=NoteV,\NoteC=NoteC}{   
					\ifthenelse{\value{DTLrowi}=1}{\tabularnewline  }{\tabularnewline}   
					\ifthenelse{\value{DTLrowi}=1 \OR \value{DTLrowi}=4\OR \value{DTLrowi}=7 \OR 	\value{DTLrowi}=10 \OR    \value{DTLrowi}=13 \OR  	\value{DTLrowi}=16 \OR 	\value{DTLrowi}=19 \OR  \value{DTLrowi}=22 }{ 
						\bfseries \Image 	&  \bfseries \sg  &  \bfseries  \sgm  & \bfseries \Ours{1} 
						
					}{ 
						\ifthenelse{ \value{DTLrowi}=2 \OR \value{DTLrowi}=5 \OR  	\value{DTLrowi}=8 \OR 	\value{DTLrowi}=11 \OR  \value{DTLrowi}=14    \OR  \value{DTLrowi}=17 \OR 	\value{DTLrowi}=20 \OR  \value{DTLrowi}=23  }{   
							\includegraphics [width=0.25\textwidth, trim={0cm 0 0.0cm 0.0cm},clip] {\Image}	& 	\includegraphics [width=0.25\textwidth, trim={0cm 0 0.0cm 0.0cm},clip] {\sg} & 
							\includegraphics [width=0.25\textwidth, trim={0cm 0 0.0cm 0.0cm},clip] {\sgm} & 
							\includegraphics [width=0.25\textwidth, trim={0cm 0 0.0cm 0.0cm},clip] {\ours}  }
						{  &  \textit{MMA} = \numDB{\sg} &  \textit{MMA}= \numDB{\sgm} &  \bfseries \textit{MMA} = \numDB{\ours}   } 
						
						\ifthenelse{ \value{DTLrowi}=3 \OR \value{DTLrowi}=6 \OR  	\value{DTLrowi}=9 \OR 	\value{DTLrowi}=12 \OR  \value{DTLrowi}=15    \OR  \value{DTLrowi}=18   }{
							\tabularnewline	 }{}
					}  
				}
			\end{tabular} 
		}  	  
	}
	\caption{Image matching against viewpoint changes on HPatches by SuperGlue, SGMNet, and Our \textit{Pair.Neigh}.}	 
	\label{tab:Supp:Vis3}    
\end{figure*}

\begin{figure*}[h]  
	\renewrobustcmd{\bfseries}{\fontseries{b}\selectfont}   
	\sisetup{detect-weight,mode=text,round-mode=places, round-precision=2}
	{\fontsize{9}{9}\selectfont   
		{  \setlength\tabcolsep{2pt}  	\renewcommand{\arraystretch}{1.2} 
			\begin{tabular}{  c@{\hskip5pt} c@{\hskip5pt} c@{\hskip5pt}}     
				\multicolumn{3}{c}{\textbf{\textit{Fountain}}}
				\\   
				\textbf{SuperGlue}   & \textbf{SGMNet} & \textbf{\Ours{1}}	
				\\
				\begin{minipage}[b]{0.28\textwidth} 
					\includegraphics [width=1\textwidth, trim={7cm 4.5cm 7cm 4.5cm},clip]{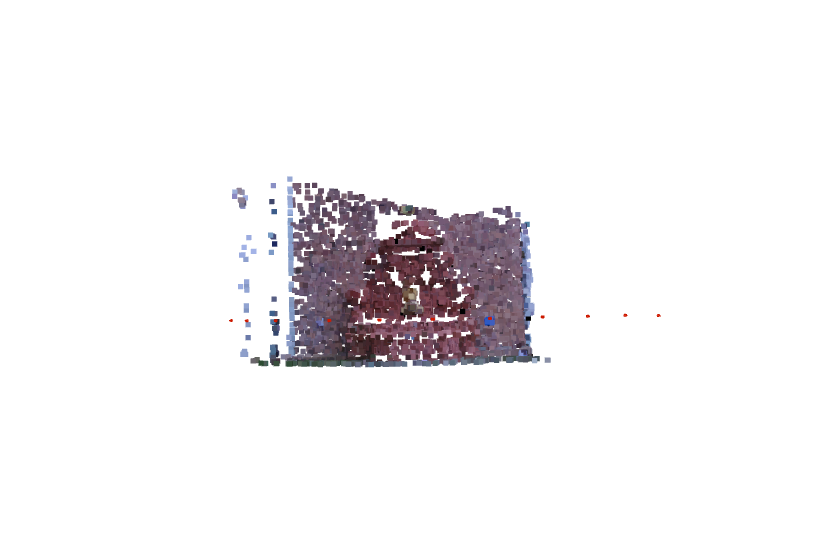} 
				\end{minipage} 
				&
				\begin{minipage}[b]{0.28\textwidth}  
					\includegraphics [width=1\textwidth, trim={7cm 4.5cm 7cm 4.5cm},clip]{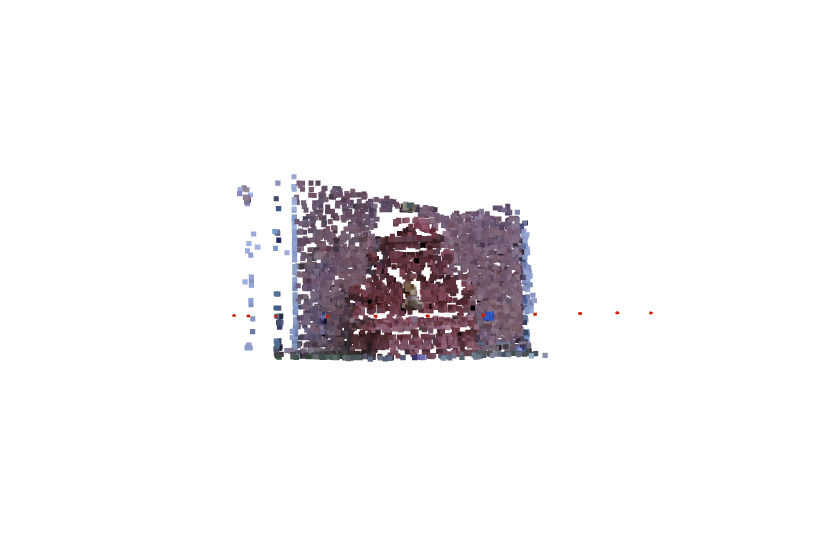} 
				\end{minipage}
				&
				\begin{minipage}[b]{0.28\textwidth} 
					\includegraphics [width=1.0\textwidth, trim={7cm 4.5cm 7cm 4.5cm},clip]{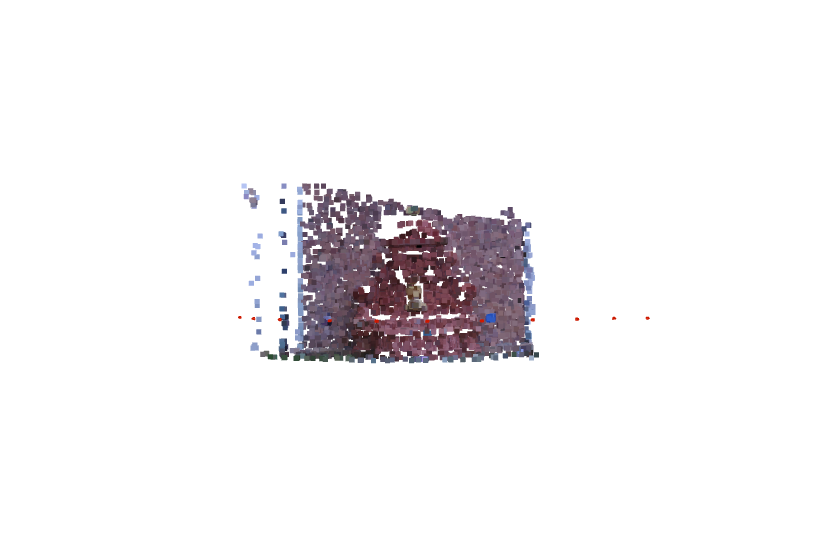} 
				\end{minipage}  
				\\
				\Reproj= 0.961, \Track= 5.14 &  \Reproj= 0.96, \Track= 4.92 & \bfseries \Reproj= 0.90, \Track= 5.14  
				\\  
				\\
				\multicolumn{3}{c}{\textbf{\textit{Herzjesu}}}
				\\   
				\textbf{SuperGlue}   & \textbf{SGMNet} & \textbf{\Ours{1}}	
				\\  
				\begin{minipage}[b]{0.20\textwidth} 
					\includegraphics [width=1\textwidth, trim={5.0cm 4.75cm 5cm 3.5cm},clip]{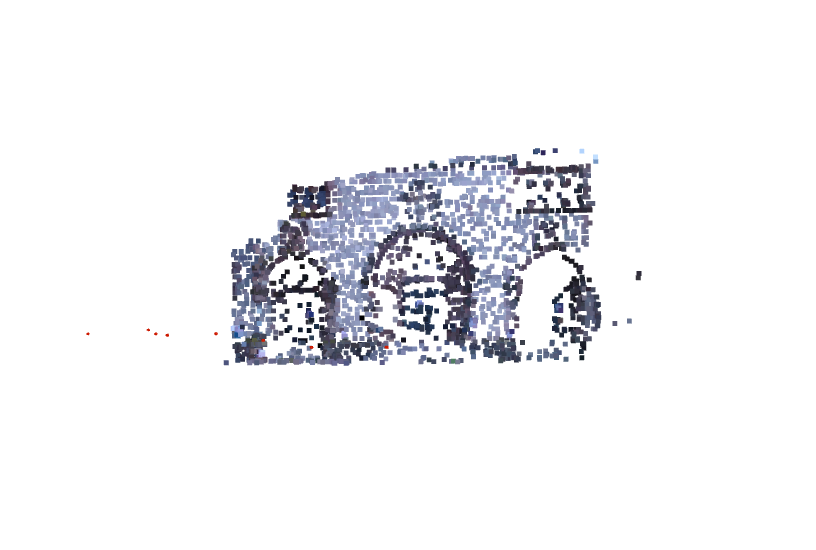} 
				\end{minipage} 
				&
				\begin{minipage}[b]{0.20\textwidth}  
					\includegraphics [width=1\textwidth, trim={5.0cm 4.75cm 5cm 3.5cm},clip]{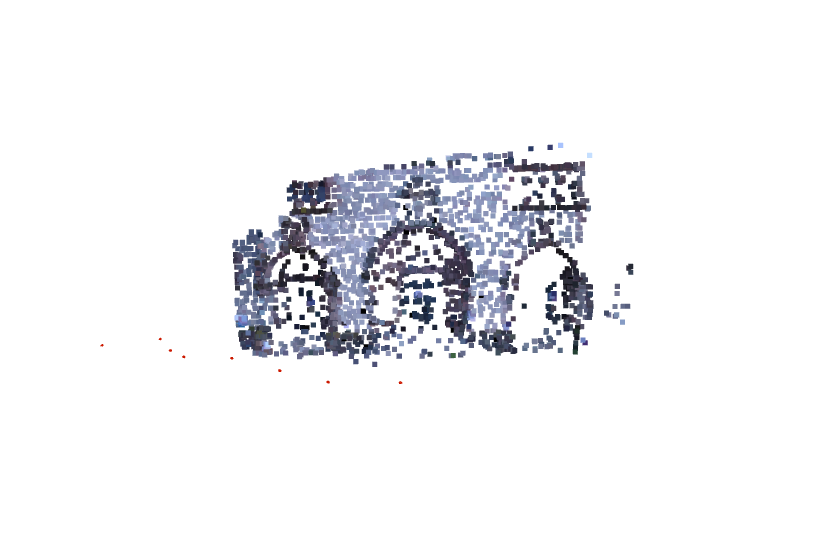} 
				\end{minipage}
				&
				\begin{minipage}[b]{0.20\textwidth} 
					\includegraphics [width=1.0\textwidth, trim={5.0cm 4.75cm 5cm 3.5cm},clip]{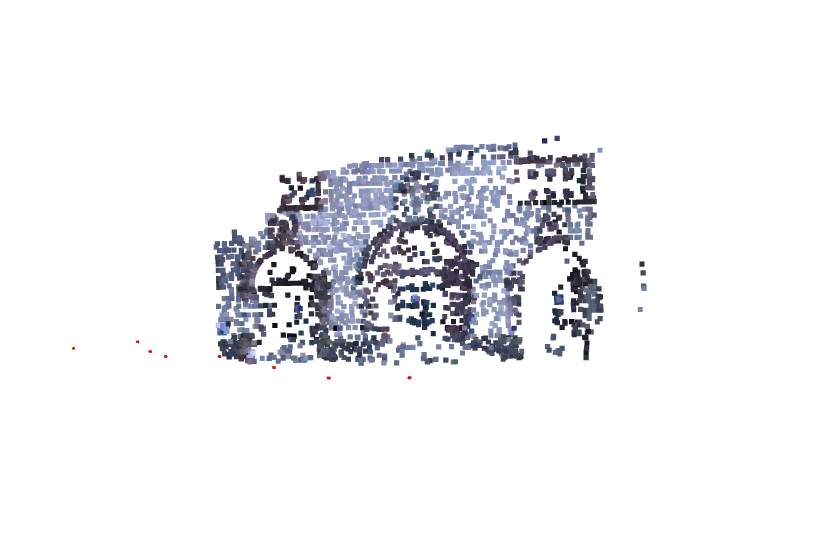} 
				\end{minipage}  
				\\
				\begin{minipage}[b]{0.20\textwidth} 
					\includegraphics [width=1\textwidth, trim={8.0cm 6.75cm 9cm 4cm},clip]{pics/SfM/Herzjesu/SGsnap.png} 
				\end{minipage} 
				&
				\begin{minipage}[b]{0.20\textwidth}  
					\includegraphics [width=1\textwidth, trim={8.0cm 6.75cm 9cm 4cm},clip]{pics/SfM/Herzjesu/SGMsnap.png} 
				\end{minipage}
				&
				\begin{minipage}[b]{0.20\textwidth} 
					\includegraphics [width=1.0\textwidth, trim={7.75cm 6.75cm 9.5cm 4cm},clip]{pics/SfM/Herzjesu/Ourssnap.png} 
				\end{minipage}
				\\
				\\
				\Reproj= 0.93, \Track= 4.45 &  \Reproj= 0.96, \Track= 4.16 & \bfseries \Reproj= 0.87, \Track= 4.53  
				\\ 
				\\ 
				\\
				\multicolumn{3}{c}{\textbf{\textit{South-building}}}
				\\ 
				\textbf{SuperGlue}   & \textbf{SGMNet} & \textbf{\Ours{1}}
				\\
				\begin{minipage}[b]{0.30\textwidth} 
					\includegraphics [width=1\textwidth, trim={3cm 2cm 3cm 3.5cm},clip]{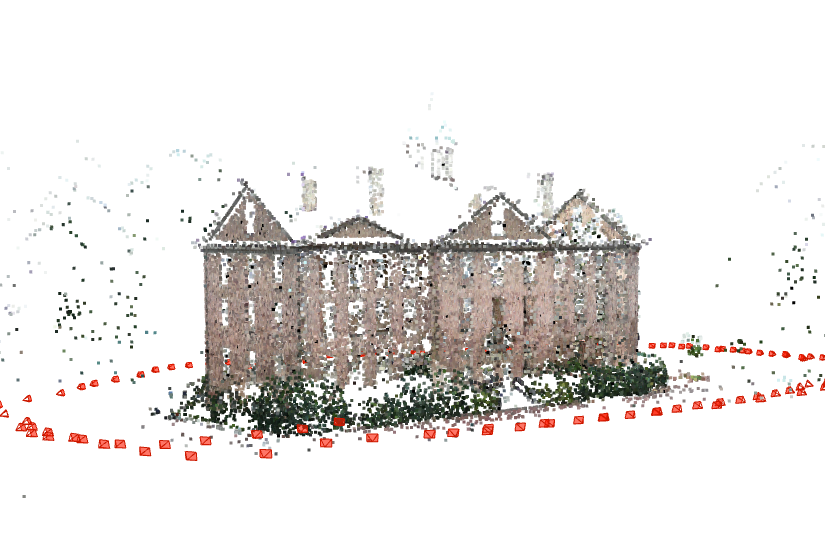} 
				\end{minipage} 
				&
				\begin{minipage}[b]{0.30\textwidth}  
					\includegraphics [width=1\textwidth, trim={3cm 2cm 3cm 3.5cm},clip]{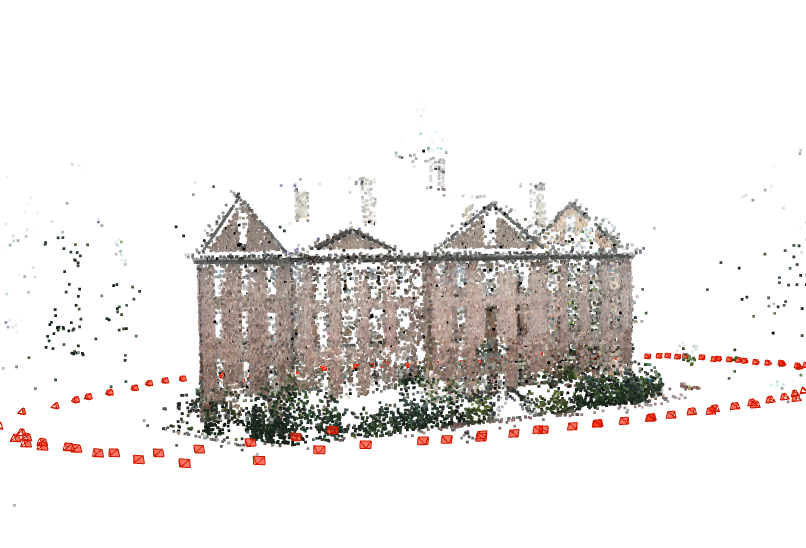} 
				\end{minipage}
				&
				\begin{minipage}[b]{0.30\textwidth} 
					\includegraphics [width=1.0\textwidth, trim={3cm 2cm 3cm 3.5cm},clip]{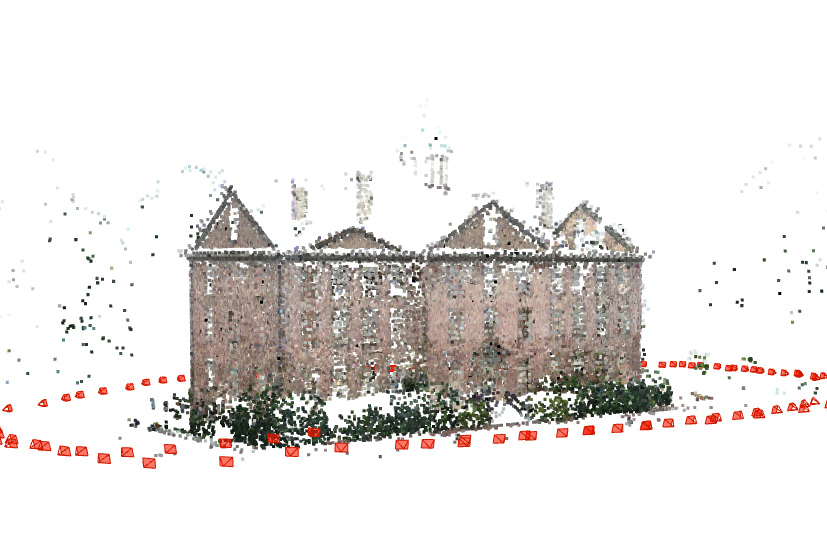} 
				\end{minipage}  
				\\
				\begin{minipage}[b]{0.30\textwidth} 
					\includegraphics [width=1\textwidth, trim={6cm 5.5cm 5.5cm 3.5cm},clip]{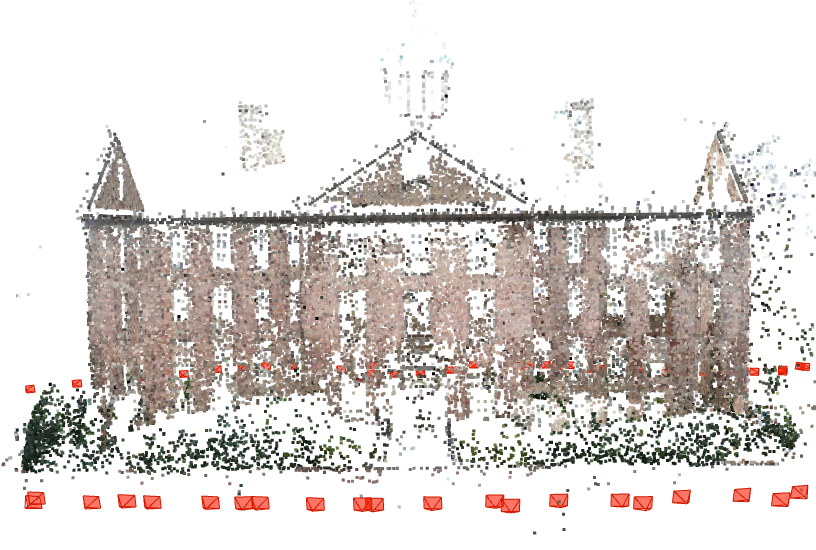} 
				\end{minipage} 
				&
				\begin{minipage}[b]{0.30\textwidth}  
					\includegraphics [width=1\textwidth, trim={6cm 5.5cm 5.5cm 3.5cm},clip]{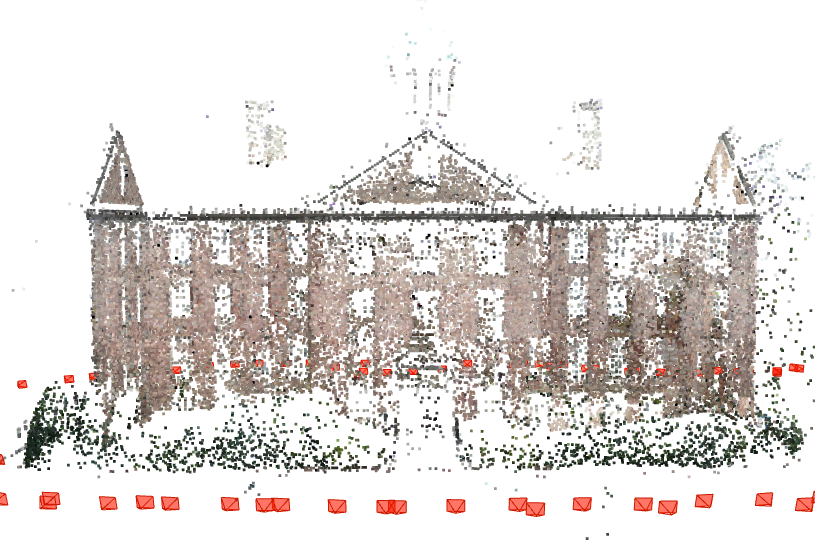} 
				\end{minipage}
				&
				\begin{minipage}[b]{0.30\textwidth} 
					\includegraphics [width=1.0\textwidth, trim={6cm 5.5cm 5.5cm 3.5cm},clip]{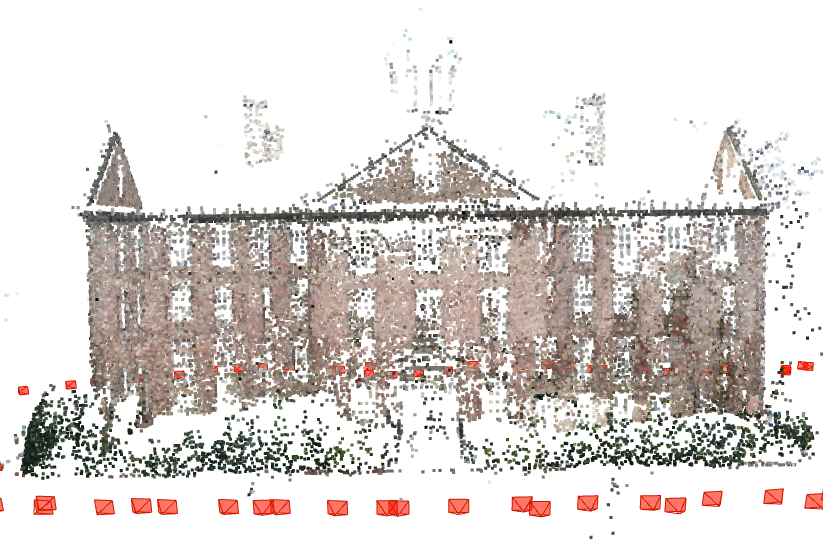} 
				\end{minipage}  
				\\
				\\
				\Reproj= 0.95, \Track= 7.90 &  \Reproj= 0.98, \Track= 6.95 & \bfseries \Reproj= 0.84, \Track= 8.27 
			\end{tabular}  
		}
	} 
	\caption{3D Reconstruction on small-size datasets by SuperGlue (left), SGMNet (middle), and Our \textit{Pair.Neigh} (right). }  
	\label{tab:Supp:3DRecon}  
\end{figure*}

\begin{figure*}[h]  
	\renewrobustcmd{\bfseries}{\fontseries{b}\selectfont}   
	\sisetup{detect-weight,mode=text,round-mode=places, round-precision=2}
	{\fontsize{9}{9}\selectfont   
		{  \setlength\tabcolsep{1pt}  	\renewcommand{\arraystretch}{1.2} 
			\begin{tabular}{    P{5.5cm} P{5.5cm} P{5.5cm}  }     
				\multicolumn{3}{c}{\textbf{\textit{Gandarkmentmarkt}}}
				\\
				\textbf{SuperGlue-10}   & \textbf{SGMNet-10} & \textbf{\Ours{1}} 	\\ 
				\begin{minipage}[b]{0.30\textwidth}  
					\includegraphics [width=1\textwidth, trim={3.5cm 2.5cm 6cm  0cm},clip]{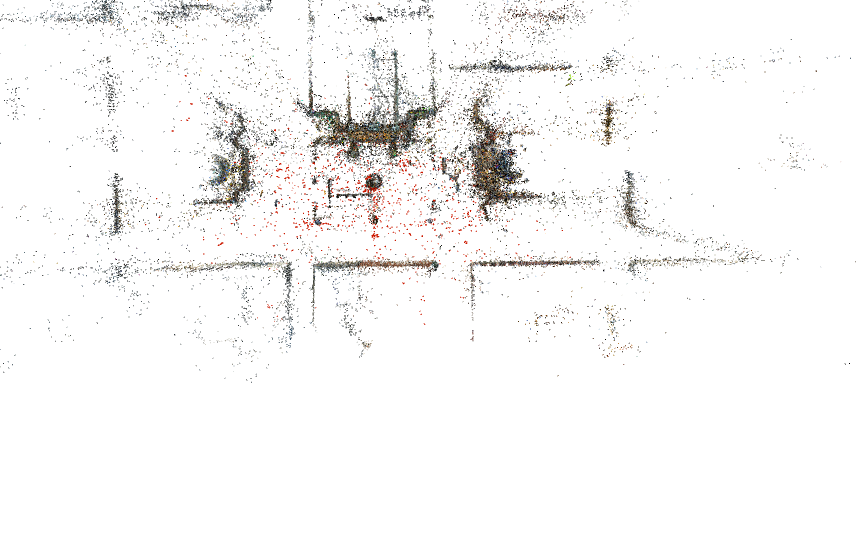}
				\end{minipage}
				&
				\begin{minipage}[b]{0.30\textwidth} 
					\includegraphics [width=1.0\textwidth, trim={3.5cm 2.5cm 5.5cm  0cm},clip]{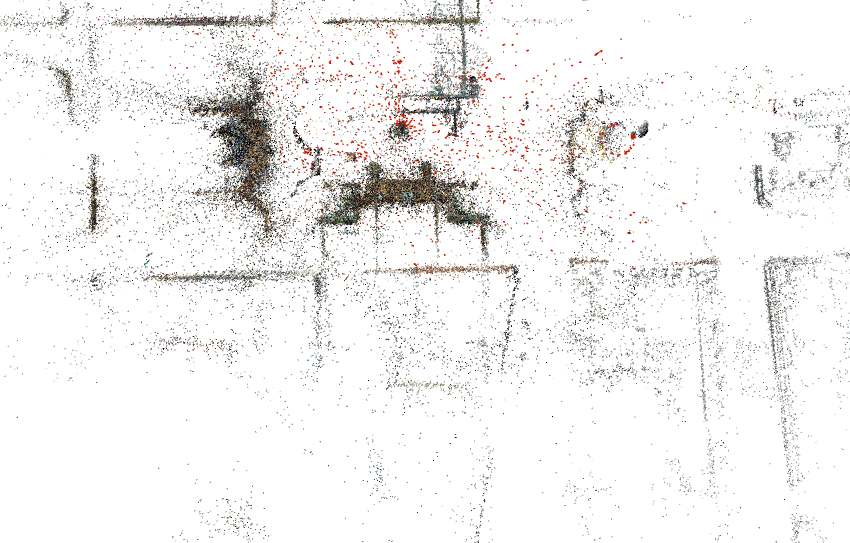}
				\end{minipage}  
				&
				\begin{minipage}[b]{0.30\textwidth} 
					\includegraphics [width=1.0\textwidth, trim={4cm 2.5cm 5.5cm  0cm},clip]{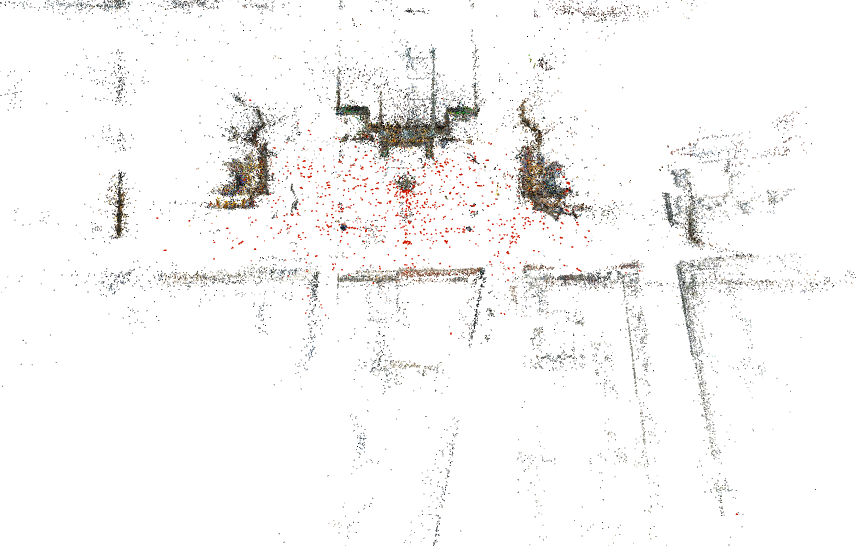}
				\end{minipage} 
				\\ 
				\begin{minipage}[b]{0.30\textwidth}  
					\includegraphics [width=1\textwidth, trim={0cm 0cm 0cm  2cm},clip]{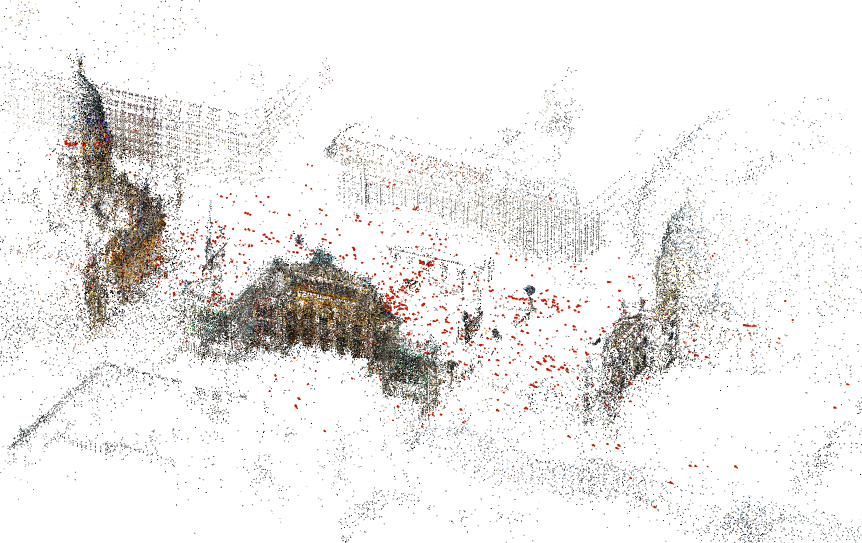}
				\end{minipage}
				&
				\begin{minipage}[b]{0.30\textwidth} 
					\includegraphics [width=1.0\textwidth, trim={0cm 0cm 0cm 2cm},clip]{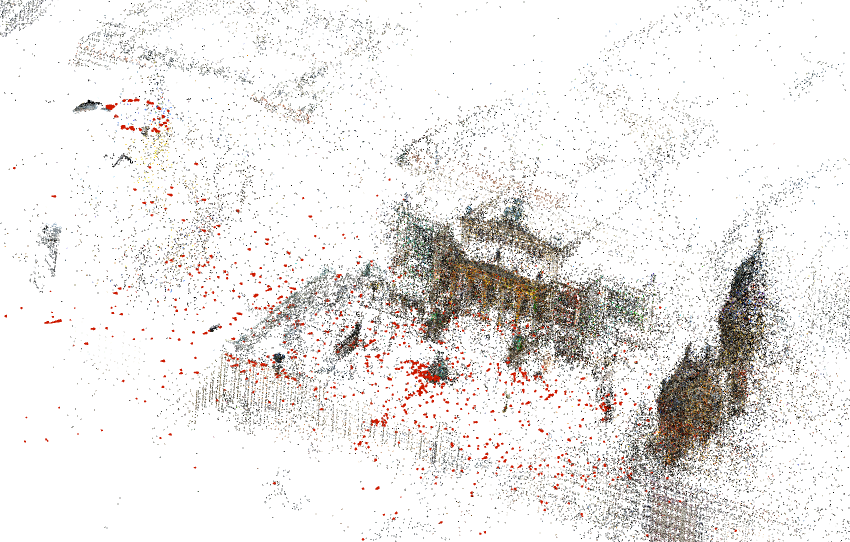}
				\end{minipage}  
				&
				\begin{minipage}[b]{0.30\textwidth} 
					\includegraphics [width=1.0\textwidth, trim={0cm 0cm 0cm  2cm},clip]{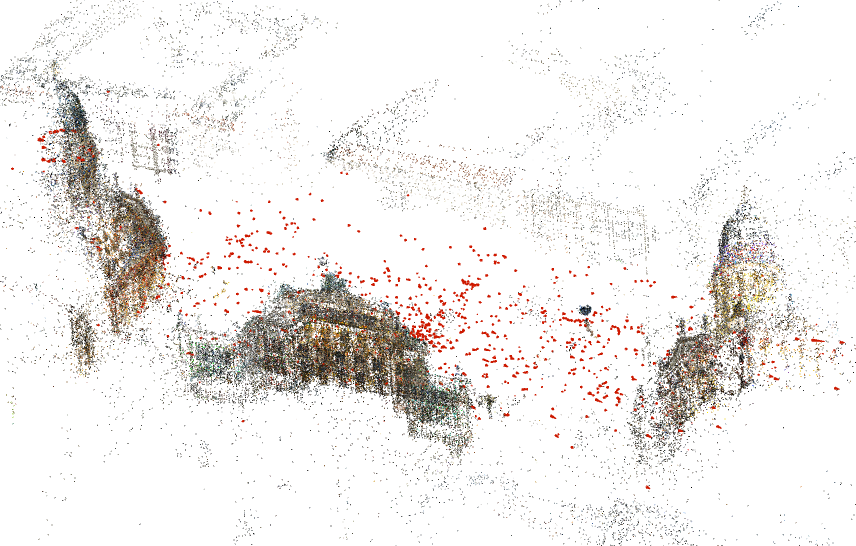}
				\end{minipage} 
				\\
				\Reproj= 1.22, \bfseries \Track= 8.05   &  \Reproj= 1.20, \Track= 7.53   & \textbf{\Reproj= 1.12},  \Track= 7.92  
				\\    
				\\
				\multicolumn{3}{c}{\textbf{\textit{Madrid Metropolis}}}
				\\
				\textbf{SuperGlue-10}   & \textbf{SGMNet-10} & \textbf{\Ours{1}}  	\\ 
				\begin{minipage}[b]{0.15\textwidth}  
					\includegraphics [width=1\textwidth, trim={4cm 1cm 7cm  5cm},clip]{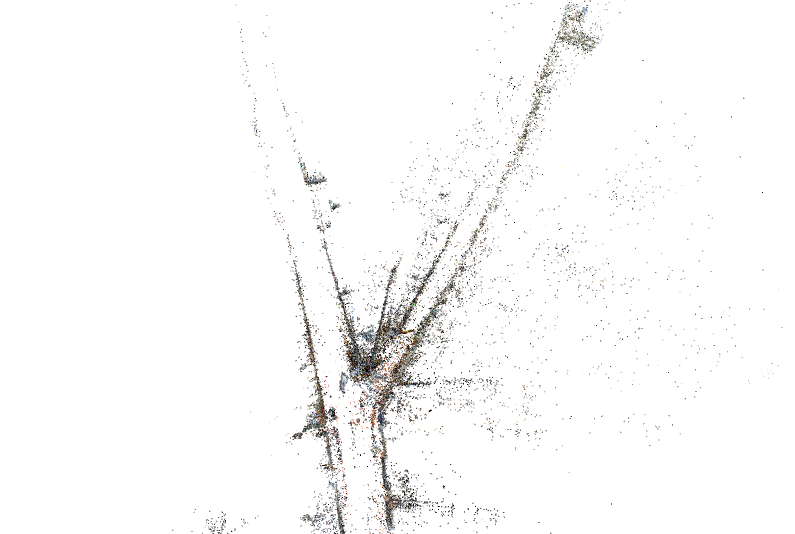}
				\end{minipage}
				&
				\begin{minipage}[b]{0.15\textwidth} 
					\includegraphics [width=1.0\textwidth, trim={4cm 1cm 7cm  5cm},clip]{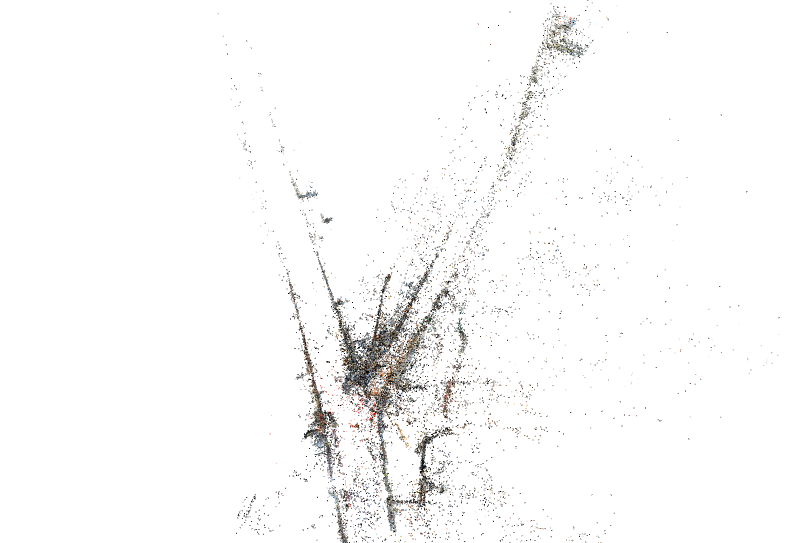}
				\end{minipage}  
				&
				\begin{minipage}[b]{0.15\textwidth} 
					\includegraphics [width=1.0\textwidth, trim={4cm 1cm 7cm  5cm},clip]{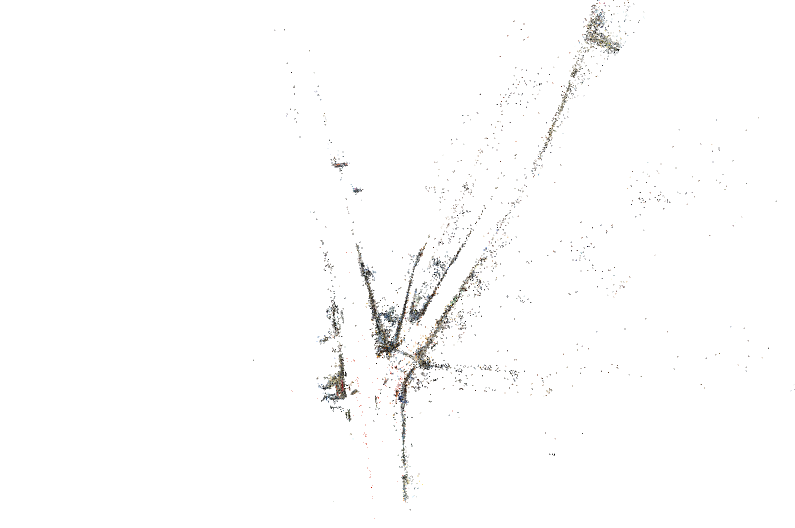}
				\end{minipage} 
				\\ 
				\begin{minipage}[b]{0.30\textwidth}  
					\includegraphics [width=1\textwidth, trim={2cm 2cm 0cm  0.5cm},clip]{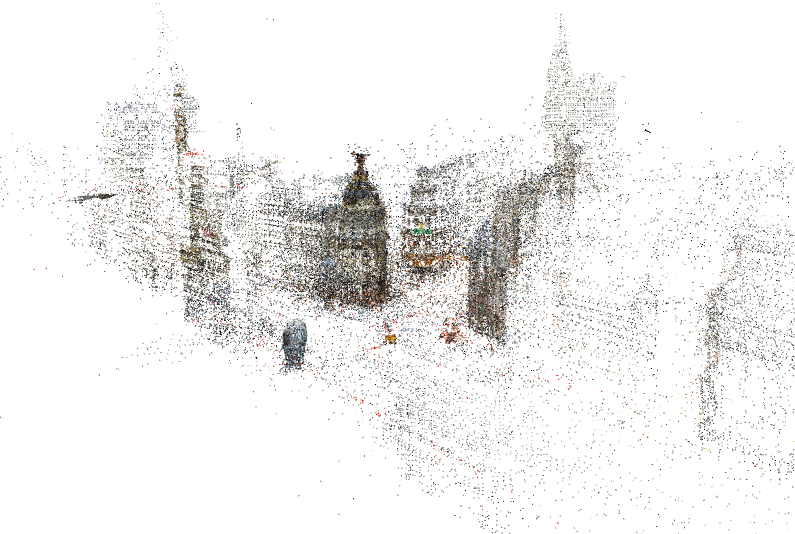}
				\end{minipage}
				&
				\begin{minipage}[b]{0.30\textwidth} 
					\includegraphics [width=1.0\textwidth, trim={2cm 2cm 0cm 0.5cm},clip]{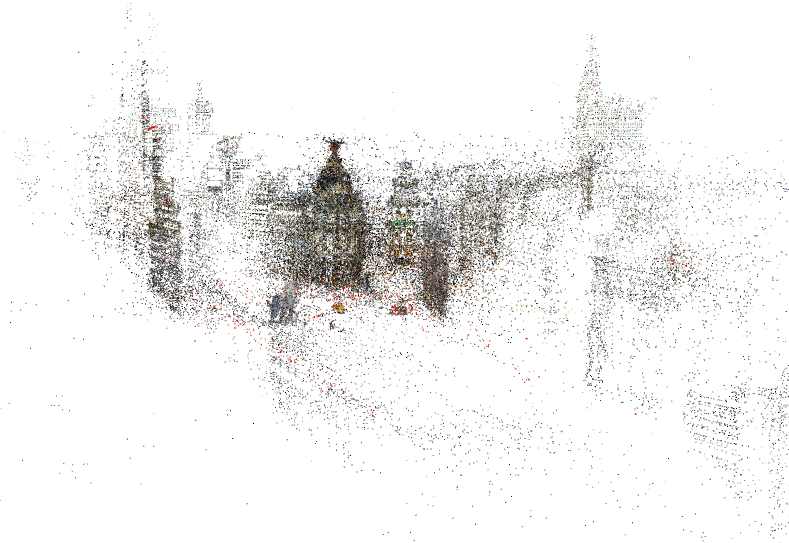}
				\end{minipage}  
				&
				\begin{minipage}[b]{0.30\textwidth} 
					\includegraphics [width=1.0\textwidth, trim={2cm 2cm 0cm  0.5cm},clip]{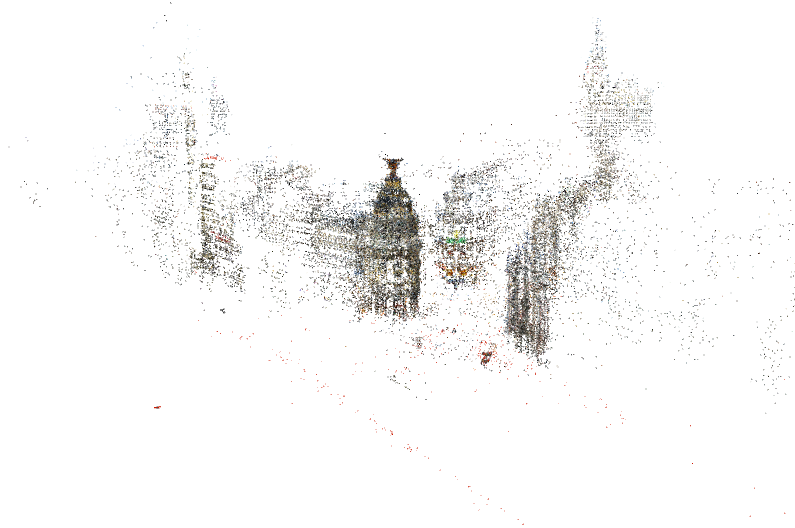}
				\end{minipage} 
				\\
				\Reproj= 1.24, \Track= 7.59    &  \Reproj= 1.21, \Track= 6.99   & \bfseries \Reproj= 1.12, \bfseries \Track= 8.51
				\\
				\\
				\multicolumn{3}{c}{\textbf{\textit{Tower of London}}}    
				\\    \textbf{SuperGlue-10}   & \textbf{SGMNet-10}  & \textbf{\Ours{1}}  	\\ 
				\begin{minipage}[b]{0.30\textwidth}  
					\includegraphics [width=1\textwidth, trim={0cm 0cm 0cm  0cm},clip]{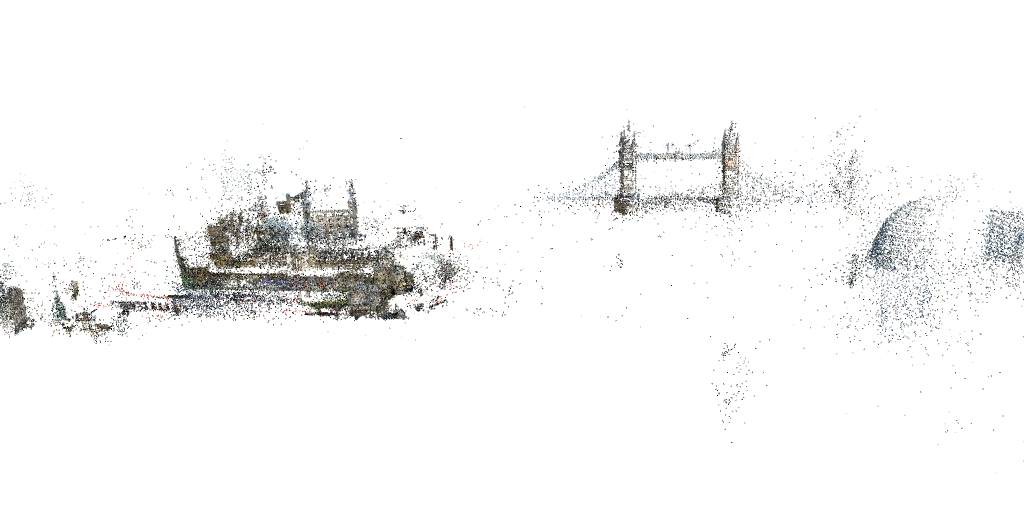}
				\end{minipage}
				&
				\begin{minipage}[b]{0.30\textwidth} 
					\includegraphics [width=1.0\textwidth, trim={0cm 0cm 0cm 0cm},clip]{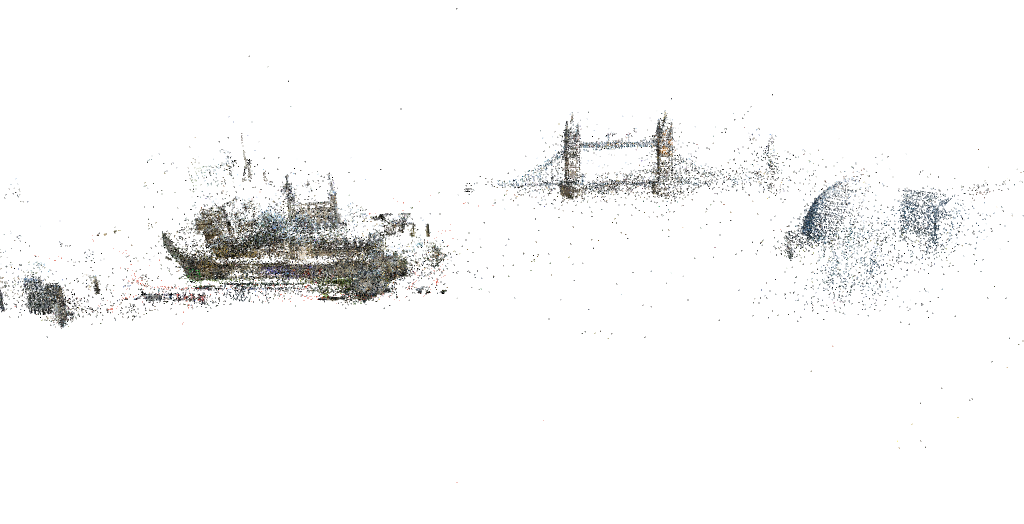}
				\end{minipage}  
				&
				\begin{minipage}[b]{0.30\textwidth} 
					\includegraphics [width=1.0\textwidth, trim={0cm 0cm 0cm  0cm},clip]{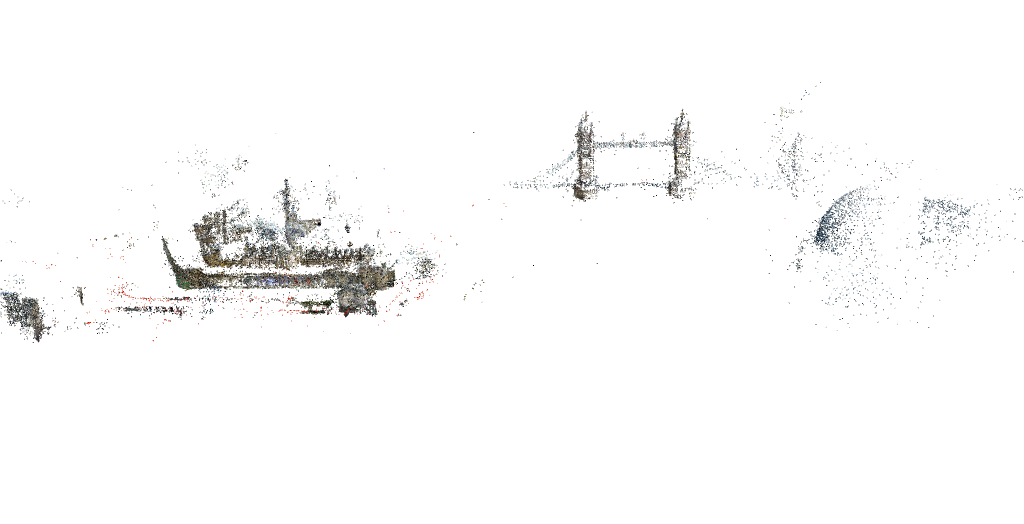}
				\end{minipage} 
				\\
				\Reproj= 1.15,  \Track= 7.27   &  \Reproj= 1.15,  \Track= 6.78  & \bfseries \Reproj= 1.04, \bfseries \Track= 8.49  
				\\    
			\end{tabular}  
		}
	} 
	\caption{3D Reconstruction on medium-size datasets  by SuperGlue (left), SGMNet (middle), and Our \textit{Pair.Neigh} (right). }  
	\label{tab:Supp:3DReconMedium}  
\end{figure*}


\end{document}